\documentclass{article}

\usepackage[main, final]{neurips_2026}

\usepackage[T1]{fontenc}
\usepackage{url}
\usepackage{graphicx}
\usepackage{subcaption}
\usepackage{wrapfig}
\usepackage{booktabs}
\usepackage{nicefrac}
\usepackage{microtype}
\usepackage[dvipsnames]{xcolor}
\usepackage{amsmath}
\usepackage{amssymb}
\usepackage{mathtools}
\usepackage{amsthm}
\usepackage{algorithm}
\usepackage{algorithmic}
\usepackage{setspace}
\usepackage{enumitem}
\usepackage{tcolorbox}
\tcbuselibrary{skins, breakable, listings}

\usepackage{listings}
\usepackage{inconsolata}
\usepackage{fancyvrb}
\usepackage{array}
\usepackage{float}
\usepackage{needspace}
\usepackage{placeins}
\usepackage{hyperref}
\usepackage[capitalize,noabbrev]{cleveref}

\lstdefinestyle{py}{
  language=Python,
  basicstyle=\ttfamily\scriptsize,
  columns=fullflexible,
  keepspaces=true,
  upquote=true,
  frame=single,
  framerule=0.2pt,
  rulecolor=\color{black!20},
  numbers=left,
  numberstyle=\ttfamily\tiny\color{black!50},
  numbersep=5pt,
  breaklines=true,
  breakatwhitespace=false,
  showstringspaces=false,
  tabsize=4,
  backgroundcolor=\color{black!2},
  keywordstyle=\bfseries\color{blue!50!black},
  commentstyle=\upshape\color{green!40!black},
  stringstyle=\color{red!40!black},
}

\lstdefinelanguage{OptiMarkdown}{}

\lstdefinestyle{md}{
  language=OptiMarkdown,
  basicstyle=\ttfamily\scriptsize,
  columns=fullflexible,
  keepspaces=true,
  upquote=true,
  frame=none,
  numbers=none,
  breaklines=true,
  breakatwhitespace=false,
  showstringspaces=false,
  tabsize=2,
  keywords={},
  deletekeywords={and,or,for,if,else,while,not,in,is,from,to,as,with,return,import},
  keywordstyle=\color{black},
  identifierstyle=\color{black},
  commentstyle=\bfseries\color{black},
  stringstyle=\color{black},
  morecomment=[l][\bfseries\color{black}]{\#},
  moredelim=**[is][\bfseries\color{black}]{**}{**},
}

\newcommand{\lt}{\textless}
\newcommand{\gt}{\textgreater}
\newcommand{\tagopen}[1]{\texttt{\lt#1\gt}}
\newcommand{\tagclose}[1]{\texttt{\lt/#1\gt}}

\newtcolorbox{bluebox}[1][]{enhanced, colback=gray!5, colframe=gray, coltitle=white, title=#1,  fonttitle=\bfseries\small, listing only, listing options={
    basicstyle=\ttfamily\footnotesize,
    breaklines=true,
    columns=fullflexible,
    keepspaces=true
  }, sharp corners=south, boxrule=0.8pt, top=4pt, bottom=-2pt, left=6pt, right=6pt,
}

\newtcolorbox{descriptionbox}[1][]{enhanced,
  breakable,
  colback=gray!5,
  colframe=Orchid,
  coltitle=white,
  title=#1,
  fonttitle=\bfseries,
  listing only,
  listing options={
    basicstyle=\ttfamily\footnotesize,
    breaklines=true,
    columns=fullflexible,
    keepspaces=true
  },
  sharp corners=south,
  boxrule=0.8pt,
  top=6pt,
  bottom=6pt,
  left=6pt,
  right=6pt,
}

\newtcolorbox{promptbox}[1][]{enhanced,
  breakable,
  colback=gray!5,
  colframe=brown,
  coltitle=white,
  title=#1,
  fonttitle=\bfseries,
  listing only,
  listing options={
    basicstyle=\ttfamily\footnotesize,
    breaklines=true,
    columns=fullflexible,
    keepspaces=true
  },
  sharp corners=south,
  boxrule=0.8pt,
  top=6pt,
  bottom=6pt,
  left=6pt,
  right=6pt,
}

\DefineVerbatimEnvironment{OldParts}{Verbatim}{commandchars=\\\{\},formatcom=\color{red}}
\DefineVerbatimEnvironment{NewParts}{Verbatim}{commandchars=\\\{\},formatcom=\color{green!60!black}}

\theoremstyle{plain}

\theoremstyle{definition}

\theoremstyle{remark}

\usepackage{amsmath,amsfonts,bm}

\def\eqref#1{equation~\ref{#1}}

\def\1{\bm{1}}

\DeclareMathAlphabet{\mathsfit}{\encodingdefault}{\sfdefault}{m}{sl}
\SetMathAlphabet{\mathsfit}{bold}{\encodingdefault}{\sfdefault}{bx}{n}

\newcommand{\optimindcode}{\href{https://github.com/microsoft/OptiGuide/tree/main/optimind}{\mbox{\raisebox{-0.12em}{\includegraphics[height=0.9em]{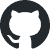}}\hspace{0.25em}GitHub}}}
\newcommand{\optimindmodel}{\href{https://huggingface.co/microsoft/optimind-rl}{\mbox{\raisebox{-0.12em}{\includegraphics[height=0.9em]{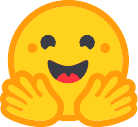}}\hspace{0.25em}Hugging Face}}}

\title{OptiMind: Improving Small Language Models for Optimization Modeling through Formulation Guidance and Multi-Turn Learning
}

\author{Xinzhi Zhang$^{*\dagger}$ \\
  Microsoft Research \\
  \texttt{xinzhizhang@microsoft.com} \\
  \And
  Zeyi Chen$^{\dagger}$ \\
  University of Washington \\
  \texttt{zeyi@uw.edu} \\
  \And
  Humishka Zope$^{\dagger}$ \\
  Stanford University \\
  \texttt{zope@stanford.edu} \\
  \And
  Hugo Barbalho \\
  Microsoft Research \\
  \texttt{hugobarbalho@microsoft.com} \\
  \And
  Konstantina Mellou \\
  Microsoft Research \\
  \texttt{kmellou@microsoft.com} \\
  \And
  Marco Molinaro \\
  Microsoft Research \\
  \texttt{mmolinaro@microsoft.com} \\
  \And
  Janardhan Kulkarni \\
  Microsoft Research \\
  \texttt{jakul@microsoft.com} \\
  \And
  Ishai Menache$^{\ddagger}$ \\
  Microsoft Research \\
  \texttt{ishai@microsoft.com} \\
  \And
  Sirui Li$^{*}$ \\
  Microsoft Research \\
  \texttt{siruili@microsoft.com}
}

\begin{document}

\newlength{\savedtextfloatsep}
\newlength{\saveddbltextfloatsep}
\newlength{\savedintextsep}
\setlength{\savedintextsep}{\intextsep}
\setlength{\intextsep}{8pt}
\raggedbottom
\setlength{\savedtextfloatsep}{\textfloatsep}
\setlength{\saveddbltextfloatsep}{\dbltextfloatsep}
\setlength{\textfloatsep}{8pt}
\setlength{\dbltextfloatsep}{8pt}

\maketitle

\begingroup
\begin{NoHyper}
\renewcommand{\thefootnote}{}
\footnotetext{$^{*}$ Equal contribution.}
\footnotetext{$^{\dagger}$ Part of this work was done during the authors' internships at Microsoft Research.}
\footnotetext{$^{\ddagger}$ Senior author.}
\end{NoHyper}
\endgroup

\begin{abstract}
Mathematical optimization drives decisions in supply chains, logistics, energy, and scheduling, but translating natural-language problems into solver-executable formulations remains a bottleneck. While frontier models show promise for this task, many deployments require small language models (SLMs) that run locally under privacy, latency, and cost constraints. Existing SLMs are limited by noisy data, ambiguous benchmarks, and one-shot inference that ignores the iterative, feedback-driven nature of optimization modeling. To address these challenges, we introduce \textsc{OptiMind}, a system that combines data cleaning and class-specific hints with training using solver feedback. \textsc{OptiMind} distills recurring formulation errors into class-specific hints that guide the repair of ambiguous problems and the regeneration of solution trajectories. This process provides reliable training data, which is then used for multi-turn RL to teach models to revise their formulations in response to solver feedback. At inference time, the same hint library guides further refinement as models revise their solutions over successive turns.
To support reliable evaluation, we clean three challenging benchmarks from the literature. On these cleaned benchmarks, our 20B \textsc{OptiMind-RL} model outperforms strong open-source baselines and reaches frontier-model performance under the same protocol. 
The full hint library, cleaned test data, cleaning logs, and evaluation code are publicly available on \optimindcode, and our trained OptiMind-RL model is available on \optimindmodel.

\end{abstract}

\section{Introduction}
\newlength{\savedmainparskip}
\setlength{\savedmainparskip}{\parskip}
\setlength{\parskip}{3.5pt}

Mathematical optimization plays a critical role across many business sectors, from supply-chain management to energy systems to logistics planning, where effective decision-making relies on solving highly complex optimization problems. While practitioners can usually describe these problems in natural language, translating them into precise mathematical formulations that optimization solvers can process remains a skill-intensive bottleneck. Crafting a correct formulation requires precise definition of decision variables, objectives, and constraints, a skill that typically takes years of specialized training in operations research to develop \citep{simchi2025democratizing}. Researchers and practitioners have begun exploring whether large language models (LLMs) can automate the formulation task: translating natural language problem descriptions into executable optimization models~\citep{pmlr-v220-ramamonjison23a,tang2024orlm,yang2025optibench}. Success on this front would lower a major barrier to broader use of optimization, democratizing its benefits. Yet current systems remain far from this goal.

Frontier reasoning models such as GPT-5, Gemini 3, and Claude Opus can often perform Linear Programming (LP), Integer Programming (ILP) and Mixed Integer and Linear Programming (MILP) formulation tasks effectively. However, many real-world applications require smaller models deployable locally. This is especially important in supply chain and logistics settings, where data privacy constraints, latency requirements, and cost considerations often make cloud-based frontier models impractical. For example, warehouses, distribution centers, and other operational sites may need to keep sensitive local data on-premises while still generating high-quality optimization formulations in real time \citep{corradini2025state,li2024small}.
Unfortunately, the performance of open-source SLMs remains far from satisfactory. This is a significant concern given the practical importance of the task: a poorly formulated optimization model can lead to substantial downstream losses, including higher operational costs, inefficient resource allocation, and increased supply-chain latency.

A core challenge is that mathematical programming formulation is a complex and specialized domain where high-quality training data is scarce. Unlike general programming, optimization modeling lacks large-scale naturally occurring corpora, such as GitHub code or Stack Overflow-style question-answer data, that provide high-quality and diverse examples of LP/ILP formulations. Existing data is often limited in coverage, uneven in quality, and prone to errors, especially when generated synthetically. As a result, SLMs receive weak supervision and struggle to generalize across problem families.

On the other hand, current inference approaches also leave significant structure unused. Most methods rely on a simple prompt containing the natural-language problem statement and a generic modeling goal, such as generating an integer programming formulation in Pyomo, OR-Tools, or GurobiPy. This ignores richer strategies that incorporate formulation schemas, class-specific modeling hints, and iterative refinement. The issue is further compounded by noisy benchmarks: ambiguous questions, missing data, and incorrect ground-truth values can produce misleading evaluations, especially when benchmarks are not carefully cleaned by domain experts. Overall, existing approaches underutilize optimization expertise at every stage: benchmark construction, data cleaning, training, and inference.
Our goal in this work is to substantially close this gap and make strong open-source ILP formulation models a realistic possibility. We ask:

\emph{Can we train small, efficient, open-source models that are competitive with frontier models on LP/ILP formulation, while remaining cheap enough for on-device or large-scale deployment?}

We address this challenge with \textsc{OptiMind}, an integrated system for improving small language models on optimization formulation. Our approach starts from the observation that models often repeat similar formulation errors on related problems. To this end, we can distill these errors into compact, class-specific hints that provide guidance to avoid common formulation errors. We then use the hints to generate more reliable solutions for training problems, while also correcting errors and resolving ambiguities in the problem statements.

Using this cleaned data, we finetune our model with reinforcement learning. Our RL pipeline incorporates multiple turns of interaction with the solver, using the resulting feedback to revise the formulation.
At inference time, we further enhance formulation reliability by combining guidance from the hint library with additional rounds of revision using solver feedback.

\textbf{Our Contributions:}

        \textbf{1. Competitive Small Models.}
We train a domain-specialized 20B language model for optimization formulation.
Built on \textsc{gpt-oss-20b}, \textsc{OptiMind-RL} achieves $78.5\%$, $92.3\%$, and \space $91.0\%$ accuracy on IndustryOR, Mamo-Complex, and OptMATH when used with our hint-guided five-turn inference pipeline.
This outperforms the strongest open-source baselines up to 32B parameters by $12.3$--$27.2$ percentage points.
It is also competitive with GPT-5 without hints at the same turn budget, exceeding it by $1.1$--$2.9$ percentage points on Mamo-Complex and OptMATH and trailing by $4.7$ percentage points on IndustryOR.

        \textbf{2. Learning to Formulate and Revise with Solver Feedback.}
    We construct a multi-turn optimization environment for training models with standard GRPO and solver feedback. The \emph{average-turn correctness reward} encourages repairing incorrect formulations and retaining correct ones. Experiments show improved generalization to unseen problem structures and nonlinear formulations, and successful training across model families using the same cleaned corpus. The RL model also largely retains the base model's tool-use and general benchmark performance.

    \textbf{3. High Quality Optimization Data and Formulation Guidance.} 
    We improve data quality for optimization-focused LLM training, evaluation, and inference along three dimensions. 
    First, we develop a scalable hint- and LLM-assisted data-cleaning pipeline that identifies and removes erroneous training examples in OR-Instruct and OptMATH, correcting up to $30\%$ of the data and providing supervised fine-tuning trajectories and ground-truth objective values for reinforcement-learning rewards.
    Second, we audit and correct widely used evaluation benchmarks IndustryOR, MAMO-Complex, and OptMATH, identifying errors in $50\%$, $51\%$, and $41\%$ of their instances, respectively.    
            Third, we introduce a library of approximately 200 class-specific hints that substantially improves inference-time performance. We have released the full hint library, cleaned test data, cleaning logs, and evaluation code on \optimindcode, and our trained OptiMind-RL model on \optimindmodel.

\begin{figure*}[t]
\centering
\resizebox{0.94\textwidth}{!}{\begin{minipage}{\textwidth}

\begin{subfigure}[t]{0.55\textwidth}
    \vspace{0pt}
    \begin{bluebox}[Mamo-Complex Transport Problem and Hint]
    \begin{minipage}[t]{\linewidth}
    \begin{spacing}{0.82}
    \ttfamily\tiny\linespread{0.9}\selectfont\raggedright
    \setlength{\abovedisplayskip}{1pt}
    \setlength{\belowdisplayskip}{1pt}
    \setlength{\abovedisplayshortskip}{1pt}
    \setlength{\belowdisplayshortskip}{1pt}

    \textbf{Problem:}
    Each region has an amount of medicines, but they each need different quantities to adequately meet their healthcare demands.  It is given the supply and demand of each node, and transportation costs per pair of nodes.

    \vspace{2pt}
    
    \textbf{Hint:}
    \begin{itemize}[leftmargin=1.2em,itemsep=0pt,topsep=1pt,parsep=0pt,partopsep=0pt]
        \item  Enforce flow conservation with the \textbf{appropriate signs}. If you have more total supply than total demand, \textbf{use inequality} instead of equality.
        \item  If the problem includes a \textbf{fixed cost} to open/use a route (edge), introduce a binary variable and include a fixed charge term in the objective
        \item You do not need to directly send the flow from supply to demand, you can \textbf{allow intermediate nodes} unless explicitly forbidden.
        \item Do not try to solve the problem by \textbf{brute force}
    \end{itemize}

    \end{spacing}
    \end{minipage}
    \end{bluebox}

\begin{bluebox}[Base Model with Hint \color{red!50}{(Incorrect)}]
    \begin{minipage}[t]{\linewidth}
    \begin{spacing}{0.82}
    \ttfamily\tiny\linespread{0.9}\selectfont\raggedright
    \setlength{\abovedisplayskip}{1pt}
    \setlength{\belowdisplayskip}{1pt}
    \setlength{\abovedisplayshortskip}{1pt}
    \setlength{\belowdisplayshortskip}{1pt}

    \textbf{Thinking:}
    ``$x_{ij}$ -- quantity of supplies shipped {\color{ForestGreen} \textbf{from region \(i\) to region \(j\)}}''
    
    \vspace{1pt}
    ``The total supply exceeds the total demand. We {\color{red} \textbf{add a dummy demand node}} that `takes' this surplus and has no supply.''
    
    \vspace{2pt}
    \textbf{Solution:}
    The solution adds edges between all pairs of nodes, not just from supply to demand, to allow the use of intermediate nodes. 

    \vspace{1pt}
    It also adds a dummy node to absorb excess supply, and connects it to all other nodes at cost 0.

    \vspace{2pt}
    \textbf{Explanation:}
    The answer is still incorrect. While the hinted base model applied Hint 3 to allow intermediate nodes, it did not apply Hint 1 to correctly handle excess supply.

    \end{spacing}
    \end{minipage}
    \end{bluebox}
    
\end{subfigure}
\hfill
\begin{subfigure}[t]{0.43\textwidth}
    \vspace{0pt}
        \begin{bluebox}[Base Model without Hint \color{red!50}{(Incorrect)}]
    \begin{minipage}[t]{\linewidth}
    \begin{spacing}{0.82}
    \ttfamily\tiny\linespread{0.9}\selectfont\raggedright
    \setlength{\abovedisplayskip}{1pt}
    \setlength{\belowdisplayskip}{1pt}
    \setlength{\abovedisplayshortskip}{1pt}
    \setlength{\belowdisplayshortskip}{1pt}

    \textbf{Solution:}
    ``A transport from a surplus node to another surplus node would not help satisfy any deficit and would only increase cost, because all transportation costs are non-negative.  
Therefore {\color{red} \textbf{we only consider arcs from each surplus node to each deficit node}}.''

    \vspace{4pt}
    \textbf{Explanation:}
    The answer is incorrect because for some nodes, it is cheaper to send units between them by using an intermediate node, even with all costs being positive.

    \end{spacing}
    \end{minipage}
    \end{bluebox}

\begin{bluebox}[\textsc{OptiMind-RL} with Hint \color{ForestGreen!50}{(Correct)}]
    \begin{minipage}[t]{\linewidth}
    \begin{spacing}{0.82}
    \ttfamily\tiny\linespread{0.9}\selectfont\raggedright
    \setlength{\abovedisplayskip}{1pt}
    \setlength{\belowdisplayskip}{1pt}
    \setlength{\abovedisplayshortskip}{1pt}
    \setlength{\belowdisplayshortskip}{1pt}

    \textbf{Thinking:}
     ``For each region $i$, the amount of supplies available after {\color{ForestGreen}\textbf{shipments plus inbound shipments}} must be at least $d_i$''
    
    "That means there's excess supply overall. {\color{ForestGreen}\textbf{The hint says:}} `If you have more total supply than total demand you must {\color{ForestGreen}\textbf{use inequality}} instead of equality'."

    \vspace{2pt}
    \textbf{Solution:}
    The solution adds edges between all pairs of nodes, not just from supply to demand, to allow the use of intermediate nodes. 

    \vspace{1pt}
    It uses inequalities in the flow conservation constraints to handle excess supply. 
    
    \vspace{2pt}
    \textbf{Explanation:}
    The answer is {\color{ForestGreen} \textbf{correct}}. The RL + hint rollout now applies both Hint 1 and Hint 3 to correctly model the problem.
    \end{spacing}
    \end{minipage}
    \end{bluebox}

\end{subfigure}

\end{minipage}}
\caption{Comparison of rollout trajectories between base model without hint,
base model with hint, and \textsc{OptiMind-RL} with hint on Mamo-Complex 
problem\_index 147. Red marks the key failures.}
\label{fig:multi_turn_rl}
\end{figure*}

\section{Description of the Formulation Task}

We study the \emph{formulation task}: translating a complete natural-language description of a decision-making problem, including all required data, into an executable MILP/ILP model. A MILP problem is formulated as
\(
\min\{c^\top x : Ax \leq b,\ x_j \in \mathbb{Z}\ \forall j \in I\},
\)
where $x \in \mathbb{R}^n$ is the vector of decision variables, $c \in \mathbb{R}^n$ is the cost vector, $A \in \mathbb{R}^{m \times n}$ and $b \in \mathbb{R}^m$ define the linear constraints, and $I \subseteq \{1,\dots,n\}$ indicates the integer-valued variables.
The input of the task is the description of the optimization problem in natural language, and the output is a Python code that constructs the optimization model, calls a solver such as Gurobi, and prints the optimal solution. Running this code therefore produces the final decisions recommended for the input problem. Fig.\ref{fig:optimind} illustrates the pipeline, and Appendix~\ref{app:inout} gives a complete input/output example.

\section{Guiding Data Improvement Through Distilled Expert Hints}
\label{sec:error_analysis}

We focus on three challenging, established optimization benchmarks in our experiments: IndustryOR~\citep{tang2024orlm}, Mamo-Complex~\citep{huang2024mamo}, and OptMATH~\citep{lu2025optmath}. These are substantially harder than earlier LP-centric datasets: strong models achieve only 20\%--50\% accuracy~\citep{jiang2025llmopt,liu2025optitree,lu2025optmath}, whereas earlier LP-centric benchmarks such as NL4LP~\citep{pmlr-v220-ramamonjison23a} and NLP4LP~\citep{ahmaditeshnizi2024optimus} reached 90--95\%+ in prior evaluations~\citep{liu2025optitree,lu2025optmath}. However, we observed that low accuracy reflected not only model limitations but, importantly, benchmark quality.

\begin{figure}[t]
\centering
\begin{minipage}{0.48\textwidth}
    \begin{center}
\includegraphics[width=\linewidth]{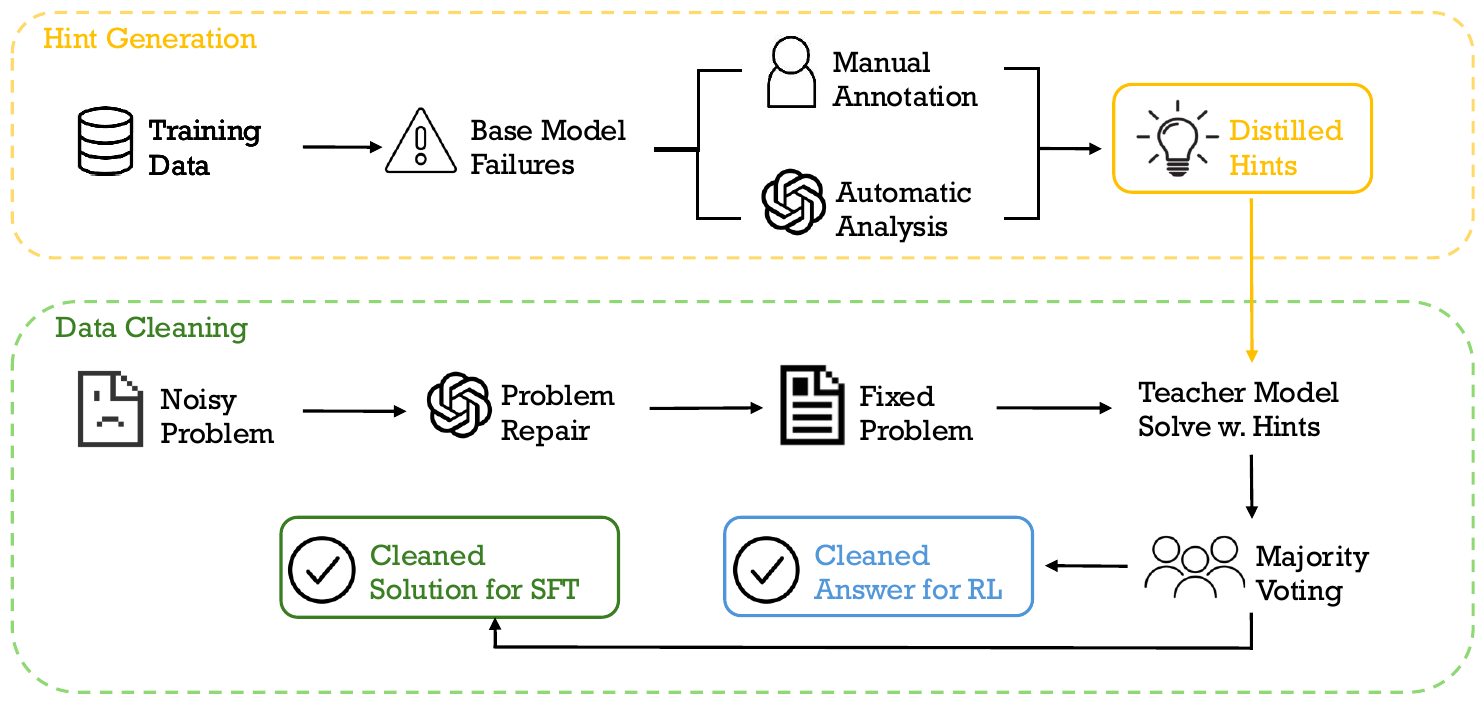}  
\end{center}
\caption{Overview of \textsc{OptiMind}'s data-improvement pipeline. We derive reusable hints from base-model errors, then use them to repair problems, regenerate solutions, and determine ground-truth objective values.}
\label{fig:data-pipeline}
\end{minipage}
\hfill
\begin{minipage}{0.48\textwidth}
   \centering
\vspace{8pt}
\includegraphics[width=\linewidth]{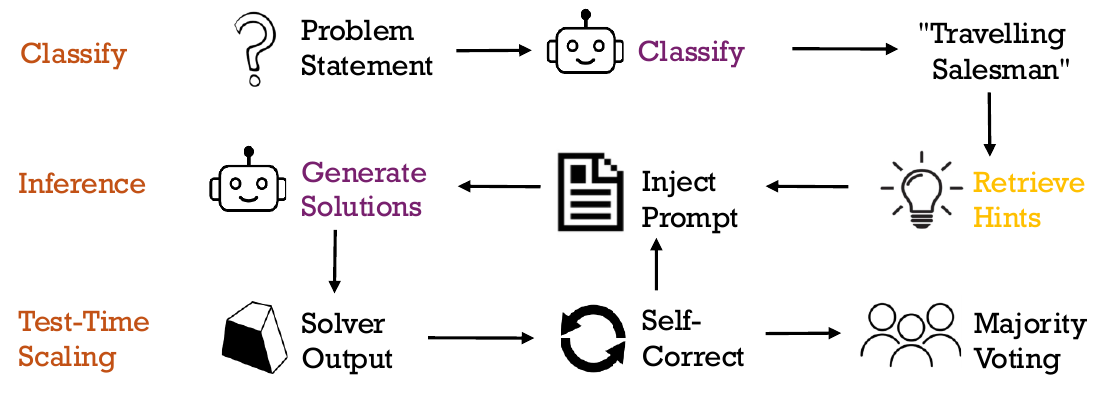}
\vspace{6pt}

\caption{\textsc{OptiMind} inference pipeline. The hint library distilled during data improvement is reused at test time for class-conditioned prompting, multi-sample decoding, and optional solver-feedback self-correction.}
\label{fig:training_inference_pipeline}
\end{minipage}
\end{figure}

\subsection{Quality Issues in Existing Datasets}

We identify significant residual errors in these benchmarks, including missing or ambiguous data, incorrect answers, infeasible instances, variable-type ambiguities, and rounding inconsistencies. In the original test sets, $50\%$ of IndustryOR instances, $51\%$ of Mamo-Complex, and $41\%$ of OptMATH instances require correction or clarification (Table~\ref{tab:error_breakdown}). Despite recent efforts to improve benchmark quality~\citep{tang2024orlm,chen2025solver}, these issues remain non-negligible; representative examples and cleaning statistics are given in Appendices~\ref{app:supp-data-correction} and~\ref{app:benchmark_cleaning}. 

To support reliable evaluation in this work and future research, our team of optimization experts (OR faculty and PhD students) reviewed the test sets of these benchmarks four times over a month. We release the cleaned benchmarks and HTML cleaning logs documenting the changes
(Appendices~\ref{app:benchmark_cleaning} and~\ref{app:anonymous_repo}). After cleaning, the same \textsc{gpt-oss-20b} model and inference strategy improved from 40\%--60\% accuracy on the original releases to 70\%--90\% on our corrected test sets.

\space
We also noticed that training data associated with these benchmarks exhibits similar quality issues. As a preliminary exploration of the reliability of this data, we applied SFT followed by RL to Qwen3-8B using the original training data. Accuracy improved from roughly $25\%$ to $49\%$, but remained well below the $\sim70\%$ achieved by \textsc{gpt-oss-20b} under cleaned evaluation. Although the different base models preclude attributing this gap to data quality alone, the result reinforces our concern that errors in the original training data may limit its usefulness for learning. Such errors affect both SFT targets and RL rewards: incorrect reference answers or ambiguous problem statements can cause training to reinforce incorrect formulations.

Unlike the test sets, which we cleaned manually, the training corpora are too large for exhaustive manual review. Regenerating training solutions with a stronger LLM offers a scalable alternative, but even strong models still make formulation errors often enough to compromise the reliability of the regenerated data. Thus, \textsc{OptiMind} begins by developing a more reliable procedure for cleaning the training data. 
\space

\subsection{Model Failures Reveal Recurring Formulation Errors}

Our key observation is that models' formulation errors are highly structured, with similar mistakes recurring across related problems. To validate this observation, we used \textsc{gpt-oss-20b} to generate formulations for training problems from OR-Instruct (the training set associated with the IndustryOR benchmark)~\citep{tang2024orlm} and OptMATH~\citep{lu2025optmath} and inspected the resulting formulation errors.\footnote{{}We remark that Mamo-Complex does not provide a corresponding training set.}
 Recurring errors include incorrectly applying MTZ subtour-elimination constraints to the fixed start node in TSP problems, swapping inflow and outflow signs in network-flow and inventory-balance constraints, omitting capacity or linking constraints in facility-location and production-planning problems, and assuming integrality when the statement only specifies continuous quantities. These patterns motivate a \emph{compact class-conditioned hint library} to help correct recurring mistakes.

\subsection{Distilling Domain Knowledge into Reusable Hints}

To turn these recurring formulation errors into reusable guidance, we build a hint library within the \textsc{OptiMind} framework.
For each recurring failure mode within a class, the library stores a short error summary and a preventive modeling rule. When generating a formulation, the model receives the hints associated with the problem's class, instructing it to apply only those relevant to the particular instance. The goal is not to encode full formulations, but to provide reusable
reminders that reduce systematic mistakes during solution regeneration, reward construction, and
inference. We classify problems using the 53 seed classes defined by the OptMATH generators, which provides good coverage of typical problem types. Table~\ref{tab:hint_examples} provides hint examples, and Figure~\ref{fig:sample-hints} exemplifies the impact of hints on formulation errors.

The hint library is built through two complementary routes\unskip: In the \textit{expert-curated} route, optimization experts inspect representative mismatches between \textsc{gpt-oss-20b} outputs and dataset answers, determine whether the disagreement is caused by a model error or a data issue, and write reusable corrective hints when the model is at fault. In the \textit{automatic} route, a stronger teacher model identifies high-confidence disagreement cases, synthesizes candidate hints, refines them into class-level guidance, and retains only hints that improve validation accuracy. The full automatic pipeline, prompt templates, and open-source teacher variant are given in Appendix~\ref{app:auto_hint_generation}.

When both sources are available, \textsc{OptiMind} merges expert-curated and automatic hints into a hybrid library. The resulting library is used throughout the data pipeline and is available at inference time as optional class-conditioned guidance.

Interestingly, even though our expert-curated hints were developed from recurring formulation errors made by GPT-OSS-20B, applying them unchanged to Qwen3.5-9B improves accuracy across all three benchmarks (Appendix~\ref{app:qwen_transfer}). We also note that the hint library can be extended to specialized domains beyond those covered by the 53 OptMATH seed classes through our automatic hint-generation pipeline, without requiring expert review.

\subsection{Cleaning Training Data with Hints}
\label{sec:training_data_cleaning}

\textsc{OptiMind} uses the hint library to clean the training corpora used to produce solution trajectories for SFT and ground-truth objective values for RL. The cleaning process has two goals: improving question quality and improving solution quality. For OR-Instruct, the training set associated with the IndustryOR benchmark, we automatically detect and fill missing problem description data using OpenAI o4-mini; this flags $180/2700$ items as incomplete, which are filled with feasible values and verified by optimization experts. For OptMATH, we follow the original generation process to regenerate 50 instances for each of the 53 seed classes, back-translate the problem descriptions with \textsc{gpt-oss-120b}, remove infeasible instances, and manually inspect three regenerated instances per class.

To improve solution labels, \textsc{OptiMind} regenerates solutions with \textsc{gpt-oss-120b} using the class-specific expert hints, then selects answers by majority voting over $K{=}64$ samples. It also balances the training set by sampling up to 100 instances per class. This yields $2.2$k cleaned OR-Instruct items and $2.3$k regenerated OptMATH questions; among OR-Instruct items, $602/2700$ final answers differ from the original labels. These cleaned problem statements and ground-truth objective values can be reused across model families. To adapt \textsc{OptiMind} to a different base model, the same corpus can be used directly for RL. For SFT, solution trajectories can be regenerated for that model family. Neither approach requires repeating the expert audit. 
Figure~\ref{fig:data-pipeline} depicts \textsc{OptiMind}'s data cleaning pipeline, and full training-data cleaning details and examples are provided in Appendix~\ref{app:data_cleaning}.

\section{From Cleaned Data to Multi-Turn RL}
\label{sec:training_rl}

The data-improvement pipeline in \S\ref{sec:error_analysis} produces cleaned problem descriptions,
regenerated solution trajectories, class-level hints, and ground-truth objective values. SFT on the
regenerated trajectories shows that models can learn to formulate optimization problems from
cleaned supervision. In our SFT setup, the model is trained to produce a solution without observing
feedback from its own execution during training. However, because optimization formulations are executable,
solver status, logs, objective values, and printed decisions can help identify syntax errors,
runtime failures, and modeling mistakes. We therefore train \textsc{OptiMind-RL} to use this
feedback when revising its generated programs. 

\textbf{From Prompted Self-Correction to Learned Repair.}
One encouraging observation in this direction is the following: Multi-turn inference shows that many remaining SFT and base-model errors are structured and
repairable once solver feedback is observed. In Table~\ref{tab:main_results_comprehensive}, the base
model and \textsc{OptiMind-SFT} both improve substantially from one to five turns. For example,
without hints, \textsc{OptiMind-SFT} improves from $85.77$ to $90.71$ on Mamo-Complex and
from $72.95$ to $81.64$ on OptMATH; the base model shows even larger multi-turn gains on these
datasets. These improvements suggest that many failures are not arbitrary generation errors, but
formulation errors that execution can help localize.

A concrete example is presented in Appendix~\ref{app:multiTurnEx}, which shows an IndustryOR instance where the optimal solution of the generated formulation shows inventory exceeding internal capacity without external storage. After inspecting this output, the language model corrects an error in the direction of an inequality on the next turn. This example motivates including solver feedback in RL rollouts to train models to identify and correct modeling errors even when execution succeeds. Appendix~\ref{app:rl-trajectories} further shows how the model corrects modeling errors, diagnoses infeasibility, and leaves correct formulations unchanged.

\textbf{Multi-Turn RL with Solver-Verified Rewards.} ~The training of \textsc{OptiMind-RL} is conducted as follows.
Each training instance consists of a cleaned problem $q$ and its ground-truth objective value
$z^*$. Given a response $y$, we extract the latest Python code block,
execute it, and parse the printed objective value $\hat z(y)$ when available. We define
\emph{executable correctness} as
\[
C(q,y)=
\mathbf{1}\!\left[
\operatorname{Exec}(y)\text{ succeeds}
\land
\hat z(y)\text{ is parsed}
\land
\frac{|\hat z(y)-z^\star|}{|z^\star|+1}
\le 10^{-6}
\right].
\]
The ground-truth objective value is hidden from the model and used only for reward computation.

A rollout is a finite-horizon dialogue between the policy and the solver. At turn $t$, the policy
generates $y^{(t)}\sim\pi_\theta(\cdot\mid h_t)$, the environment executes the latest available
candidate program $\bar y^{(t)}$, and the resulting stdout, stderr, solver status, and printed objective
are appended to the next history. If a later response does not contain a new code block, we carry
forward the latest candidate. We train with $T=3$ turns and reward average executable correctness: $
R(\tau)=\frac{1}{T}\sum_{t=1}^{T} C(q,\bar y^{(t)}).$ As Table~\ref{tab:rl_hint_ablation} shows, averaging the reward across turns supports both accurate initial formulations and effective self-correction using solver feedback; changing a correct formulation into an incorrect one lowers the trajectory reward, encouraging the model to retain correct solutions. 
We optimize this reward with GRPO~\citep{deepseekmath} in the \textsc{Slime}
framework~\citep{slime_github}; prompts, execution details, and hyperparameters are provided in
Appendix~\ref{app:training-details}.

 \section{Enhancing Formulation Robustness at Test Time}
\label{sec:inference}

The training procedure discussed in Section~\ref{sec:training_rl} produces a model specialized in optimization formulation. In addition, we use class-specific hints at test time to improve the correctness of the model's formulations. Given a new optimization problem, the model first identifies its canonical problem class, retrieves the corresponding error summaries and hints, and uses them as a conditional checklist when generating the formulation. The model applies only hints whose conditions are present in the problem statement; for example, a backlog-balance hint applies only when backorders are explicitly allowed.

When more inference budget is available, we apply test-time scaling techniques such as majority voting and self-correction to improve the robustness of generated formulations. For self-correction, we repeatedly provide the model with its most recent program and execution output, including runtime error messages and Gurobi logs, and ask it to revise the program in the next turn if it detects a coding error or believes its solution is still incorrect. We illustrate the pipeline in Figure~\ref{fig:training_inference_pipeline} and provide detailed prompts and inference pseudocode in Appendix~\ref{app:inference_details}.

\section{Experiments}\label{sec:experiments}

\subsection{Experimental Setup}
\label{sec:experimental_setup}

\textbf{Datasets.}
We evaluate on rigorously cleaned versions of three challenging optimization-modeling benchmarks:
\textbf{IndustryOR}~\citep{tang2024orlm}, \textbf{Mamo-Complex}~\citep{huang2024mamo}, and
\textbf{OptMATH}~\citep{lu2025optmath}. Throughout the paper, we use OptMATH to denote the subset of linear problems in the original OptMATH dataset~\citep{lu2025optmath}, reflecting our focus on ILP and MILP problems. We provide a comprehensive study of the nonlinear OptMATH subset in Appendix~\ref{app:nonlinear}. These benchmarks contain 469 expert-validated instances. We use these cleaned sets as the main
evaluation suite because benchmark correctness is especially important for measuring formulation
ability: ambiguous statements or incorrect ground-truth objective values can otherwise reward incorrect models.
In addition to the three standard benchmarks, we test \textsc{OptiMind}'s generalization by constructing an eight-problem out-of-taxonomy dataset from OR papers on arXiv, which we denote as the \emph{arXiv dataset}. We also evaluate the out-of-taxonomy LogiOR dataset; details are provided in Appendices~\ref{app:ood_generalization} and~\ref{app:standalone_logior}.
Additional details on benchmark selection, cleaning, and parallel-cleaned variants are provided in Appendix~\ref{app:evaluation_experiment_details}.

\textbf{Metrics.}
Our primary metric is solution accuracy, reported after the first generation turn and after five turns
of solver-feedback self-correction. Each model is prompted to produce an executable Python script
using GurobiPy. A prediction is marked correct if the printed objective value $\hat{z}$ matches the
dataset's ground-truth objective value $z^\star$ under the normalized-error criterion
$|\hat{z}-z^\star|/(|z^\star|+1) < 10^{-6}$, following OptMATH~\citep{lu2025optmath}. 
\space 
This objective-value matching serves as a proxy for formulation correctness. To validate this commonly used metric, we also performed an audit on OptiMind-RL outputs: among those whose objective values matched the ground-truth objective values, our optimization experts audited 30 randomly sampled outputs across the three benchmarks and found that all 30 formulations were correct.
Details on objective parsing,
majority voting, RL reward computation, and repeated evaluations are given in
Appendix~\ref{app:evaluation_experiment_details}.

\textbf{Baselines and model variants.}
We compare OptiMind with open-source optimization-modeling baselines up to 32B parameters,
including \textsc{gpt-oss-20B}, \textsc{Qwen3-32B}, and \textsc{SIRL-Gurobi32B}~\citep{chen2025solver} (based on the Qwen3-32B family),
as well as frontier models \textsc{o4-mini} and \textsc{GPT-5}. 
We evaluate the base model, \textsc{gpt-oss-20b}, and our two trained variants, \textsc{OptiMind-SFT} and
\textsc{OptiMind-RL}, and ablate data cleaning, error-analysis-based hints, and test-time scaling to quantify the contribution of each component.

\textbf{Evaluation protocol.}
We evaluate our trained models alongside the baselines under the same hint settings and number of self-correction turns to compare the effects of training and inference-time guidance (Table~\ref{tab:main_results_comprehensive}). These comparisons use single-sample decoding ($K=1$); majority voting is examined separately in the test-time-scaling study.

\subsection{ Main Results and Component Contributions}

\begin{figure}[ht]
\centering
\includegraphics[width=0.8\linewidth]{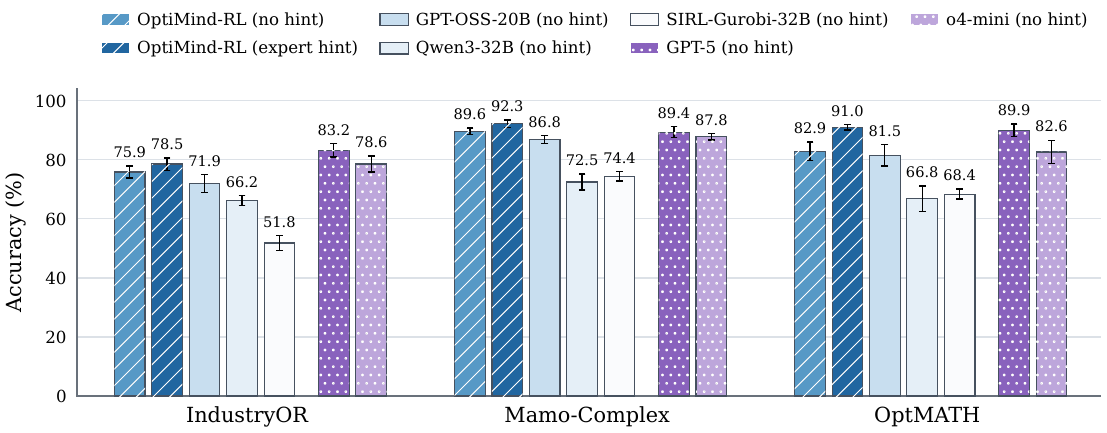}
\caption{Average accuracy for different models under 5 turns. Models colored in blue are open-sourced models up to 32B parameters, and models in purple are closed-sourced frontier models.}
\label{fig:main_bar_plot}
\end{figure}

We now assess the impact of OptiMind's pipeline on solution accuracy.

\textbf{Overall comparison.}
Figure~\ref{fig:main_bar_plot} reports results on our three expert-cleaned benchmarks, with Table~\ref{tab:main_results_comprehensive} providing additional details. In the single-turn, no-hint scenario, \textsc{OptiMind-RL} improves the base model by $+5.7$ percentage points on IndustryOR, $+20.3$ points on Mamo-Complex, and $+9.0$ points on OptMATH. Under our full inference pipeline with expert hints and 5-turn correction, the gains over the single-turn, no-hint base model increase to $+12.3$, $+27.2$, and $+24.3$ points, respectively. Overall, \textsc{OptiMind-RL} clearly outperforms similarly sized open-source models on all three benchmarks and remains competitive with much larger frontier models when all models use five turns, with expert hints for \textsc{OptiMind-RL} and no hints for the baselines (Table~\ref{tab:main_results_comprehensive}). We observe consistent trends on additional cleaned benchmarks from prior work (see Table~\ref{tab:other_benchmarks} in Appendix~\ref{app:additional_eval_results}).

\textbf{Contributions of training and inference.}
\space
Table~\ref{tab:component_decomposition} compares OptiMind's training and inference settings at $K=1$, with SFT and RL trained independently from the same base model. The results show accuracy gains from training, hints, and additional turns of solver feedback. 
At one turn without hints, cleaned-data SFT exceeds the base by $2.7/20.7/6.3$ points on IndustryOR, Mamo-Complex, and OptMATH, respectively, while RL differs from SFT by $+3.0/-0.4/+2.8$ points. With expert hints, RL exceeds SFT by $5.3/0.9/5.2$ points at one turn and $3.3/0.2/2.5$ at five turns. Within RL, expert hints add $3.8/4.6/9.5$ points at one turn, and iterative feedback from one to five turns adds another $2.8/2.3/5.8$ points with expert hints.
Under five-turn inference with expert hints, RL gains over SFT are largest on IndustryOR and smallest on Mamo-Complex, consistent with differing benchmark headroom (Appendix~\ref{app:benchmark_headroom}; additional ablations in \S\ref{sec:ablation}).

\begin{table}[htbp]
\caption{Component comparisons under $K=1$. SFT and multi-turn RL are independently trained from GPT-OSS-20B. All values are accuracy (\%); full deviations appear in Table~\ref{tab:main_results_comprehensive}.}
\label{tab:component_decomposition}
\centering\small\setlength{\tabcolsep}{2.5pt}
\renewcommand{\arraystretch}{0.94}
{\begin{tabular}{llrrrrrr}
\toprule
& & \multicolumn{2}{c}{IndustryOR} & \multicolumn{2}{c}{Mamo-Complex} & \multicolumn{2}{c}{OptMATH} \\
\cmidrule(lr){3-4}\cmidrule(lr){5-6}\cmidrule(lr){7-8}
Model & Hints & 1 turn & 5 turns & 1 turn & 5 turns & 1 turn & 5 turns \\
\midrule
GPT-OSS-20B & None & 66.23 & 71.89 & 65.03 & 86.80 & 66.68 & 81.52 \\
SFT & None & 68.97 & 72.43 & 85.77 & 90.71 & 72.95 & 81.64 \\
Multi-turn RL & None & 71.97 & 75.88 & 85.36 & 89.58 & 75.70 & 82.92 \\
SFT & Expert & 70.50 & 75.25 & 89.09 & 92.04 & 79.93 & 88.50 \\
Multi-turn RL & Expert & 75.76 & 78.54 & 90.00 & 92.26 & 85.17 & 90.99 \\
\bottomrule
\end{tabular}}
\end{table}

\begin{figure}[ht]
\centering
\begin{minipage}{0.45\linewidth}
    \centering
    \includegraphics[width=\linewidth]{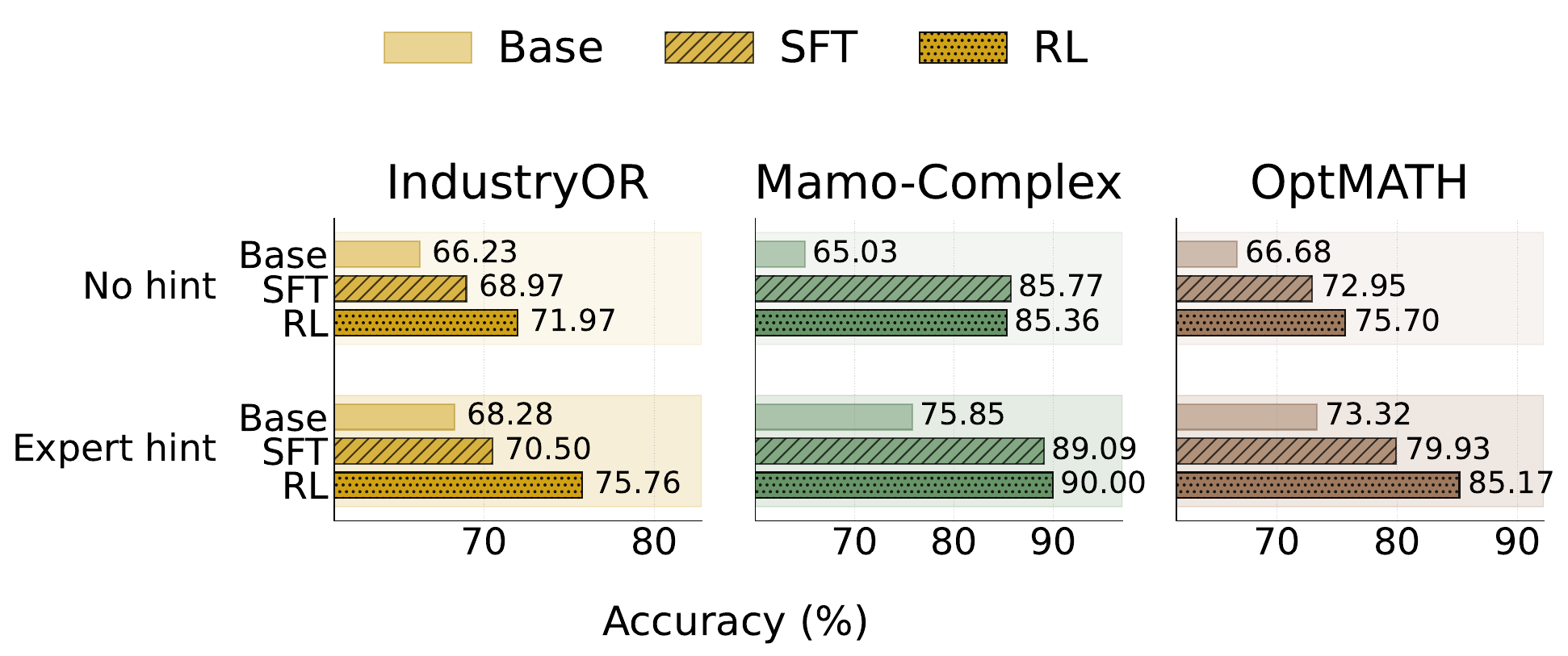}

\end{minipage}
\hfill
\begin{minipage}{0.53\linewidth}
    \centering
    \includegraphics[width=\linewidth]{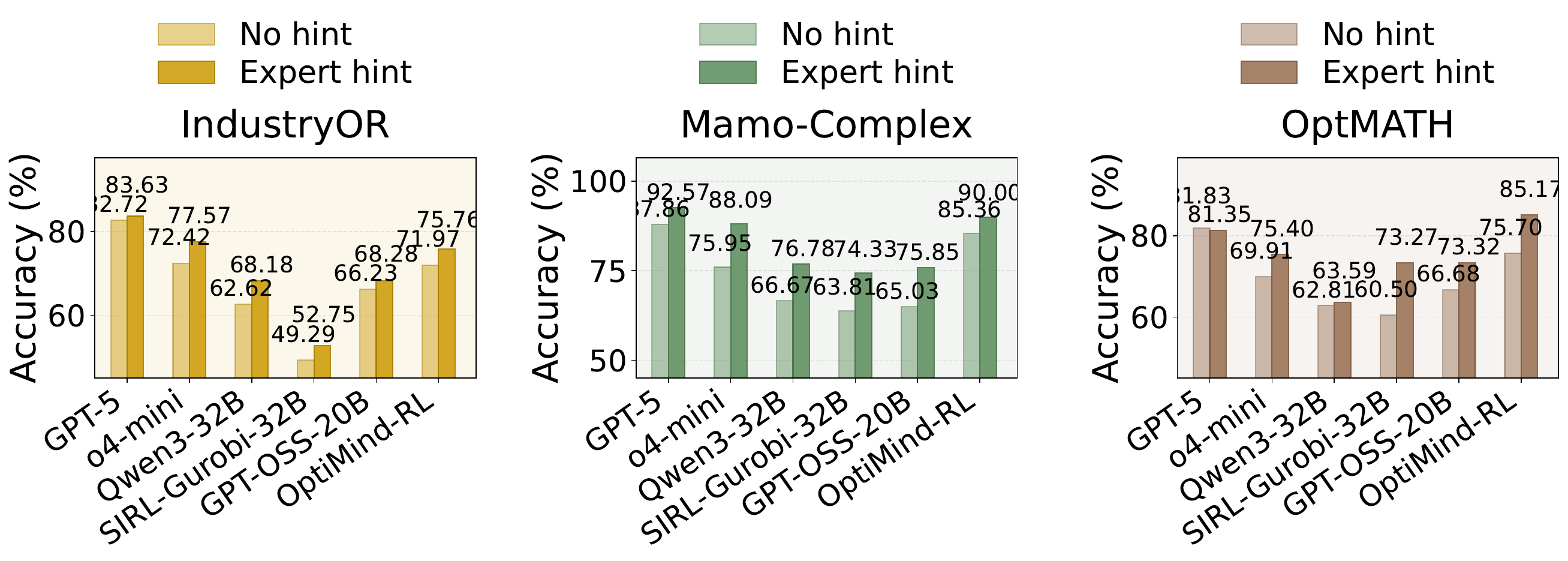}

\end{minipage}

\caption{
Left: training comparison across \textsc{gpt-oss-20b}, \textsc{OptiMind-SFT}, and \textsc{OptiMind-RL} under first-turn, no-majority-voting evaluation, with and without expert hints.
Right: first-turn accuracy of different models on three cleaned benchmarks with and without expert-curated hints, showing consistent improvements from hints.
}
\label{fig:training_and_hint_comparison}
\end{figure}

\textbf{Zooming in on OptiMind's Hint mechanisms.}
Using class-specific error analysis and the resulting hints at inference time improves single-turn accuracy on all three benchmarks.
\emph{Expert-curated} hints generalize broadly across model families: Fig.~\ref{fig:training_and_hint_comparison} (Right) reports the ``no hint'' vs.\ ``expert-curated hint'' comparison for a wide range of baselines and shows gains in nearly all settings, with only a small \textsc{GPT-5} dip on OptMATH.
Typical gains from ``no hint'' to ``expert-curated hints'' across the baselines and our RL model (single-turn, no majority voting) range from $+0.9\%$ to $+5.6\%$ on IndustryOR, $+3.3\%$ to $+12.1\%$ on Mamo-Complex, and $-0.5\%$ to $+12.8\%$ on OptMATH, where all non-\textsc{GPT-5} models improve on OptMATH.
Interestingly, even \textsc{GPT-5} improves with expert-curated hints by up to $+4.7\%$ on Mamo-Complex, suggesting that the hints encode domain-specific guidance that is complementary to general model capabilities.

\emph{Fully-automatic} and \emph{hybrid} hint sets also improve performance in some settings, though less consistently across models (see Appendix~\ref{app:additional_eval_results}). 
Because these hints were tuned for the \textsc{gpt-oss-20b} family, they are most effective for our base, SFT, and RL models. 
Expert-curated hints give larger gains on IndustryOR and Mamo-Complex, whereas automatic or hybrid hints are especially strong on OptMATH; hybrid hints often offer the best overall tradeoff, achieving best or near-best performance for \textsc{OptiMind-SFT} and the strongest reported OptMATH result for \textsc{OptiMind-RL} (Fig.~\ref{fig:different_hint_compare}).

\textbf{Effect of test-time scaling.}
We evaluate the effect of two test-time scaling techniques, multi-turn self-correction and majority voting (MV), when additional computation is available. 
Both provide consistent accuracy improvements over single-turn inference for \textsc{gpt-oss-20b} base and our SFT model, with and without hints; see 
Fig.~\ref{fig:classes_ablation} in Appendix~\ref{app:additional_eval_results}. MV and multi-turn are complementary, but their respective marginal benefit diminishes once multiple turns are used or compute increases. For \textsc{OptiMind-RL}, Table~\ref{tab:main_results_comprehensive} shows that five-turn solver-feedback correction improves over one-turn inference, with gains ranging from $+1.6\%$ to $+7.2\%$. Overall, RL training plus hints with a small number of self-correction turns delivers a strong accuracy-compute tradeoff across benchmarks.

\subsection{Additional Ablation Studies} \label{sec:ablation}

We ablate the RL reward environment, data cleaning, hint generation, classification, and inference strategies. Given time and compute limits, the detailed data-recipe and automatic-hint ablations use SFT checkpoints; for RL, we fix the cleaned data recipe and ablate the reward environment.

\textbf{Ablation on RL environment.}
We compare three RL reward environments: (1) a single-turn reward on the first generated program, (2) a terminal reward on only the final corrected program, and (3) our default multi-turn averaged reward over the solver-feedback rollout. Table~\ref{tab:rl_hint_ablation} shows that single-turn RL is strongest on OptMATH with expert and hybrid hints, while multi-turn RL remains strongest on IndustryOR and Mamo-Complex under expert hints. On the other hand, terminal reward can achieve competitive 5-turn accuracy, but it substantially weakens first-turn accuracy. We therefore use the averaged reward as our default because it preserves the benefits of multi-turn repair while also encouraging early correctness and stable behavior after solver feedback.

\textbf{Ablation on training data.}
To illustrate the effectiveness of our cleaned training data, we evaluate seven alternative data configurations: (1) \emph{Original data:} the original OptMATH set plus COPTPy-to-GurobiPy translations from \textsc{o4-mini}, keeping only instances whose final solutions match the provided answers; (2) \emph{No OR-Instruct fixing:} training on the original \textsc{OR-Instruct} questions without fixing missing-data issues; (3) \emph{No OptMATH back-translation:} training on the original \textsc{OptMATH} questions without back-translation; (4) \emph{Single-shot solution:} generating solutions without majority voting; (5) \emph{No-hint:} removing error-analysis hints when re-generating solutions; (6) \emph{Answer-aligned data:} keeping the original problem and final answer, generating 64 solutions with gpt-oss-120B, and selecting the sample that matches the original answer even if that answer is imperfect; and (7) \emph{20b data:} replacing gpt-oss-120B with gpt-oss-20B as the response generator for training trajectories.

Fig.~\ref{fig:ablation_data} compares these configurations under the first-turn, no-hint setting and reports the weighted average accuracy across the three benchmarks (weighted by sample count). The data-recipe ablation on SFT gives a weighted score of 57.9 with original data versus 81.7 with the full recipe (base: 68.2), demonstrating the effectiveness of the data-recipe. All ablations underperform our full data recipe, which combines missing-data fixes, manual resolution of key ambiguities, OptMATH re-generation with back-translation, and gpt-oss-120B-generated trajectories. Using gpt-oss-20B to generate data is competitive but consistently weaker than using gpt-oss-120B. These data-recipe ablations use SFT models; the selected cleaned data and ground-truth objective values remain fixed for RL.

\textbf{Ablation on classification.}
We also test the classification component on the Mamo-Complex benchmark, comparing five configurations: (1) classification performed by \textsc{gpt-oss} (our default); 
(2) human-expert manually assigned classes, serving as an approximate oracle;
(3) applying all class-specific hints together in the prompt without classification; (4) random assignment of classes; and (5) no hints. Table~\ref{tab:ablation_classification} shows that configuration~(1) achieves the best performance and is even slightly better than human experts  assigning classes, suggesting that the \textsc{gpt-oss}-based classifier is well aligned with the error patterns from our analysis. Against expert labels, the automatic classifier matches 200 out of 210 \textsc{Mamo-Complex} instances. Configuration~(3) performs slightly worse than (1) but still remains competitive, substantially outperforming random classification in~(4) or no hints in (5). This indicates that our class structure and hints capture meaningful error modes, and that targeted hint selection remains helpful and robust even when classification is imperfect.

\textbf{Ablation on automatic hint generation.}
We ablate our automatic-hint pipeline at three stages (Appendix~\ref{app:auto_hint_generation}): 
(i) \emph{raw {\upshape\textsc{GPT-5}} hints} (before stability refinement or validation-based filtering),
(ii) \emph{stability-refined hints} (after stability refinement and before filtering),
and (iii) \emph{stability-refined+validation-filtered hints} (our final set).
Table~\ref{tab:ablation_auto_hint} shows that refinement improves stability and validation-based filtering further boosts accuracy.
As with the data-recipe ablation, this study is run on the base and SFT models rather than retraining a separate RL checkpoint for each hint-generation variant.

\subsection{Generalization Across Model Families and Tasks}
\space

\space

\textbf{\textcolor{black}{OptiMind generalizes across model families.}}
We apply our full pipeline to Qwen3.5-9B without additional expert annotation. For multi-turn RL, we reuse the cleaned problem instances and ground-truth objective values from our GPT-OSS-20B experiments. For SFT, we follow the same trajectory-generation and filtering procedure, using Qwen3.5-122B-A10B to generate candidate solutions and retaining only those that match our ground-truth objective values in the cleaned data. We also reuse the expert-curated hint library unchanged. Table~\ref{tab:qwen_results} reports mean accuracies, with experimental details in Appendix~\ref{app:qwen_transfer}.
Qwen-SFT substantially improves first-turn accuracy. With multi-turn RL, accuracy increases by $2.83$--$16.10$ percentage points from turn one to turn five, whereas SFT changes by $-1.61$ to $+2.66$ points. At five turns, RL exceeds SFT across all three benchmarks with or without hints by $6.29$--$15.76$ points. 
\space

\textbf{\textcolor{black}{OptiMind generalizes to out-of-distribution test benchmarks.}}
\space
On the arXiv dataset, multi-turn RL has the highest observed mean accuracy without hints at both turn budgets for both model families (Appendix~\ref{app:ood_generalization}). On held-out nonlinear OptMATH, multi-turn RL exceeds SFT without hints, and two hints distilled from three separate seed instances further improve five-turn performance (Table~\ref{tab:optmath_nonlinear_hints}). We also report the standalone LogiOR evaluation in Appendix~\ref{app:standalone_logior}.

Finally, we evaluate \textsc{OptiMind-RL} on agentic tool-calling tasks~\citep{liu2024apigen,liu2024toolacewinningpointsllm} and general-language benchmarks using the LM Evaluation Harness~\citep{eval-harness} to assess performance beyond optimization formulation. We observe that \textsc{OptiMind-RL} largely preserves GPT-OSS-20B's performance on the evaluated agentic tool-calling and LM-eval suites, whereas \textsc{OptiMind-SFT} shows degradation, including incorrect tool-calling syntax in the agentic evaluation (Appendix~\ref{app:general_capabilities}).

\section{Related Work}

A growing body of work studies natural-language-to-optimization formulation benchmarks. Early datasets such as NL4LP~\citep{pmlr-v220-ramamonjison23a} focused on LPs, while later benchmarks expanded to ILP/MILP, industrial, logistics, nonlinear, and long-context settings, including Mamo~\citep{huang2024mamo}, NLP4LP~\citep{ahmaditeshnizi2024optimus}, ComplexOR~\citep{xiao2024chainofexperts}, IndustryOR~\citep{tang2024orlm}, OptiBench~\citep{yang2025optibench}, OptMATH~\citep{lu2025optmath}, and LogiOR~\citep{logior}. These benchmarks enable systematic evaluation of LLM-based optimization modeling, but recent analyses identify missing data, ambiguity, infeasible instances, and incorrect answers~\citep{xiao2025survey,survey}. Parallel efforts therefore release cleaned benchmarks, including SIRL~\citep{chen2025solver} and LogiOR~\citep{logior}. Our pipeline uses class-level error analysis and hints to repair benchmarks, regenerate training data, and support SFT, RL reward computation, and inference.

Recent work also uses solver feedback, retrieval, and agentic prompting for optimization modeling. LLMOPT~\citep{jiang2025llmopt}, AlphaOPT~\citep{alphaopt}, and OptiTree~\citep{liu2025optitree} mainly use these mechanisms as inference-time scaffolds: generate a formulation, execute or check it, and prompt or guide repair. Other methods incorporate feedback during training: SIRL~\citep{chen2025solver} uses solver-informed verifiable rewards, StepORLM~\citep{steporlm} uses generative process supervision for credit assignment, and related RL approaches exploit solver execution, feasibility, objective values, or process-level signals~\citep{orr1,storm,evom,motwani2026autoor}. Another related line treats solver interaction as a multi-step decision process~\citep{solver_in_the_loop}, but focuses on debugging existing or sabotaged OR models rather than generating formulations from natural language. 
The CALM training pipeline~\citep{storm} also uses repeated code execution during RL; OptiMind additionally uses reusable class-specific hints for training-data curation and inference within a validated solver-grounded environment.
Appendix~\ref{app:structured_comparison} summarizes these distinctions against closely related optimization-modeling systems.

\section{Conclusion}

We present \textsc{OptiMind}, a framework for teaching small language models optimization formulation through expert-cleaned data, class-specific hints, and multi-turn RL with solver feedback. \textsc{OptiMind} turns one-shot code generation into an interactive formulate--execute--revise process. On expert-cleaned IndustryOR, Mamo-Complex, and OptMATH, a specialized 20B model surpasses strong open-source baselines and reaches frontier-model performance. The full hint library, cleaned test data, cleaning logs, and evaluation code are publicly available on \optimindcode, and our trained OptiMind-RL model is available on \optimindmodel.

\begin{ack}
The authors thank Luke Marshall and Patricia Kovaleski for insightful discussions. The authors further thank the anonymous reviewers for their valuable feedback. The first author would like to acknowledge support from NSF CAREER Award and Packard Fellowships during the work.
\end{ack}

\FloatBarrier
\flushbottom
\setlength{\intextsep}{\savedintextsep}
\setlength{\textfloatsep}{\savedtextfloatsep}
\setlength{\dbltextfloatsep}{\saveddbltextfloatsep}
\setlength{\parskip}{\savedmainparskip}
\bibliographystyle{plainnat}
\bibliography{ref,applicationsIP}

\clearpage
\appendix

\section{\texorpdfstring{Structured Comparison to Prior Work}{Structured Comparison to Prior Work}}\label{app:structured_comparison}

Table~\ref{tab:prior_work_comparison} summarizes the main technical distinctions between OptiMind and closely related optimization-modeling systems. The key difference is that OptiMind uses one reusable class-level hint library to connect benchmark cleaning, training-data regeneration, SFT validation, multi-turn RL, and inference-time guidance.

\begin{table}[H]
\caption{Structured comparison with related optimization-modeling systems.}
\label{tab:prior_work_comparison}
\centering
\footnotesize
\setlength{\tabcolsep}{3pt}
\renewcommand{\arraystretch}{0.95}
\begin{tabular}{@{}>{\raggedright\arraybackslash}p{2.7cm} >{\raggedright\arraybackslash}p{5.5cm} >{\raggedright\arraybackslash}p{1.5cm} >{\raggedright\arraybackslash}p{3.4cm}@{}}
\toprule
\textbf{Method} & \textbf{Core methodology} & \textbf{Policy optimization} & \textbf{Inference control} \\
\midrule
ORLM~\citep{tang2024orlm} & OR-Instruct synthetic data and SFT & -- & -- \\
OptMATH~\citep{lu2025optmath} & Bidirectional data synthesis for optimization modeling & -- & -- \\
LLMOPT~\citep{jiang2025llmopt} & Multi-instruction tuning and self-correction & -- & Self-correction \\
SIRL~\citep{chen2025solver} & Solver-informed RL with executable code/model verification & RLVR & Instance-enhanced self-consistency \\
StepORLM~\citep{steporlm} & Solver outcome verification plus generative process supervision & W-DPO & GenPRM verification/reranking \\
Solver-in-the-Loop~\citep{solver_in_the_loop} & MDP-based OR model debugging with IIS/solver actions & GRPO & MDP repair policy \\
CALM~\citep{storm} & \leavevmode Flaw-specific hint intervention for trajectory curation, followed by SFT and GRPO & GRPO & Code-interpreter reflection \\
OptiTree~\citep{liu2025optitree} & Hierarchical thought-tree search for formulation & -- & Tree-search decomposition \\
AlphaOPT~\citep{alphaopt} & Self-improving experience library from failed attempts & -- & Library retrieval and refinement \\
\textbf{Ours} & \leavevmode Hint distillation, data improvement, independently trained SFT/GRPO variants, and solver-guided inference & GRPO & Hint injection and self-correction \\
\bottomrule
\end{tabular}

\medskip
\begin{tabular}{@{}>{\raggedright\arraybackslash}p{2.7cm} *{2}{>{\centering\arraybackslash}p{1.3cm}} >{\centering\arraybackslash}p{1.4cm} >{\centering\arraybackslash}p{1.5cm} >{\centering\arraybackslash}p{1.8cm} >{\centering\arraybackslash}p{1.5cm}@{}}
\toprule
\textbf{Method} & \textbf{Finetuning} & \textbf{Improves data} & \textbf{External memory} & \textbf{Solver/tool feedback} & \textbf{Multi-turn train-time rollout} & \textbf{Class-level hints} \\
\midrule
ORLM & \checkmark & \checkmark & -- & -- & -- & -- \\
OptMATH & \checkmark & \checkmark & -- & \checkmark\ (data filtering) & -- & -- \\
LLMOPT & \checkmark & -- & -- & \checkmark & -- & -- \\
SIRL & \checkmark & \checkmark & -- & \checkmark & -- & -- \\
StepORLM & \checkmark & \checkmark & -- & \checkmark & -- & -- \\
Solver-in-the-Loop & \checkmark & \checkmark\ (debug data) & -- & \checkmark & \checkmark\ (debugging) & -- \\
CALM & \checkmark & \checkmark & -- & \checkmark & \checkmark & -- \\
OptiTree & -- & -- & -- & -- & -- & -- \\
AlphaOPT & -- & -- & \checkmark & \checkmark & -- & -- \\
\textbf{Ours} & \checkmark & \checkmark & \checkmark & \checkmark & \checkmark & \checkmark \\
\bottomrule
\end{tabular}
\end{table}

\clearpage
\section{Additional Evaluation Results} \label{app:additional_eval_results}

\subsection{Comprehensive Expert-Cleaned Benchmark Results}\label{app:benchmark_headroom}

Table~\ref{tab:main_results_comprehensive} gives the full single-sample results behind the condensed main comparison, including standard deviations and automatic-hint variants.
OptiMind-RL uses expert hints and five turns in the main comparison. We also report automatic and hybrid hints: in particular, OptMATH accuracy is 90.99 with expert hints and 91.46 with hybrid hints.

\begin{table}[!ht]
\caption{Comprehensive results on expert-cleaned benchmarks in the single-sample setting (majority $K=1$) under different hint configurations. For the \textsc{gpt-oss-20b}-and-below block, the best result in each metric column is bolded.}
\label{tab:main_results_comprehensive}
\centering
\scriptsize
\setlength{\tabcolsep}{3pt}
\renewcommand{\arraystretch}{1.05}
\resizebox{\linewidth}{!}{\begin{tabular}{llcccccc}
\toprule
& & \multicolumn{2}{c}{\textbf{IndustryOR}} & \multicolumn{2}{c}{\textbf{Mamo-Complex}} & \multicolumn{2}{c}{\textbf{OptMATH}} \\
\cmidrule(lr){3-4}\cmidrule(lr){5-6}\cmidrule(lr){7-8}
\textbf{Model} & \textbf{Hints} & \textbf{1 turn} & \textbf{5 turns} & \textbf{1 turn} & \textbf{5 turns} & \textbf{1 turn} & \textbf{5 turns} \\
\midrule
\textsc{gpt-5} & no hint
& $82.72 \pm 2.15$ & $83.23 \pm 2.29$
& $87.86 \pm 1.55$ & $89.36 \pm 1.92$
& $81.83 \pm 2.35$ & $89.93 \pm 2.12$ \\
& expert
& $83.63 \pm 2.17$ & $85.15 \pm 1.65$
& $92.57 \pm 1.19$ & $93.42 \pm 0.62$
& $81.35 \pm 2.92$ & $90.41 \pm 1.85$ \\
& automatic
& $65.65 \pm 2.71$ & $77.78 \pm 1.00$
& $86.57 \pm 1.28$ & $91.42 \pm 1.51$
& $77.85 \pm 2.64$ & $91.18 \pm 2.37$ \\
& hybrid
& $77.52 \pm 4.31$ & $83.58 \pm 1.51$
& $78.33 \pm 1.01$ & $91.90 \pm 0.67$
& $79.73 \pm 2.34$ & $91.18 \pm 2.22$ \\
\midrule
\textsc{o4-mini} & no hint
& $72.42 \pm 2.33$ & $78.58 \pm 2.72$
& $75.95 \pm 2.38$ & $87.78 \pm 1.15$
& $69.91 \pm 3.20$ & $82.64 \pm 3.98$ \\
& expert
& $77.57 \pm 2.55$ & $80.30 \pm 1.97$
& $88.09 \pm 2.02$ & $91.67 \pm 1.52$
& $75.40 \pm 2.73$ & $86.10 \pm 2.69$ \\
& automatic
& $71.04 \pm 2.33$ & $79.12 \pm 2.54$
& $78.57 \pm 3.33$ & $86.03 \pm 0.72$
& $73.68 \pm 2.78$ & $87.18 \pm 2.50$ \\
& hybrid
& $76.76 \pm 5.77$ & $79.29 \pm 1.30$
& $83.33 \pm 4.20$ & $89.16 \pm 1.57$
& $75.87 \pm 2.10$ & $86.55 \pm 2.46$ \\
\midrule
\textsc{Qwen3-32B} & no hint
& $62.62 \pm 2.73$ & $66.16 \pm 1.74$
& $66.67 \pm 2.67$ & $72.47 \pm 2.65$
& $62.81 \pm 1.72$ & $66.82 \pm 4.28$ \\
& expert
& $68.18 \pm 2.67$ & $72.47 \pm 2.65$
& $76.78 \pm 1.05$ & $84.05 \pm 1.48$
& $63.59 \pm 1.53$ & $77.51 \pm 1.72$ \\
& automatic
& $62.92 \pm 2.93$ & $66.56 \pm 3.28$
& $68.85 \pm 2.66$ & $77.57 \pm 2.14$
& $62.73 \pm 4.56$ & $75.04 \pm 4.26$ \\
& hybrid
& $64.34 \pm 3.23$ & $68.98 \pm 3.12$
& $73.33 \pm 2.68$ & $81.23 \pm 2.59$
& $66.73 \pm 3.33$ & $77.78 \pm 3.70$ \\
\midrule
\textsc{SIRL-Gurobi32B} & no hint
& $49.29 \pm 3.39$ & $51.81 \pm 2.61$
& $63.81 \pm 1.53$ & $74.44 \pm 1.66$
& $60.50 \pm 1.61$ & $68.37 \pm 1.70$ \\
& expert
& $52.75 \pm 2.18$ & $55.44 \pm 2.04$
& $74.33 \pm 1.65$ & $77.72 \pm 1.57$
& $73.27 \pm 0.80$ & $75.38 \pm 1.34$ \\
& automatic
& $50.30 \pm 2.06$ & $51.81 \pm 2.23$
& $75.47 \pm 1.54$ & $78.09 \pm 2.12$
& $71.21 \pm 1.28$ & $74.70 \pm 0.87$ \\
& hybrid
& $50.70 \pm 2.18$ & $53.13 \pm 2.82$
& $74.61 \pm 1.00$ & $78.90 \pm 1.45$
& $73.90 \pm 1.44$ & $75.58 \pm 1.18$ \\
\hline\hline
\textsc{gpt-oss-20b} & no hint
& $66.23 \pm 3.77$ & $71.89 \pm 3.05$
& $65.03 \pm 3.91$ & $86.80 \pm 1.33$
& $66.68 \pm 4.70$ & $81.52 \pm 3.60$ \\
& expert
& $68.28 \pm 2.61$ & $75.05 \pm 1.84$
& $75.85 \pm 2.74$ & $89.47 \pm 1.27$
& $73.32 \pm 4.05$ & $87.39 \pm 1.91$ \\
& automatic
& $61.31 \pm 2.65$ & $73.43 \pm 2.07$
& $68.14 \pm 1.91$ & $85.38 \pm 1.43$
& $73.38 \pm 3.82$ & $88.83 \pm 2.46$ \\
& hybrid
& $65.25 \pm 3.96$ & $75.05 \pm 1.65$
& $74.10 \pm 2.71$ & $87.57 \pm 1.15$
& $74.65 \pm 4.57$ & $87.41 \pm 3.14$ \\
\midrule
\textsc{OptiMind-SFT} & no hint
& $68.97 \pm 2.15$ & $72.43 \pm 3.43$
& $85.77 \pm 2.19$ & $90.71 \pm 1.24$
& $72.95 \pm 3.05$ & $81.64 \pm 3.42$ \\
& expert
& $70.50 \pm 3.29$ & $75.25 \pm 2.57$
& $89.09 \pm 1.27$ & $92.04 \pm 1.05$
& $79.93 \pm 2.24$ & $88.50 \pm 1.81$ \\
& automatic
& $69.39 \pm 3.08$ & $74.44 \pm 2.97$
& $82.76 \pm 2.88$ & $88.71 \pm 1.38$
& $83.75 \pm 2.53$ & $90.69 \pm 1.48$ \\
& hybrid
& $70.70 \pm 2.77$ & $74.04 \pm 2.97$
& $87.79 \pm 2.05$ & $91.01 \pm 0.96$
& $84.27 \pm 2.68$ & $90.08 \pm 1.59$ \\
\midrule
\textsc{OptiMind-RL} & no hint
& $71.97 \pm 2.36$ & $75.88 \pm 2.05$
& $85.36 \pm 1.34$ & $89.58 \pm 1.13$
& $75.70 \pm 2.57$ & $82.92 \pm 3.07$ \\
& expert
& $\bf{75.76 \pm 2.14}$ & $\bf{78.54 \pm 2.07}$
& $\bf{90.00 \pm 1.73}$ & $\bf{92.26 \pm 1.23}$
& $85.17 \pm 2.05$ & $90.99 \pm 0.93$ \\
& hybrid
& $74.24 \pm 0.87$ & $76.52 \pm 1.49$
& $88.15 \pm 1.08$ & $89.70 \pm 0.98$
& $\bf{87.58 \pm 0.31}$ & $\bf{91.46 \pm 1.23}$ \\
\end{tabular}}
\end{table}

\FloatBarrier
\begin{figure}[H]
\centering
\includegraphics[width=0.75\linewidth]{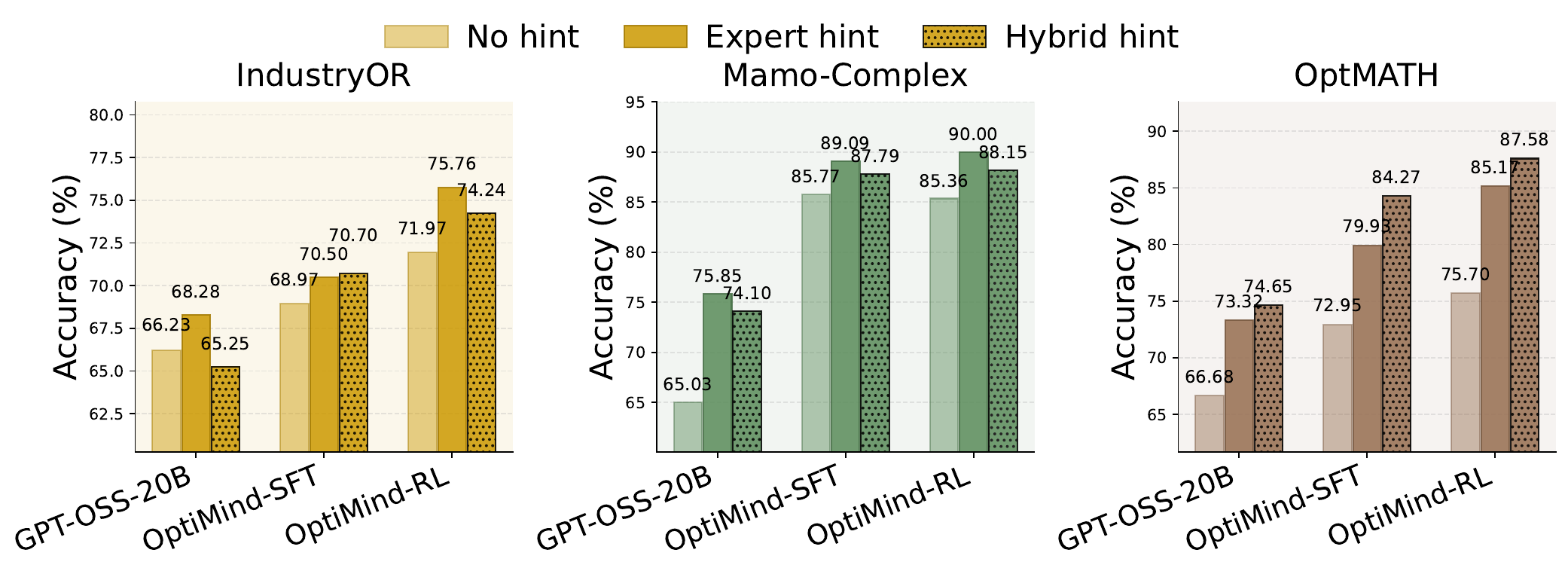}

\caption{First-turn accuracy of \textsc{gpt-oss-20b}, \textsc{OptiMind-SFT}, and \textsc{OptiMind-RL} under different hint configurations on the expert-cleaned benchmarks.}
\label{fig:different_hint_compare}
\end{figure}

\paragraph{Benchmark headroom and RL gains.}
The size of the gains from RL is consistent with differences in the remaining benchmark headroom (Table~\ref{tab:main_results_comprehensive}). IndustryOR has more room for improvement and shows the clearest improvement by RL (75.25 for SFT versus 78.54 for RL, both with five turns and expert hints), although GPT-5 with the same expert-hint setting remains stronger at 85.15. On Mamo-Complex and OptMATH, strong systems cluster within a narrow high-accuracy range once expert hints and iterative feedback are used: on Mamo-Complex, five-turn SFT and RL reach 92.04 and 92.26 accuracy, close to 93.42 for five-turn GPT-5 with hints, with overlapping one-standard-deviation intervals; on OptMATH, five-turn SFT and RL reach 88.50 and 90.99 accuracy, with RL attaining a slightly higher mean accuracy than the 90.41 of five-turn GPT-5 under the same expert-hint setting.

\Needspace{6\baselineskip}
\subsection{Per-class Breakdown of Accuracy}

\begin{figure}[tbp]\begin{center}
\includegraphics[width=0.9\linewidth]{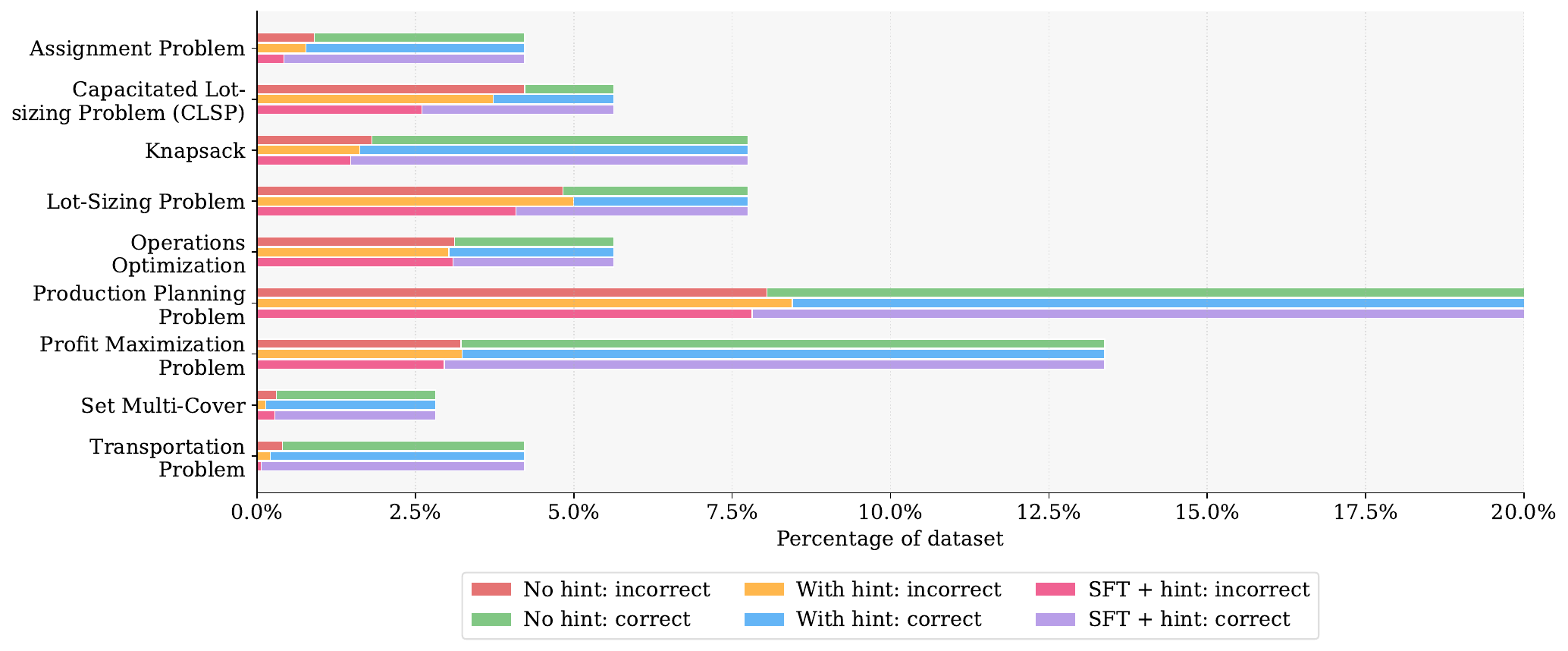}

\includegraphics[width=0.8\linewidth]{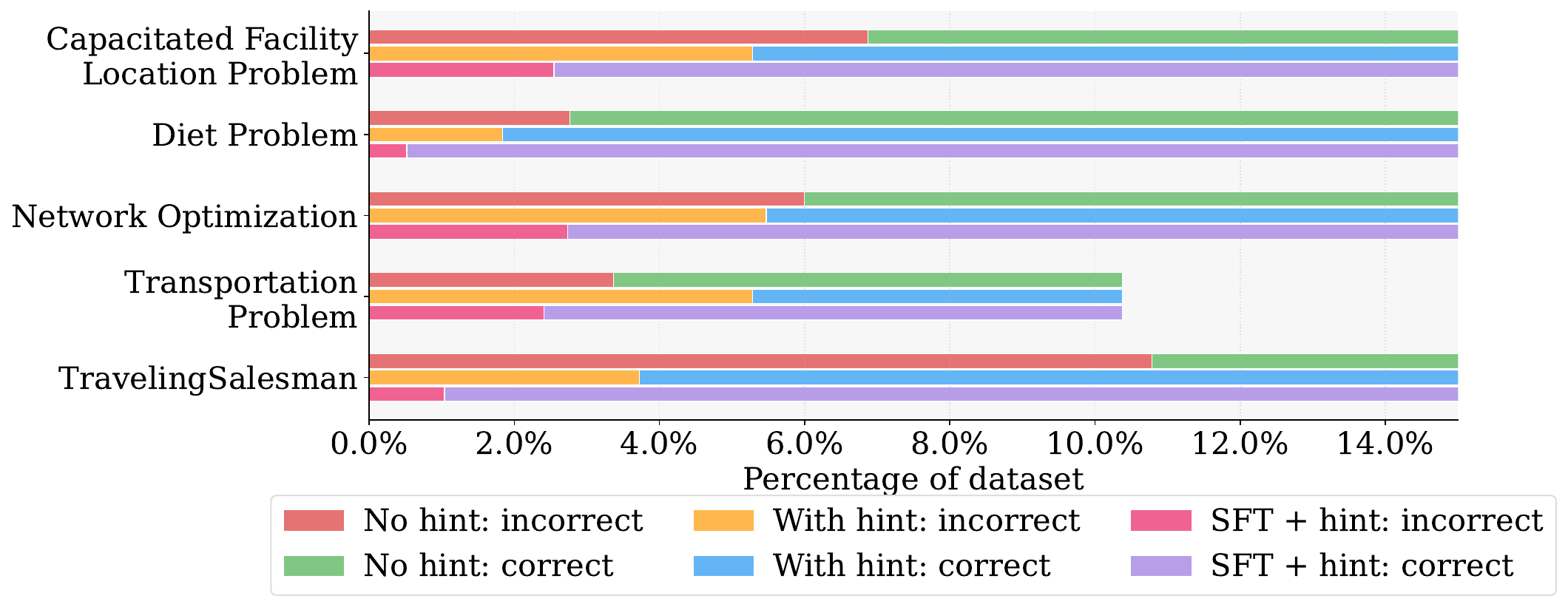}

\includegraphics[width=0.9\linewidth]{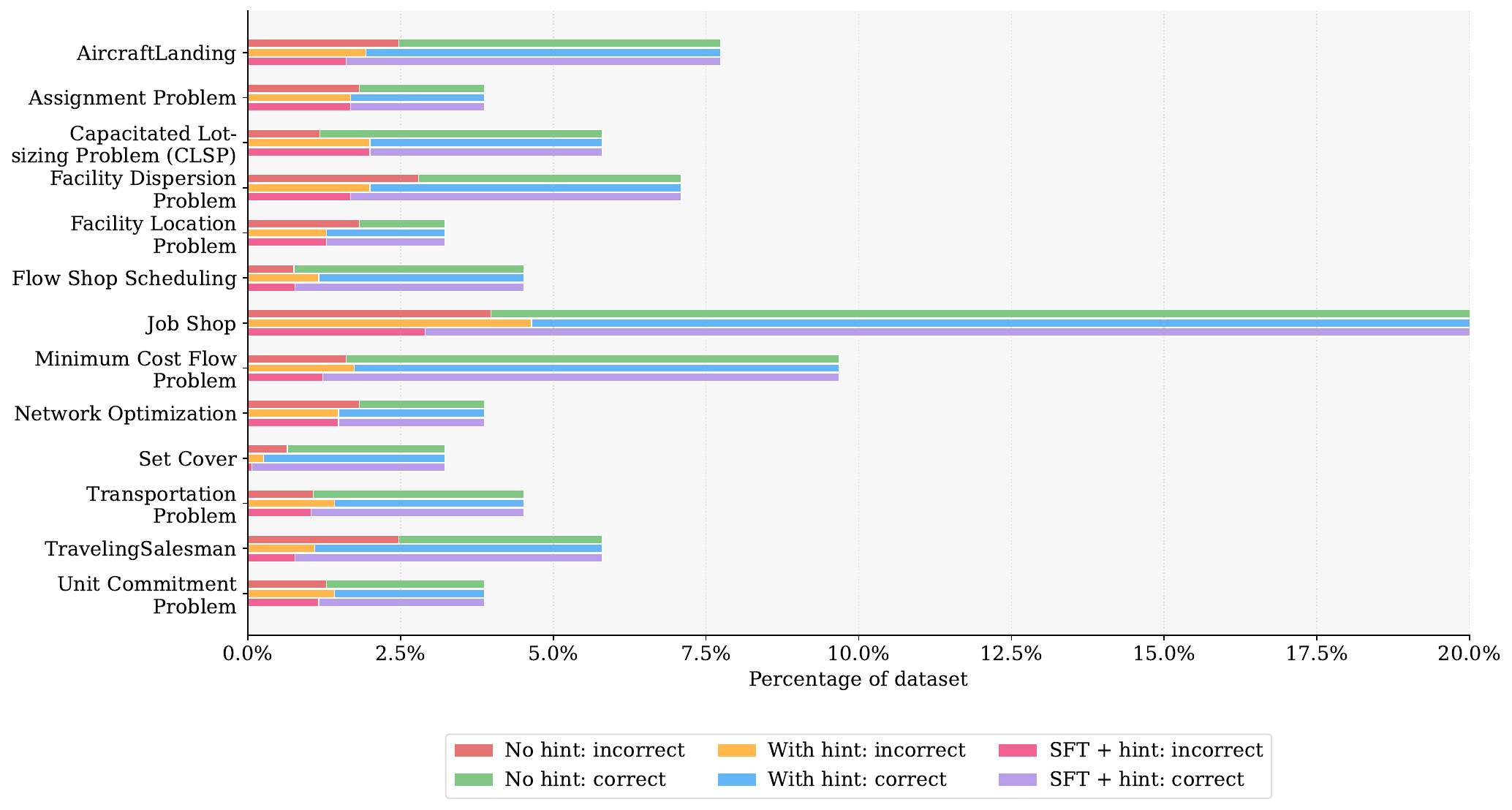}

\end{center}
\caption{Accuracy by problem type on IndustryOR (top), Mamo-Complex (middle) and OptMATH (bottom), measured via first-turn evaluation with gpt-oss-20B (no majority voting). Problem types occurring in less than 2.5\% of instances are omitted in the figure. We are using hybrid hints.}
\label{fig:breakdown_more}
\end{figure}

\space Figure~\ref{fig:breakdown_more} presents the per-problem-class effect of error-analysis hints at inference time for \textsc{gpt-oss-20b} across all three cleaned benchmarks: IndustryOR, Mamo-Complex, and OptMATH.

To illustrate the pattern more concretely, we use Mamo-Complex as a representative example because its instances are concentrated in five dominant problem classes. 
The gains are broad across classes, with the largest improvements appearing in classes where errors often arise from sign conventions, such as flow-conservation constraints, or from structurally complex constraints, such as subtour elimination in TSP. 
\space The SFT+hint setting further improves over base+hint across these major classes. 
Although IndustryOR and OptMATH have more heterogeneous class distributions, they show the same overall trend: error-analysis hints improve performance across a wide range of problem classes.

\subsection{Results on Parallel-Cleaned Benchmarks}

\begin{figure}
    \centering
      \includegraphics[width=0.7\linewidth]{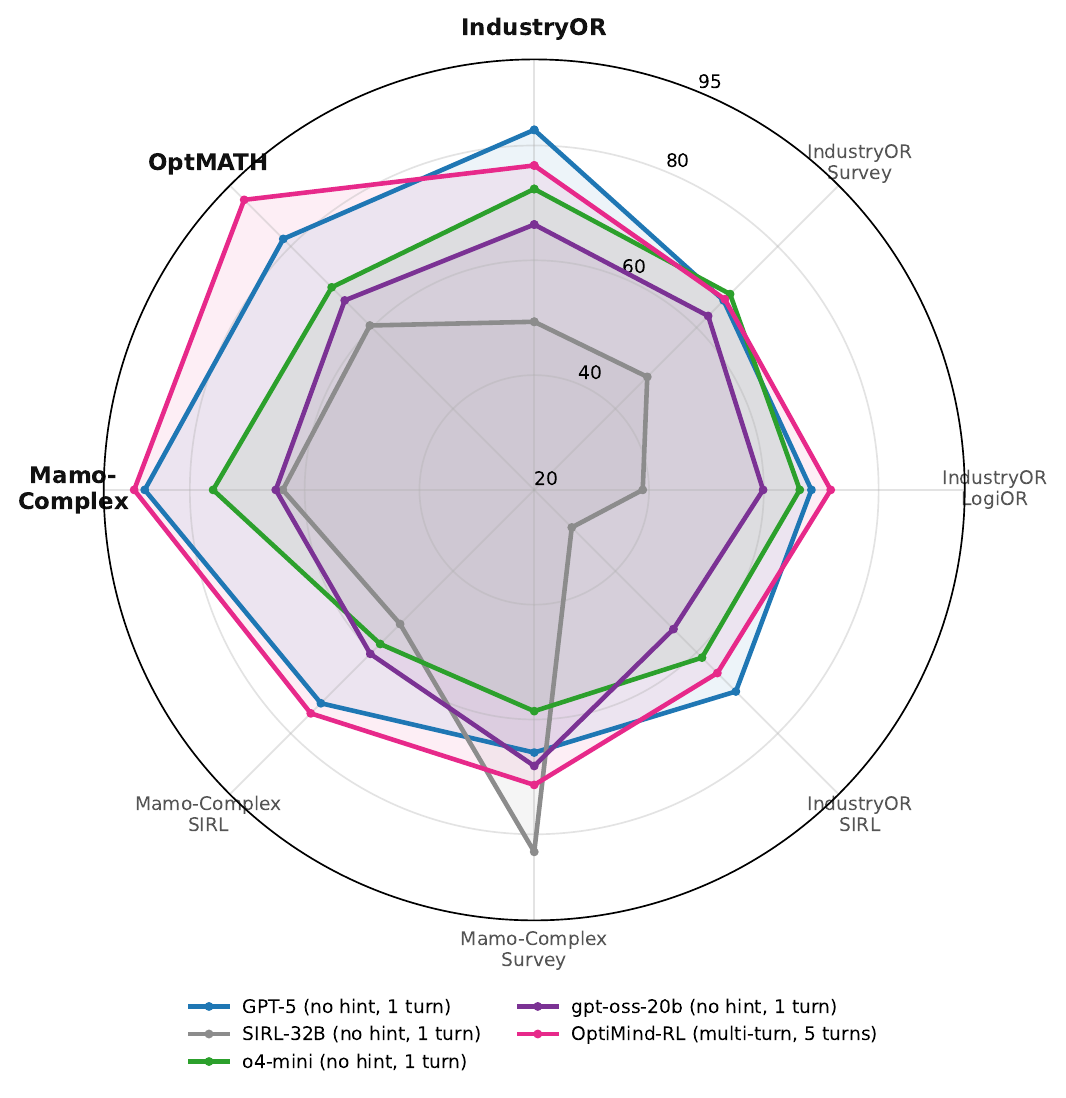}

  \caption{ Mean accuracy across expert-cleaned and independently cleaned benchmark versions. GPT-5, SIRL-Gurobi32B, o4-mini, and GPT-OSS-20B use no hints and one turn; OptiMind-RL uses expert hints and five turns. All settings use $K=1$. Values are drawn from Tables~\ref{tab:main_results_comprehensive} and~\ref{tab:other_benchmarks}.}
  \label{fig:benchmarks_radar}
\end{figure}

We evaluate on versions of IndustryOR and Mamo-Complex independently cleaned by SIRL, LogiOR, and Survey (Table~\ref{tab:other_benchmarks}; Figure~\ref{fig:benchmarks_radar}). OptiMind-RL has higher mean accuracy than SFT on four of the five independently cleaned benchmarks at both turn budgets, including both Mamo-Complex variants. Across Tables~\ref{tab:main_results_comprehensive} and~\ref{tab:other_benchmarks}, RL is higher on seven of the eight expert-hint benchmark variants; the only exception is IndustryOR-LogiOR, where SFT leads by $0.72$ points at one turn and $1.35$ points at five turns.

\begin{table}[!ht]
\caption{Accuracies under different base models and hint settings on benchmarks cleaned by parallel works. SIRL stands for \cite{chen2025solver}, LogiOR stands for \cite{logior}, and Survey stands for \cite{survey}. Best reported result in each metric column is bolded. All values are mean accuracy (\%). All comparisons use $K=1$.}
\label{tab:other_benchmarks}
\centering
\resizebox{\linewidth}{!}{\begin{tabular}{ll cc cc cc cc cc}
\toprule
& & \multicolumn{2}{c}{\textbf{IndustryOR-SIRL}} & \multicolumn{2}{c}{\textbf{IndustryOR-LogiOR}} & \multicolumn{2}{c}{\textbf{IndustryOR-Survey}} & \multicolumn{2}{c}{\textbf{Mamo-Complex-SIRL}} & \multicolumn{2}{c}{\textbf{Mamo-Complex-Survey}} \\
\cmidrule(lr){3-4}\cmidrule(lr){5-6}\cmidrule(lr){7-8}\cmidrule(lr){9-10}\cmidrule(lr){11-12}
\textbf{Model} & \textbf{Hints} & \textbf{1 turn} & \textbf{5 turns} & \textbf{1 turn} & \textbf{5 turns} & \textbf{1 turn} & \textbf{5 turns} & \textbf{1 turn} & \textbf{5 turns} & \textbf{1 turn} & \textbf{5 turns} \\
\midrule
\textsc{gpt-5} & no & 69.67 & 71 & 68.29 & 70.12 & 66.67 & 68.25 & 72.57 & 74.05 & 65.76 & 66.21 \\
 & expert & \textbf{72} & \textbf{73} & \textbf{76.82} & \textbf{78.04} & 70.23 & 69.04 & \textbf{88.17} & \textbf{89.65} & 85.58 & 86.48 \\
\midrule
\textsc{o4-mini} & no & 61.33 & 64.67 & 66.26 & 74.79 & 68.25 & 69.84 & 57.96 & 69.95 & 58.55 & 66.67 \\
 & expert & 68.67 & 70 & 71.54 & 74.39 & 66.67 & 63.49 & 81.77 & 86.53 & 78.82 & 86.48 \\
\midrule
\textsc{SIRL-Gurobi32B} & no & 29.25 & 29.87 & 38.9 & 39 & 47.85 & 47.85 & 53.05 & 53.1 & 83.06 & 82.97 \\
 & expert & 30.8 & 31.2 & 35.1 & 35.6 & 46.67 & 46.67 & 65.56 & 65.61 & 77.02 & 77.2 \\
\midrule
\textsc{gpt-oss-20b} & no & 54.3 & 62.7 & 59.87 & 65.97 & 62.85 & 67.38 & 60.39 & 77.83 & 68.1 & 78.82 \\
 & expert & 57.3 & 63.3 & 66.09 & 70 & 62.38 & 67.38 & 73.59 & 85.27 & 77.47 & 85.85 \\
\midrule
\textsc{OptiMind-SFT} & no & 57.3 & 63.7 & 63.71 & 67.68 & 67.61 & 68.57 & 79.8 & 85.32 & 85.22 & 85.94 \\
 & expert & 59.2 & 62.6 & 67.68 & 71.64 & 69.04 & 71.42 & 84.43 & 87.68 & 87.38 & 90.09 \\
\midrule
\textsc{OptiMind-RL} & no & 61.33 & 65.11 & 69.65 & 72.63 & 67.99 & 68.25 & 72.86 & 74.04 & 72.25 & 75.05 \\
 & expert & 61.82 & 66.55 & 66.96 & 70.29 & \textbf{71.65} & \textbf{71.65} & 86.16 & 89.16 & \textbf{90.09} & \textbf{92.34} \\
\bottomrule
\end{tabular}}

\end{table}

\begin{figure}
\centering
\begin{subfigure}[t]{\linewidth}
\centering
\resizebox{\linewidth}{!}{\begin{tabular}{l c l cc cc cc}
\toprule
& & &
\multicolumn{2}{c}{\bf IndustryOR} &
\multicolumn{2}{c}{\bf Mamo-Complex} &
\multicolumn{2}{c}{\bf OptMATH} \\
\cmidrule(r){4-5} \cmidrule(r){6-7} \cmidrule(r){8-9}
\multicolumn{1}{l}{\bf Model} &
\multicolumn{1}{l}{\bf K} &
\multicolumn{1}{l}{\bf Hints} &
\multicolumn{1}{c}{\bf 1 turn} & \multicolumn{1}{c}{\bf 5 turns} &
\multicolumn{1}{c}{\bf 1 turn} & \multicolumn{1}{c}{\bf 5 turns} &
\multicolumn{1}{c}{\bf 1 turn} & \multicolumn{1}{c}{\bf 5 turns} \\
\midrule

\textsc{gpt-oss-20b} & 1 & no hint
& $66.23 \pm 3.77$ & $71.89 \pm 3.05$
& $65.03 \pm 3.91$ & $86.80 \pm 1.33$
& $74.87 \pm 3.40$ & $86.45 \pm 2.13$ \\
& & hybrid
& $65.25 \pm 3.96$ & $75.05 \pm 1.65$ 
& $74.10 \pm 2.71$ & $87.57 \pm 1.15$
& $78.61 \pm 3.66$ & $87.59 \pm 2.78$ \\
\midrule

\textsc{gpt-oss-20b} & 8 & no hint
& $75.85 \pm 2.04$ & $77.47 \pm 1.26$ 
& $89.71 \pm 1.12$ & $89.66 \pm 0.89$ 
& $90.31 \pm 1.17$ & $90.39 \pm 1.64$ \\
& & hybrid
& $78.28 \pm 0.98$ & $78.38 \pm 1.36$ 
& $89.95 \pm 1.21$ & $89.71 \pm 1.27$
& $91.64 \pm 0.97$ & $90.23 \pm 1.12$ \\

\midrule\midrule

\textsc{OptiMind-SFT} & 1 & no 
& $68.97 \pm 2.15$ & $72.43 \pm 3.43$ 
& $85.77 \pm 2.19$ & $90.71 \pm 1.24$ 
& $84.82 \pm 2.95$ & $89.62 \pm 2.80$ \\
& & hybrid
& $\bf{70.70 \pm 2.77}$ & $\bf{74.04 \pm 2.97}$ 
& $\bf{87.79 \pm 2.05}$ & $\bf{91.01 \pm 0.96}$ 
& $\bf{87.76\pm 2.36}$ & $\bf{90.33 \pm 1.31}$ \\

\midrule
\textsc{OptiMind-SFT} & 8 & no 
& $78.48 \pm 1.43$ & $79.09 \pm 1.58$ 
& $92.95 \pm 0.58$ & $\bf{93.09 \pm 0.66}$
& $\bf{93.55 \pm 0.69}$ & $\bf{93.94 \pm 1.00}$  \\
& & hybrid
& $\bf{79.58 \pm 0.92}$ & $\bf{79.69 \pm 0.88}$ 
& $\bf{93.00 \pm 0.73}$ & $93.00 \pm 0.67$
& $90.62 \pm 0.90$ & $91.09 \pm 0.75$ \\
\bottomrule
\end{tabular}}\end{subfigure}
\hfill
\vspace{4.0em}
\begin{subfigure}[t]{0.99\linewidth}
\centering
\includegraphics[width=0.99\linewidth]{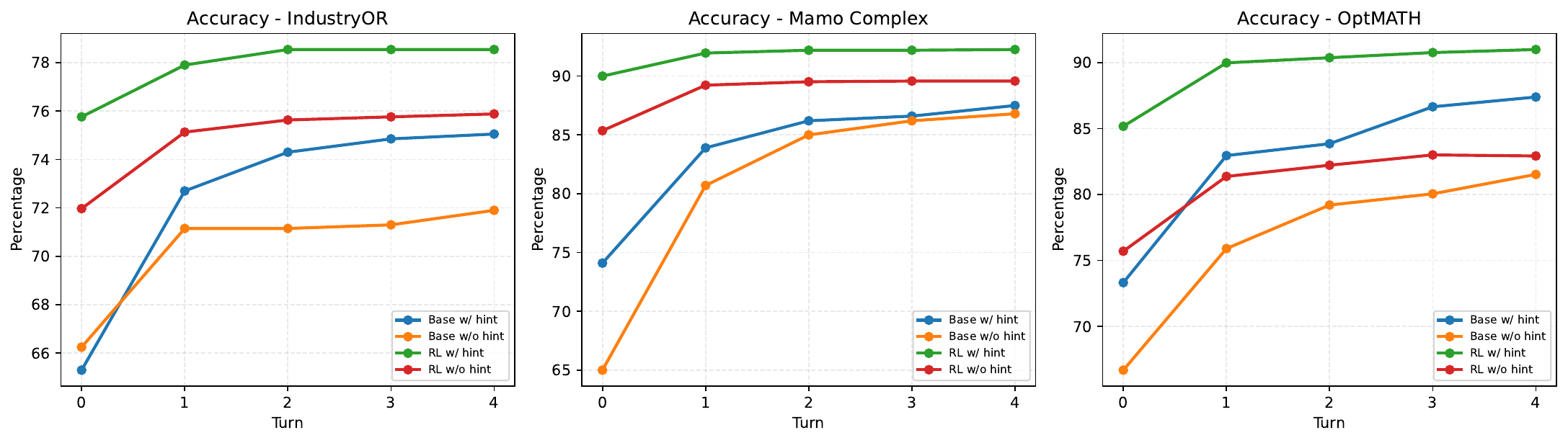}

\end{subfigure}
\caption{\textbf{Top}: Accuracies of the \textsc{gpt-oss-20b} base model and \textsc{OptiMind-SFT} across majority voting $K$ and turn counts on the three benchmarks. \textbf{Bottom}: Accuracy  of \textsc{gpt-oss-20b} base model and \textsc{OptiMind-RL} over five self-correcting turns on the three benchmarks. All evaluations utilize hybrid hints.}
 \label{fig:classes_ablation}
\end{figure}

\space

\FloatBarrier
\space
\Needspace{6\baselineskip}
\subsection{Transfer to Qwen3.5-9B}
\label{app:qwen_transfer}

To test whether the full pipeline is tied to GPT-OSS-20B, we applied it to Qwen3.5-9B without model-specific expert re-annotation. For SFT, we distill solution trajectories from Qwen3.5-122B-A10B and retain only those whose answers match the ground truth on the OptiMind training instances. Multi-turn RL uses the same cleaned problem statements and ground-truth objective values as our main OptiMind-RL model, without new teacher-generated trajectories. The same expert hints are reused without regenerating or adapting them. Mean accuracy (\%) and sample standard deviation are reported in Table~\ref{tab:qwen_results}.

The Qwen evaluation uses 99 IndustryOR, 210 Mamo-Complex, and 128 linear OptMATH instances. The corresponding CSV contains 130 rows, but two have empty question and answer fields (zero-based rows 67 and 76) and are excluded from evaluation, leaving 128 valid instances.

Each Table~\ref{tab:qwen_results} configuration uses five inference runs (seeds 0--4), temperature 0.6, top-$p=0.95$, and $K=1$, with all five turns executed without early stopping. One-turn and five-turn accuracy refer to the first and fifth responses in the same trajectories, rather than success at any preceding turn. Standard deviations use the $n-1$ denominator across runs at a fixed checkpoint and benchmark; they do not measure variability across training seeds. Qwen-RL-multi denotes checkpoint v3 (\texttt{iter\_0000399}) throughout Tables~\ref{tab:qwen_results}--\ref{tab:standalone_logior}. In Table~\ref{tab:qwen_results}, the base and SFT models use a dynamic total token budget of 60,000, while RL uses 131,072 with a per-turn completion cap of 24,576.

\begin{table}[htbp]
\caption{Accuracy (\%) for Qwen3.5-9B and fine-tuned variants under no-hint and expert-hint configurations. Values are mean $\pm$ sample standard deviation across five inference runs (seeds 0--4), with $K=1$.}
\label{tab:qwen_results}
\centering
\small
\setlength{\tabcolsep}{3pt}
\resizebox{\linewidth}{!}{\begin{tabular}{llcccccc}
\toprule
& & \multicolumn{2}{c}{\textbf{IndustryOR}}
& \multicolumn{2}{c}{\textbf{Mamo-Complex}}
& \multicolumn{2}{c}{\textbf{OptMATH}} \\
\cmidrule(lr){3-4}\cmidrule(lr){5-6}\cmidrule(lr){7-8}
\textbf{Model} & \textbf{Hints}
& \textbf{1 turn} & \textbf{5 turns}
& \textbf{1 turn} & \textbf{5 turns}
& \textbf{1 turn} & \textbf{5 turns} \\
\midrule
Qwen3.5-9B & None & $38.59 \pm 3.94$ & $61.82 \pm 3.30$ & $41.90 \pm 2.21$ & $76.10 \pm 1.09$ & $44.69 \pm 3.55$ & $65.00 \pm 2.78$ \\
& Expert & $48.08 \pm 6.76$ & $65.05 \pm 3.00$ & $59.52 \pm 3.90$ & $85.90 \pm 2.47$ & $59.38 \pm 3.41$ & $73.91 \pm 2.25$ \\
\midrule
Qwen-SFT & None & $66.87 \pm 2.52$ & $67.88 \pm 4.25$ & $84.95 \pm 2.25$ & $83.81 \pm 2.99$ & $77.03 \pm 2.63$ & $79.69 \pm 4.17$ \\
& Expert & $66.46 \pm 1.66$ & $64.85 \pm 3.81$ & $86.19 \pm 1.75$ & $86.19 \pm 1.06$ & $78.28 \pm 1.50$ & $79.69 \pm 2.41$ \\
\midrule
Qwen-RL-multi & None & $69.29 \pm 1.53$ & $78.79 \pm 1.24$ & $74.00 \pm 1.60$ & $90.10 \pm 1.19$ & $74.22 \pm 5.15$ & $86.25 \pm 1.05$ \\
& Expert & $\mathbf{77.78 \pm 1.24}$ & $\mathbf{80.61 \pm 2.41}$ & $\mathbf{88.10 \pm 2.45}$ & $\mathbf{94.00 \pm 0.43}$ & $\mathbf{81.25 \pm 3.22}$ & $\mathbf{87.97 \pm 1.52}$ \\
\bottomrule
\end{tabular}}
\end{table}

\textit{The pipeline transfers across base models, and RL requires less model-specific data generation.} Qwen-SFT substantially improves first-turn accuracy, confirming that the cleaned corpus is not tied to GPT-OSS. However, it requires newly distilled and filtered trajectories from a strong Qwen teacher. Qwen-RL directly reuses the GPT-OSS training data without additional post-processing.

\textit{Multi-turn RL shows larger accuracy gains from additional solver-feedback turns and expert hints than SFT.} From turn one to turn five, multi-turn RL gains 2.83--16.10 points, whereas SFT changes by only $-1.61$ to $+2.66$ points. At five turns, expert hints improve multi-turn RL on all three benchmarks by 1.72--3.90 points, while the SFT changes range from $-3.03$ to $+2.38$ points.

\textit{Multi-turn RL improves accuracy across the evaluated settings.} At five turns, RL exceeds SFT in all six matched dataset/hint settings by 6.29--15.76 points. Both models receive the same initial prompts within each hint condition and use solver feedback, consistent with stronger iterative revision by RL. This advantage extends beyond the three primary benchmarks: in the no-hint cross-model evaluation, multi-turn RL achieves the highest observed five-turn mean within each backbone on both the held-out arXiv dataset and the standalone 92-problem LogiOR benchmark (Tables~\ref{tab:ood_custom_results} and~\ref{tab:standalone_logior}).

\Needspace{6\baselineskip}
\subsection{\texorpdfstring{Generalization on the arXiv Dataset}{Generalization on the arXiv Dataset}}
\label{app:ood_generalization}
\label{app:generalization_results}

To test whether OptiMind rigidly depends on the predefined taxonomy, we evaluated the GPT-OSS-20B base model and its SFT, single-turn RL, and multi-turn RL variants on the arXiv dataset, an 8-problem held-out evaluation set outside the training taxonomy based on recent optimization papers from arXiv and Optimization Online, covering applications such as sports scheduling with travel distance, open-pit mine production planning, and political districting. These problems do not cleanly fit the 53 classes, and we did not generate custom hints for them; instead, we reused the existing hint library as a stress test of whether distilled optimization expertise transfers to novel domains. We also evaluate our recipe on Qwen3.5-9B, comparing its base, SFT, and multi-turn RL variants; see Appendix~\ref{app:qwen_transfer} for adaptation details. Table~\ref{tab:ood_custom_results} reports the resulting accuracies.

For the single-turn and multi-turn \textsc{OptiMind-RL} models, we evaluate each hint condition over ten runs with seeds 0--9. Each run uses temperature 0.9, top-$p=1.0$, and one sample per turn ($K=1$), and continues for all five turns without early stopping. For these four rows, the one-turn and five-turn accuracies are measured within the same runs and reported as the mean and sample standard deviation across runs, using the $n-1$ denominator.

For Qwen, the base and SFT models use ten runs (seeds 0--9), while Qwen-RL-multi uses five (seeds 0--4) at the same v3 checkpoint as Table~\ref{tab:qwen_results}. All three use temperature 0.6, top-$p=0.95$, $K=1$, a total token budget of 131,072, and a per-turn completion cap of 24,576. Every run evaluates all eight problems for five turns without early stopping; the first- and fifth-turn accuracies are summarized as mean $\pm$ sample standard deviation.

\begin{table}[htbp]
\caption{Accuracy (\%), reported as mean $\pm$ sample standard deviation, on the arXiv dataset ($K=1$). Qwen base/SFT use ten runs and Qwen-RL-multi uses five. ``Reused expert'' denotes the existing hint library; no custom hints were created for this set.}
\label{tab:ood_custom_results}
\centering
\small
\setlength{\tabcolsep}{5pt}
\begin{tabular}{llcc}
\toprule
\textbf{Model} & \textbf{Hints} & \textbf{1 turn} & \textbf{5 turns} \\
\midrule
GPT-OSS-20B & None & $30.00 \pm 8.74$ & $42.50 \pm 10.54$ \\
GPT-OSS-20B & Reused expert & $32.50 \pm 10.54$ & $45.00 \pm 10.54$ \\
OptiMind-SFT & None & $33.75 \pm 8.44$ & $47.50 \pm 12.91$ \\
OptiMind-SFT & Reused expert & $38.75 \pm 12.43$ & $47.50 \pm 7.91$ \\
OptiMind-RL-single & None & $43.75 \pm 10.62$ & $45.00 \pm 12.08$ \\
OptiMind-RL-single & Reused expert & $47.50 \pm 5.27$ & $48.75 \pm 3.95$ \\
OptiMind-RL-multi & None & $48.75 \pm 7.10$ & $51.25 \pm 3.95$ \\
OptiMind-RL-multi & Reused expert & $42.50 \pm 8.74$ & $48.75 \pm 3.95$ \\
\midrule
Qwen3.5-9B & None & $12.50 \pm 10.21$ & $40.00 \pm 11.49$ \\
Qwen-SFT & None & $36.25 \pm 7.10$ & $43.75 \pm 10.62$ \\
Qwen-RL-multi & None & $40.00 \pm 10.46$ & $47.50 \pm 10.46$ \\
\bottomrule
\end{tabular}
\end{table}

Without hints, multi-turn RL has the highest observed mean at both turn budgets across both backbones, outperforming SFT by 15.00/3.75 points at one/five turns on GPT-OSS-20B and 3.75/3.75 points on Qwen3.5-9B. For GPT-OSS-20B, reusing the existing hints does not improve multi-turn RL's mean accuracy on this deliberately unmatched set, so its best result does not depend on a favorable taxonomy match.

A representative example is a social seat-assignment instance that is not one of the original 53 categories but contains an assignment substructure: the classifier maps it to the assignment class and retrieves the corresponding hints. With these hints, the model corrects an error that omitted valid student pairs when their order was reversed.
We include the custom instances and representative model-output trajectories in the supplementary material to make this stress test reproducible.

\Needspace{6\baselineskip}
\subsection{Standalone LogiOR Evaluation}
\label{app:standalone_logior}

The standalone LogiOR benchmark~\citep{logior} contains 92 problems and is distinct from the IndustryOR-LogiOR cleaned benchmark variant in Table~\ref{tab:other_benchmarks}. LogiOR provides a mathematical formulation for every problem, allowing us to reproduce and check each ground-truth objective value with a solver. We also repeat this evaluation with the Qwen3.5-9B base model and its SFT and multi-turn RL variants, using our recipe as detailed in Appendix~\ref{app:qwen_transfer}. Table~\ref{tab:standalone_logior} reports no-hint results because these out-of-taxonomy problem families fall outside our class-specific hint taxonomy.

We use ten inference runs (seeds 0--9) for each GPT-OSS variant, five (seeds 0--4) for each Qwen variant, and three API repetitions. All runs use $K=1$ and execute five turns; the reported one- and five-turn scores are accuracies at those turns within the same trajectories. GPT-OSS runs use temperature 0.9, top-$p=1.0$, and medium reasoning effort. Qwen runs use temperature 0.6, top-$p=0.95$, a total token budget of 131,072, and a per-turn completion cap of 24,576. The API requests leave seed, temperature, top-$p$, reasoning effort, and completion limit unspecified; its three repetitions are not deterministically seeded runs.

\begin{table}[htbp]
\caption{Accuracy (\%) on the standalone 92-problem LogiOR benchmark without hints ($K=1$). Values are mean $\pm$ sample standard deviation across ten GPT-OSS runs, five Qwen runs, and three API repetitions.}
\label{tab:standalone_logior}
\centering
\small
\begin{tabular}{lcc}
\toprule
\textbf{Model} & \textbf{1 turn} & \textbf{5 turns} \\
\midrule
GPT-OSS-20B & $55.22 \pm 4.34$ & $66.30 \pm 3.97$ \\
OptiMind-SFT & $60.65 \pm 2.98$ & $66.63 \pm 2.46$ \\
OptiMind-RL-single & $63.80 \pm 1.70$ & $67.93 \pm 2.47$ \\
OptiMind-RL-multi & $63.91 \pm 4.55$ & $68.04 \pm 3.88$ \\
GPT-5 & $59.78 \pm 1.09$ & $66.30 \pm 0.00$ \\
\midrule
Qwen3.5-9B & $33.48 \pm 2.36$ & $56.30 \pm 2.09$ \\
Qwen-SFT & $56.52 \pm 2.17$ & $64.13 \pm 1.88$ \\
Qwen-RL-multi & $59.78 \pm 2.03$ & $68.26 \pm 2.81$ \\
\bottomrule
\end{tabular}
\end{table}

Both fine-tuned variants have higher mean accuracy than the base model, and multi-turn RL has the highest mean within each backbone at both turn budgets. At five turns, accuracy increases from 66.30 for the base model to 66.63 for SFT and 68.04 for multi-turn RL on GPT-OSS, and from 56.30 to 64.13 and 68.26 on Qwen. At one turn, accuracy increases from 55.22 to 60.65 and 63.91 on GPT-OSS, and from 33.48 to 56.52 and 59.78 on Qwen. Compared with GPT-5, GPT-OSS RL is higher by 4.13 points at one turn and 1.74 points at five turns, while Qwen RL matches GPT-5 at one turn and surpasses it by 1.96 points at five turns.

\Needspace{24\baselineskip}
\subsection{General Capability Evaluation}
\label{app:general_capabilities}
\label{app:capability_retention}

We also benchmarked the OptiMind-SFT and RL models on general agentic tool-calling benchmarks (averaged from xLAM~\citep{liu2024apigen} and ToolACE~\citep{liu2024toolacewinningpointsllm}) and LM-eval~\citep{eval-harness} benchmarks (averaged from ARC-Challenge, HellaSwag, WinoGrande, and TruthfulQA-MC2). Table~\ref{tab:general_capabilities} reports the results. For agentic tool calling, SFT degraded the base model performance from 64.75\% to 0.00\% due to incorrect tool-calling syntax, indicating severe forgetting. At the same time, OptiMind-RL achieves 66.80\%. In LM-eval, the base model's performance of 56.50\% degrades to 48.28\% with SFT, while OptiMind-RL achieves 56.11\%. Overall, OptiMind-RL improves optimization performance while largely retaining the base model's performance on the evaluated tool-calling and general-language benchmarks.

For the optimization task, we report the average accuracy over IndustryOR, Mamo-Complex, and OptMATH in the first-turn, no-hint setting.

\begin{table}[htbp]
\caption{General capability evaluation. Agentic scores average xLAM and ToolACE; LM-eval scores average ARC-Challenge, HellaSwag, WinoGrande, and TruthfulQA-MC2. Optimization reports average accuracy over IndustryOR, Mamo-Complex, and OptMATH in the first-turn, no-hint setting.}
\label{tab:general_capabilities}
\centering
\small
\begin{tabular}{lccc}
\toprule
\textbf{Model} & \textbf{Optimization} & \textbf{Agentic} & \textbf{LM-eval} \\
\midrule
GPT-OSS-20B & 65.79 & 64.75 & 56.50 \\
OptiMind-SFT & 78.19 & 0.00 & 48.28 \\
OptiMind-RL & \textbf{79.48} & \textbf{66.80} & 56.11 \\
\bottomrule
\end{tabular}
\end{table}

\clearpage

\newpage

\section{Ablation Studies Results}\label{app:ablation-results}

\begin{figure}[H]
\centering
\includegraphics[width=0.7\linewidth]{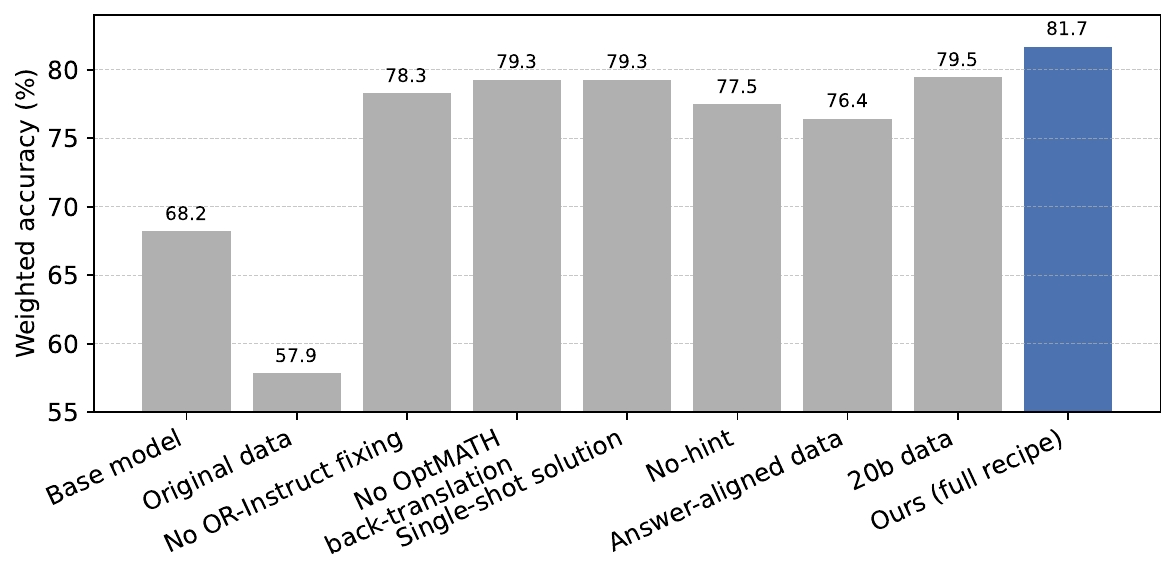}

\caption{Ablation on training-data configurations. We compare seven variants of our training data recipe under the first-turn, no-hint setting and report weighted average accuracy across IndustryOR, Mamo-Complex, and OptMATH.}
\label{fig:ablation_data}
\end{figure}

We present the additional ablation studies details in Table \ref{tab:rl_hint_ablation}, Fig \ref{fig:ablation_data}
, Table \ref{tab:ablation_classification}, and Table \ref{tab:ablation_auto_hint}.

\begin{table}[H]
\caption{Ablation for RL rewards. We report accuracies across 1 and 5 turns under the same prompt and evaluation setup. The terminal-reward row optimizes only final-turn correctness. Terminal reward achieves strong final-turn accuracy but
degrades first-turn accuracy, while average-turn reward better balances early correctness, repair, and
stability.}
\label{tab:rl_hint_ablation}
\begin{center}
\resizebox{0.98\linewidth}{!}{
\begin{tabular}{l l cc cc cc}
\toprule
\textbf{RL variant} & \textbf{Hint strategy} &
\multicolumn{2}{c}{\textbf{IndustryOR}} &
\multicolumn{2}{c}{\textbf{Mamo-Complex}} &
\multicolumn{2}{c}{\textbf{OptMATH}} \\
\cmidrule(lr){3-4} \cmidrule(lr){5-6} \cmidrule(lr){7-8}
& & \textbf{1 turn} & \textbf{5 turns}
& \textbf{1 turn} & \textbf{5 turns}
& \textbf{1 turn} & \textbf{5 turns} \\
\midrule
Single-turn RL & No hint
& $71.21 \pm 2.90$ & $75.38 \pm 2.52$
& $86.01 \pm 0.86$ & $90.83 \pm 0.74$
& $75.93 \pm 1.45$ & $81.52 \pm 2.44$ \\
& Expert
& $72.73 \pm 2.02$ & $75.51 \pm 1.73$
& $85.12 \pm 1.71$ & $89.94 \pm 1.15$
& $\mathbf{90.06 \pm 1.49}$ & $\mathbf{92.86 \pm 1.24}$ \\
& Hybrid
& $71.21 \pm 2.47$ & $75.13 \pm 2.42$
& $85.42 \pm 1.04$ & $89.11 \pm 0.81$
& $89.75 \pm 1.84$ & $92.31 \pm 0.87$ \\
\midrule
Terminal reward only & No hint
& $64.69 \pm 3.15$ & $79.80 \pm 2.64$
& $78.54 \pm 3.12$ & $89.45 \pm 0.68$
& $67.36 \pm 1.35$ & $80.87 \pm 1.81$ \\
& Expert
& $64.60 \pm 3.80$ & $\mathbf{80.07 \pm 1.98}$
& $79.81 \pm 1.41$ & $91.84 \pm 1.00$
& $71.56 \pm 2.39$ & $90.60 \pm 1.09$ \\
& Hybrid
& $59.67 \pm 2.21$ & $79.52 \pm 2.47$
& $74.27 \pm 2.51$ & $89.99 \pm 0.48$
& $72.28 \pm 1.49$ & $91.85 \pm 0.72$ \\
\midrule
\textsc{OptiMind-RL} & No hint
& $71.97 \pm 2.36$ & $75.88 \pm 2.05$
& $85.36 \pm 1.34$ & $89.58 \pm 1.13$
& $75.70 \pm 2.57$ & $82.92 \pm 3.07$ \\
& Expert
& $\mathbf{75.76 \pm 2.14}$ & $78.54 \pm 2.07$
& $\mathbf{90.00 \pm 1.73}$ & $\mathbf{92.26 \pm 1.23}$
& $85.17 \pm 2.05$ & $90.99 \pm 0.93$ \\
& Hybrid
& $74.24 \pm 0.87$ & $76.52 \pm 1.49$
& $88.15 \pm 1.08$ & $89.70 \pm 0.98$
& $87.58 \pm 0.31$ & $91.46 \pm 1.23$ \\
\bottomrule
\end{tabular}}
\end{center}
\end{table}

\begin{table}[H]
\caption{Ablation on classification and hinting strategies on Mamo-Complex. We report first-turn and 5-turn accuracies (in percentage points).}
\label{tab:ablation_classification}
\begin{center}
\resizebox{0.55\linewidth}{!}{
\begin{tabular}{l c c}
\toprule
\textbf{Classification / Hint Strategy} & \textbf{1 turn} & \textbf{5 turn} \\
\midrule
\textsc{gpt-oss} classifier (ours) & 83.81 & 89.90 \\
Human expert classification & 81.04 & 89.61 \\
All hints, no classification          & 78.57 & 89.23 \\
Random classes                        & 72.57 & 87.23 \\
No hints                              & 70.00 & 86.57 \\
\bottomrule
\end{tabular}}
\end{center}
\end{table}

\begin{table}[H]
\caption{Ablation on automatic hint generation stages. We report first-turn accuracies (in percentage points) for the base model (gpt-oss-20B) and \textsc{OptiMind-SFT} on each benchmark.}
\label{tab:ablation_auto_hint}
\begin{center}
\resizebox{0.95\linewidth}{!}{
\begin{tabular}{l c c c c c c}
\toprule
\textbf{Hint} &
\multicolumn{3}{c}{\textbf{gpt-oss-20B}} &
\multicolumn{3}{c}{\textbf{\textsc{OptiMind-SFT}}} \\
\cmidrule(lr){2-4} \cmidrule(lr){5-7}
& \textbf{IndustryOR} & \textbf{Mamo-Complex} & \textbf{OptMATH}
& \textbf{IndustryOR} & \textbf{Mamo-Complex} & \textbf{OptMATH} \\
\midrule
Raw \textsc{GPT-5} hints  & 62.19 & 66.57 & 75.91 & 66.86 & 80.95 & 83.43 \\
Refined hints             & 63.83 & 67.85 & 76.34 & 67.47 & 82.22 & 85.54 \\
Refined + filtered hints  & 61.31 & 68.14 & 77.03 & 69.39 & 82.76 & 87.11\\
\bottomrule
\end{tabular}}
\end{center}
\end{table}

\newpage
\section{Sample Problem Description and Output Optimization Modeling} \label{app:inout}

\begin{figure}
    \centering
    \includegraphics[width=0.6\linewidth]{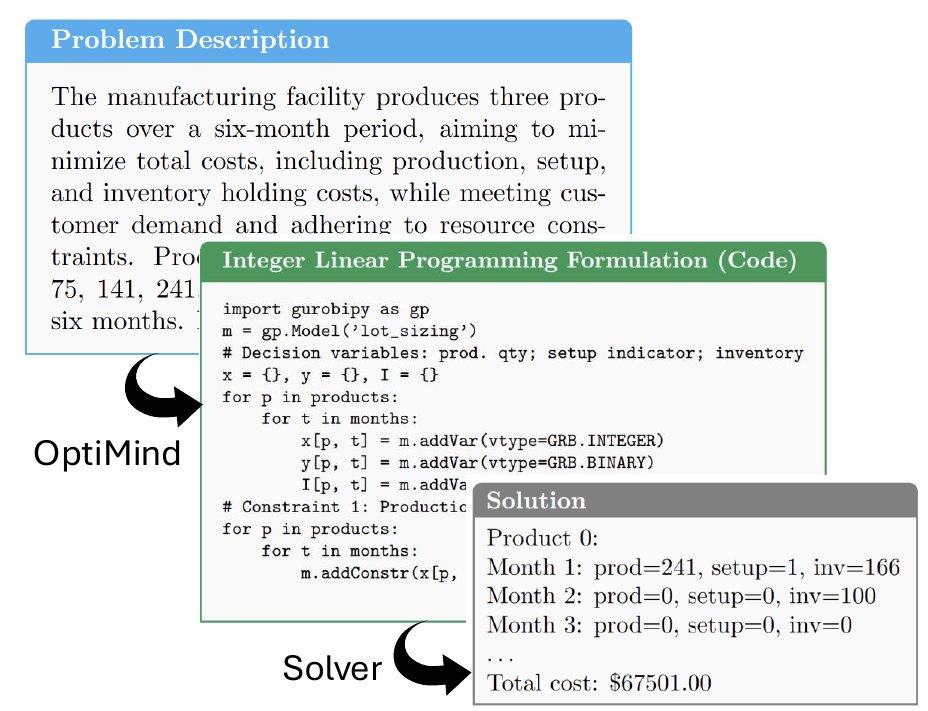}
        \vspace{6pt}
  \caption{ From problem description to solution.}
  \label{fig:optimind}
\end{figure}

Below we have a sample problem description from the dataset OptMATH:

\noindent\fbox{\begin{minipage}{\dimexpr\textwidth-2\fboxsep-2\fboxrule\relax}
\textbf{Question:} Maximize the total profit by determining the optimal production, inventory, and sales plan for three products (Product 0, Product 1, and Product 2) over a five-period planning horizon. The profit per unit sold is \$269 for Product 0, \$282 for Product 1, and \$241 for Product 2. The holding cost for each unit stored is \$15 per period. At the start of the planning horizon (Period 0), the production of each product must equal the sum of its sales and inventory for that period. For each subsequent period, the inventory carried over from the previous period plus the production in the current period must equal the sum of sales and inventory for the current period. The total production time required for all products on each machine type must not exceed the available capacity. The capacities are 480 hours per period for grinders, 320 hours per period for drills, and 160 hours per period for borers. Each unit of Product 0 requires 1 hour on grinders, 1 hour on drills, and 2 hours on borers. Each unit of Product 1 requires 1 hour on grinders, 1 hour on drills, and 2 hours on borers. Each unit of Product 2 requires 1 hour on grinders, 2 hours on drills, and 1 hour on borers. The inventory of each product at the end of any period cannot exceed 80 units. The maximum number of units that can be sold for each product in each period is as follows: Product 0: 48 in Period 0, 43 in Period 1, 58 in Period 2, 58 in Period 3, and 61 in Period 4; Product 1: 54 in Period 0, 56 in Period 1, 45 in Period 2, 46 in Period 3, and 40 in Period 4; Product 2: 57 in Period 0, 52 in Period 1, 68 in Period 2, 40 in Period 3, and 60 in Period 4. At the end of the planning horizon (Period 4), the inventory of each product must be exactly 20 units. Ensure that the final inventory levels meet the target requirements and that production does not exceed machine capacities.\\

\# Note:\\
- The Code must include:```python\\

import gurobipy as gp\\
from gurobipy import GRB\\
```\\
- Make sure the model variable is named `model`.\\
- Avoid using "$<$" $>$" in Gurobi constraints; instead, use "$<=$" or "$>=$" as appropriate.\\
- Carefully determine whether the variable is an integer or a continuous variable.
\end{minipage}}

Below we have the ground-truth output code formulates this problem as an MILP; the output also includes an initial description of the mathematical formulation in markdown format, which is omited for brevity:

\begin{lstlisting}[language=Python]
import gurobipy as gp
from gurobipy import GRB

# Data
products = [0, 1, 2]
periods = [0, 1, 2, 3, 4]
machines = ['grinders', 'drills', 'borers']

profit = {0: 269, 1: 282, 2: 241}
holding_cost = 15

machine_capacity = {
    'grinders': 480,
    'drills': 320,
    'borers': 160
}

machine_time = {
    'grinders': {0: 1, 1: 1, 2: 1},
    'drills': {0: 1, 1: 1, 2: 2},
    'borers': {0: 2, 1: 2, 2: 1}
}

max_sales = {
    0: {0: 48, 1: 43, 2: 58, 3: 58, 4: 61},
    1: {0: 54, 1: 56, 2: 45, 3: 46, 4: 40},
    2: {0: 57, 1: 52, 2: 68, 3: 40, 4: 60}
}

# Model
model = gp.Model("Production_Planning")

# Decision Variables
x = model.addVars(products, periods, name="Production")
s = model.addVars(products, periods, name="Sales")
I = model.addVars(products, periods, name="Inventory")

# Objective Function
model.setObjective(
    gp.quicksum(profit[i] * s[i, t] for i in products for t in periods) -
    gp.quicksum(holding_cost * I[i, t] for i in products for t in periods),
    GRB.MAXIMIZE
)

# Constraints
# Initial Inventory Balance (Period 0)
model.addConstrs((x[i, 0] == s[i, 0] + I[i, 0] for i in products), name="Initial_Balance")

# Inventory Balance (Periods 1-4)
model.addConstrs((I[i, t-1] + x[i, t] == s[i, t] + I[i, t] for i in products for t in periods if t >= 1), name="Inventory_Balance")

# Machine Capacity Constraints
model.addConstrs((gp.quicksum(machine_time[m][i] * x[i, t] for i in products) <= machine_capacity[m] for m in machines for t in periods), name="Machine_Capacity")

# Sales Constraints
model.addConstrs((s[i, t] <= max_sales[i][t] for i in products for t in periods), name="Sales_Constraint")

# Inventory Constraints
model.addConstrs((I[i, t] <= 80 for i in products for t in periods), name="Inventory_Constraint")

# Final Inventory Target
model.addConstrs((I[i, 4] == 20 for i in products), name="Final_Inventory_Target")

# Optimize
model.optimize()

# Results Interpretation
if model.status == GRB.OPTIMAL:
    print("Optimal Solution Found!")
    print(f"Total Profit: ${model.ObjVal:.2f}")
    
    for i in products:
        print(f"\nProduct {i}:")
        for t in periods:
            print(f"Period {t}: Production={x[i,t].X:.2f}, Sales={s[i,t].X:.2f}, Inventory={I[i,t].X:.2f}")
else:
    print("No optimal solution found.")
\end{lstlisting}

\clearpage
\newpage
\section{Examples of RL Rollout Trajectories}
\label{app:rl-trajectories}

\normalcolor
The following examples focus on representative multi-turn RL behaviors rather
than tabulating every solver run.  Each case shows the problem statement, the
local mistake in the initial trajectory, and the later self-correction that
changes the formulation while leaving unrelated model structure intact.

\noindent\textbf{{}{Illustrated behavior}: \textbf{semantic objective repair from a solver-valid but wrong program}.}\par
\noindent
In IndustryOR \#74, the first program executes successfully and returns a
zero-profit optimum, but that solution is economically suspicious: same-quarter
buy-and-sell decisions should earn the posted margin and should not pay a
storage fee.  The later RL response identifies this semantic mismatch and
changes only the objective coefficient for immediate sales.

\begin{descriptionbox}[IndustryOR \#74: Storage-Cost Repair]
\begin{lstlisting}[style=md,basicstyle=\ttfamily\footnotesize,escapeinside={(*@}{@*)}]
# Problem statement
**A timber company buys at the beginning of each quarter and may sell in the
same quarter or store timber for later sale.** Warehouse capacity is 200,000
m^3.  Storage cost is (a + b u) yuan per m^3, where a = 70, b = 100, and u is
the number of quarters stored.  All inventory must be sold by the end of
autumn, and the objective is to maximize annual profit in units of 10,000 yuan.

Quarter   purchase price   sale price   max sale
Winter          410            425          100
Spring          430            440          140
Summer          460            465          200
Autumn          450            455          160

# RL turn 0: executable but wrong
The initial trajectory builds a valid Gurobi model, but its objective treats
every sale as if it had been stored.

**Wrong local term: same-quarter sales pay storage cost.**

  net_per_unit =
      sale_price[t] - purchase_price[k] - (a + b * (t - k))

For t == k, this subtracts a = 70 even though the timber is bought and sold in
the same quarter.  Winter's immediate margin becomes 425 - 410 - 70 = -55
instead of +15, so the solver-valid model prefers doing nothing.

# RL turn 2 correction: diagnosis and local repair
Thinking sample after solver feedback:

  The model returned zero profit.  Maybe storage cost is only for stored
  inventory.  Our model wrongly applies storage cost to all sales even in the
  same quarter.  For t == k, storage cost should be zero.

**Corrected local term: immediate sale is exempt from storage cost.**

  if t == k:
      net_per_unit = sale_price[t] - purchase_price[k]
  else:
      storage_cost = a + b * (t - k)
      net_per_unit = sale_price[t] - purchase_price[k] - storage_cost

# Later RL sample
The later RL response keeps the same flow variables and capacity constraints,
and preserves the repaired piecewise objective:

  Immediate sale, no storage cost.
  Storage cost applies only for timber carried to a later quarter.

(*@\textbf{{}{Illustrated behavior}:}@*) **semantic repair without a runtime exception**.  The
policy uses a suspicious optimal objective as evidence of a formulation error
and edits the economic meaning of one objective term.
\end{lstlisting}
\end{descriptionbox}

\normalfont\normalcolor

\noindent\textbf{{}{Illustrated behavior}: \textbf{conservation-law diagnosis from infeasibility}.}\par
\noindent
In Mamo-Complex \#193, the first program fails because it forces final stock to
equal each warehouse's requirement exactly.  Since total initial stock exceeds
total required stock, exact equality leaves no place for surplus.  The later RL
response detects the aggregate conservation contradiction and changes equality
requirements into lower-bound requirements.

\begin{descriptionbox}[Mamo-Complex \#193: Surplus-Preserving Transshipment Repair]
\begin{lstlisting}[style=md,basicstyle=\ttfamily\footnotesize,escapeinside={(*@}{@*)}]
# Problem statement
**Seven warehouses may send supplies to one another at route-dependent costs.**
The goal is to meet every warehouse's required stock level at minimum movement
cost.  Existing stock can remain in place; the problem asks that needs are met,
not that surplus inventory disappears.

Warehouse       1    2    3    4    5    6    7
Initial        57  180  288  255  477  222  314
Required      499  173  293  158   15   78  278

**Total initial stock = 1793, total required stock = 1494, surplus = 299.**

# RL turn 0: infeasible equality model
The initial trajectory models the final stock at each warehouse as exactly the
required amount.

**Wrong local constraint: surplus has no feasible destination.**

  initial[i] + inflow[i] - outflow[i] == demand[i]

Summing these equalities over all warehouses cancels every internal shipment.
The model therefore requires 1793 == 1494, which is impossible.

# RL turn 2 correction: aggregate-balance reasoning
Thinking sample after infeasibility feedback:

  Summing all balance equations cancels internal flows, so exact equality
  would require total initial stock to equal total demand.  Since supply is
  larger, the extra units need to remain somewhere.  Final stock should be at
  least the required amount.

**Corrected local constraint: requirements are lower bounds.**

  initial[i] + inflow[i] - outflow[i] >= demand[i]

# Later RL sample
The later RL response keeps the same shipment variables and transportation
objective, and preserves the lower-bound stock constraints:

  final supply must be at least demand
  minimize total transportation cost

(*@\textbf{{}{Illustrated behavior}:}@*) **structural repair from infeasibility**.  The policy
does not patch the data or rewrite the route costs; it identifies the conserved
quantity and changes only the balance semantics.
\end{lstlisting}
\end{descriptionbox}

\normalfont\normalcolor

\clearpage
\newpage
\section{Details of Automatic Hint Generation}\label{app:auto_hint_generation}\label{app:error_hint_details}

\subsection{Details of Error-Analysis and Hints}\label{app:error-analysis}

\begin{figure}[tbp]\begin{center}
\includegraphics[width=0.9\linewidth]{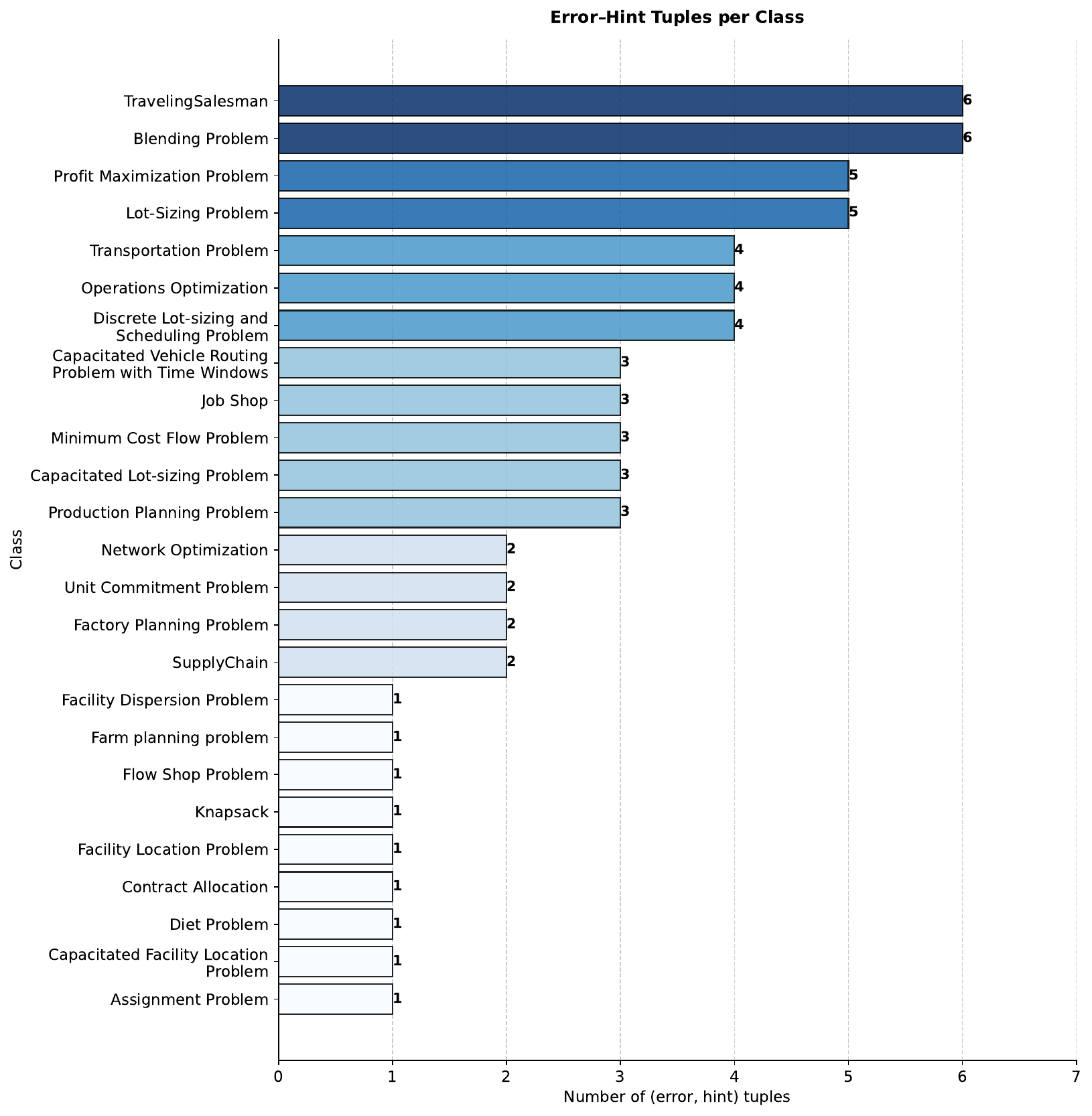}  

\end{center}
\caption{Number of error-hint pairs that experts collected per class.}
\label{fig:per_class_hint_num}
\end{figure}

This section provides the details for Section~\ref{sec:error_analysis}. It describes how we mine recurring formulation failures, how expert-curated hints are written, and how the expert and automatic hint sets are merged into a hybrid library.

\subsection{Class-Based Failure Mining}
\label{app:failure_mining}

We first classify all training problems from OR-Instruct~\citep{tang2024orlm} and OptMATH~\citep{lu2025optmath} into the 53 seed classes defined by the OptMATH problem generators. These classes cover a broad range of common MILP-style problems, including transportation, facility location, lot sizing, production planning, assignment, scheduling, and routing. We then run the base model, \textsc{gpt-oss-20b}, on each training problem and group the generated attempts by predicted class.

Within each class, we inspect recurring disagreements between generated answers and available references. Many errors are class-specific rather than instance-specific. For example, in Traveling Salesman Problem instances, the model often applies MTZ subtour-elimination constraints to the fixed start node, making the ordering constraints infeasible. In network-flow and inventory problems, the model often reverses the sign convention in flow conservation. If $B[u]$ denotes net supply at node $u$, with positive values representing surplus, the correct balance is
\[
    \text{outflow}(u) - \text{inflow}(u) = B[u],
\]
but we often observe the inflow and outflow terms swapped. In facility-location, production-planning, and assignment problems, common errors include incorrect integrality assumptions, missing capacity coupling, and missing linking constraints.

These repeated patterns motivate class-conditioned hints. Each hint consists of an error summary and a preventive modeling rule. The hint is intended to be reusable across instances in the same class, rather than a full formulation template for a single problem.

\begin{table}[t]
\caption{Representative examples of error-analysis hints from optimization problem classes. }
\label{tab:hint_examples}
\centering
\begingroup
\small
\setlength{\tabcolsep}{3pt}
\renewcommand{\arraystretch}{1.08}
\resizebox{0.92\linewidth}{!}{\begin{tabular}{@{}>{\raggedright\arraybackslash}p{0.16\linewidth}
                >{\raggedright\arraybackslash}p{0.42\linewidth}
                >{\raggedright\arraybackslash}p{0.36\linewidth}@{}}
\toprule
\textbf{Problem Class} & \textbf{Common error pattern} & \textbf{Hint} \\
\midrule
Traveling Salesman
&
Apply \textbf{MTZ subtour-elimination constraints} to the start city, making the ordering constraints infeasible.
&
Fix the start city's position, and apply the MTZ constraint for each visited pair of cities except the start city.
\\
\midrule
Job Shop
&
Use \textbf{two} binaries for a pair of operations without exclusivity, e.g., both ``$a$ before $b$'' and ``$b$ before $a$'' allows inconsistent ordering.

\smallskip
\textbf{Big-M} is too small, making the disjunction invalid and incorrectly ruling out feasible schedules.
&
Use \textbf{one} ordering binary per unordered pair: $y_{\{a,b\}}=1$ means operation $a$ is scheduled before $b$.

\smallskip
Set $M$ to $100$ plus the sum of processing times of all operations.
\\
\midrule
Capacitated Facility Lot-Sizing
&
Backlog is often modeled with the \textbf{wrong sign on $B_t$} or by using a ``net inventory'' term ($I_t - B_t$) inconsistently, which breaks flow conservation.
&
Use the exact balance equation $I_{t-1} + x_t + B_t = I_t + d_t + B_{t-1}$.
\\
\bottomrule
\end{tabular}}
\endgroup
\end{table}

\subsection{Expert-Curated Hints}
\label{app:expert_curated_hints}

For each problem class, we run \textsc{gpt-oss-20b} to produce candidate answers and identify 10--20 instances where the generated answer disagrees with the original label. Such a mismatch may be caused by either a model-side formulation error or a data-side issue such as an ambiguous statement, missing parameter, infeasible instance, or incorrect ground-truth objective value. Optimization experts review each selected item to identify the source of disagreement.

When the model is at fault, the annotator writes two short fields:
\begin{enumerate}
    \item an \emph{error summary}, describing the conceptual modeling mistake; and
    \item a \emph{preventive hint}, describing how to avoid the mistake in future instances of the same class.
\end{enumerate}
These annotations are automatically aggregated into a dictionary mapping each problem class to a list of \texttt{(error summary, preventive hint)} pairs. We then make a full standardization pass to align wording, remove duplicate phrasing, and ensure that hints are reusable rather than tied to instance-specific numbers. For quality control, at least three experts annotate overlapping representative problems, and disagreements are resolved by majority vote or discussion.

\begin{figure*}[t]
    \centering
    \begin{subfigure}[t]{0.30\textwidth}
        \begin{bluebox}[Hint Example for TSP]
        \begin{spacing}{0.81} 
        \ttfamily\tiny\linespread{0.9}
    
        \textbf{Error description:} ``When solving TSP, it is easy to make mistakes by not enforcing the MTZ subtour elimination constraints correctly.''
        
        \vspace{7pt}
        \textbf{Hint:} 
        ``When solving TSP using MTZ, fix one city's position, e.g., $u[0]=0$, and apply $u[i] - u[j] + n*x[i,j] \leq n - 1$ for the remaining cities $i,j \neq 0$ with $i \neq j$; do not apply this constraint for the city whose position you fixed.''
        \end{spacing}
    \end{bluebox}
    \end{subfigure}
    \hfill
    \begin{subfigure}[t]{0.37\textwidth}
        \begin{bluebox}[Response without Hint]
        \begin{spacing}{0.82} 
        \ttfamily\tiny\linespread{0.9}
         \textbf{Thought:} ``We need to solve TSP with 6 nodes, directed distances given. We need to formulate TSP as integer program: binary variables $x[i,j]=1$ if travel from $i$ to $j$. Constraints: each city has exactly one outgoing edge and one incoming edge. Subtour elimination: use Miller-Tucker-Zemlin constraints. {\color{red}For all $i\neq j$, $u[i] - u[j] + n*x[i,j] \leq n-1$}. Also need to ensure start city 0? Usually we set $u[0]=0$. We'll implement accordingly.''
         \end{spacing}
    \end{bluebox}
    \end{subfigure}
    \hfill
    \begin{subfigure}[t]{0.29\textwidth} 
       \begin{bluebox}[Response with Hint]
       \begin{spacing}{0.82}  
        \ttfamily\tiny
         \textbf{Thought:} ``Use MTZ subtour elimination: choose node 0 as fixed with $u[0]=0$, $u[i]$ integer $1,\ldots,n-1$ for $i>0$. {\color{ForestGreen}The hint says not to apply the constraint for the city whose position is fixed. So for $i \neq j$, $i>0$, $j>0$: $u[i] - u[j] + n*x[i,j] \leq n-1$.}''
         \vspace{1pt}
         \end{spacing}
    \end{bluebox}
    \end{subfigure}
    \caption{Example hint for the Traveling Salesman Problem and comparison of the model's response before and after applying the hint. }
    \label{fig:sample-hints}
\end{figure*}

\subsection{Automatic Hint Generation} \label{app:auto-hint-details}

This section describes the detail automatic hint generation pipeline used in \S\ref{sec:error_analysis}. Our goal is to replace a large portion of manual expert effort with a scalable distillation procedure from a stronger teacher model. Intuitively, if a teacher model can (i) produce a reliable reference solution and (ii) explain why the base model's attempt fails, then we can automatically synthesize the same kind of error summaries and preventive hints that experts would write.

We generate the fully-automatic hints in five stages: 

\begin{enumerate}
    \item \textbf{Identify high-confidence disagreement cases.} We run \textsc{gpt-oss-20B} on the training set to obtain a candidate solution attempt, and we run \textsc{GPT-5} with $K=8$ samples and take majority voting as a high-confidence teacher reference. The solution matching rate is approximately $80\%$ on OR-Instruct and $90\%$ on OptMATH. To reduce noise from ambiguous instances, we restrict to problems where at least $6/8$ \textsc{GPT-5} samples agree and where the teacher majority differs from \textsc{gpt-oss-20B}, yielding 350 candidate problems from OR-Instruct and 175 from OptMATH.
    \item \textbf{Synthesize candidate hints.} For each selected mismatch, we prompt \textsc{GPT-5} to produce a concise hint that targets the likely failure mode of the base model. Specifically, we prompt the teacher model to produce an error summary and a preventive hint, following the format used for expert-curated hints described in Section~\ref{app:expert_curated_hints}. We filter out cases that are ambiguous or that correspond to trivial coding/syntax mistakes, resulting in 200 nontrivial hints from OR-Instruct and 100 from OptMATH.
    \item \textbf{De-duplicate within each problem class.} Since many hints are highly similar (e.g., multiple instances trigger the same conceptual mistake), we aggregate hints by problem class and prompt \textsc{GPT-5} to merge duplicates and near-duplicates. This reduces redundancy and yields a compact set of unique hints (291 in our final library).
    \item \textbf{Stability refinement.} We apply an additional refinement pass that rewrites each hint to be safe and reusable across variations within the same class. This step aims to remove instance-specific details and reduce brittleness when the hint is applied to different parameterizations or problem statements.
        \item \textbf{Validation-based filtering.} Finally, we select the most useful hints by applying each hints individually on the validation set, keeping only those that exceed a $4\%$ improvement threshold. This step removes ineffective or occasionally harmful hints and produces the final automatic hint set used in our experiments.
\end{enumerate}

The default implementation above uses \textsc{GPT-5} as the teacher, but the procedure is teacher-agnostic; Appendix~\ref{app:open_teacher_hints} reports a variant using \textsc{gpt-oss-120b} as an open-source teacher.

Table~\ref{tab:auto_hint_examples} shows two representative automatic hints. They are intentionally short, operational, and recognizable to OR practitioners, capturing recurring modeling errors such as mishandled self-loops and missing node-wise flow conservation.

\begin{table}[H]
\caption{Examples of automatically generated hints.}
\label{tab:auto_hint_examples}
\begin{center}
\resizebox{\linewidth}{!}{

\begin{tabular}{p{5cm} p{10cm}}
\toprule
\textbf{Error pattern} & \textbf{Hint} \\
\midrule
Self-loop arcs $(i=j)$ are mishandled, either by adding undefined self-arcs or by forbidding self-arcs that represent local use, holding, or storage. &
Build the arc set from the input data; exclude self-loops only when the instance has no explicit $(i,i)$ arcs, and otherwise preserve them or replace them with explicit local-use variables and matching cost/capacity constraints. \\
Missing node-wise flow conservation allows flow to be created or destroyed when the model only constrains sources and sinks. &
Impose one flow-balance constraint at each node and commodity, e.g., outgoing flow minus incoming flow equals net supply, while adding separate restrictions only when the problem explicitly forbids inflow to sources or outflow from sinks. \\
\bottomrule
\end{tabular}
}
\end{center}
\end{table}

Below we present two examples to illustrate the role of the hint refinement stages: (i) a case study showing two automatically generated hints that were flagged as duplicates and merged during LLM-based de-duplication, and (ii) a comparison of a raw \textsc{GPT-5} hint before and after stability refinement, showing how iterative prompting removes brittle, instance-specific wording and produces a more generally applicable hint. Prompt templates for each stage are provided in Appendix~\ref{app:auto-hint-prompts}.

\begin{descriptionbox}[Example of Duplicated Hints and Automatic Merged Hint]
\ttfamily\small
\textbf{Hint 1:}

('The solution imposes integer restrictions on decision variables even though the original problem specifies only nonnegativity and upper bounds, not integrality. Adding integrality changes the feasible region and can yield a different, suboptimal objective value relative to the correct continuous LP.', 'Do not add integrality constraints unless the problem explicitly requires discrete units; by default, model production quantities as continuous and enforce only the stated bounds and constraints.')

\medskip

\textbf{Hint 2:}

('The solution imposes integer decision variables even though the problem formulation allows continuous production quantities. This unnecessarily restricts the feasible region and can alter the optimal objective value compared to the intended linear program.', 'Do not assume integrality unless the problem explicitly requires discrete units; model production quantities as continuous with bounds, and only introduce integer variables when indivisibility or discrete choices are stated.')

\medskip

\textbf{Hint 3:}

('The solution imposes integer decision variables without justification, turning a linear program into an integer program and potentially changing the optimal value and complexity. The problem statement allows nonnegative quantities but does not require integrality.', 'Do not assume integrality unless explicitly stated; use continuous variables for quantities by default, and only add integer domains when the problem mandates discrete units or explicitly requests them.')

\medskip
\textbf{Merged Hint after Deduplication:}

('Unnecessarily imposing integrality on decision variables when the problem specifies continuous quantities with nonnegativity and bounds.', 'Do not add integrality unless explicitly required; model production/allocation quantities as continuous with the stated bounds to preserve the intended LP feasible set and optimal value.')

\end{descriptionbox}

\vspace{1em}

\begin{descriptionbox}[Comparison of Hints Before and After Refinement (Diet Problem)]
\ttfamily\small
\textbf{Hint Before Refinement:}
Add a binding requirement (equality to required need or a minimum performance/time constraint) so minimization cannot drive all decisions to zero. If allocating continuous time, convert whole-module totals to per-unit rates and include logical upper bounds implied by completion times or capacities.

\medskip
\textbf{Hint After Refinement:}
Add a lower bound requirement (output >= required\_need) rather than a strict equality unless the instance semantics truly demand exact fulfillment, and only do this when the instance provides a trustworthy required\_need parameter that is physically achievable (<= instance capacity and time horizon); if not, use a soft constraint with slack and penalty to avoid infeasibility; convert whole module totals to per unit rates only when unit counts are fixed parameters or when the product of ratequantity can be linearized (e.g., binary multipliers with standard linearization or McCormick envelopes); do not introduce per unit continuous rates for heterogeneous, indivisible, or decision variable module counts without proper integer modeling, and avoid adding ad hoc upper bounds unless they are computed from instance data (capacity, horizon) and verified not to cut off feasible solutions; do not apply these changes blindly to maximization problems or instances where the trivial zero solution is actually valid.
\end{descriptionbox}

\clearpage
\subsection{Automatic Hint Generation Prompts}\label{app:auto-hint-prompts}

\begin{center}
\begin{minipage}[t]{\columnwidth}
\begin{promptbox}[Prompt for Generating Hints]
\footnotesize
You are an expert in optimization modeling and mathematical formulation.
I will give you three inputs:
(1) a problem description,
(2) a ground truth correct solution (reference),
(3) a solution produced by a language model (to be analyzed, likely incorrect).

Your task is to help train future language models to avoid common modeling mistakes.

A) First, analyze the problem and the provided "Ground Truth Solution" to understand the correct modeling approach and whether the problem is solvable as stated.
\begin{itemize}
  \item If the problem is solvable and the Ground Truth is valid, proceed to step C.
  \item Otherwise, proceed to step B.
\end{itemize}

B) If the problem is NOT solvable as stated (e.g., ambiguous):
\begin{itemize}
  \item \tagopen{error\_type}ambiguous problem\tagclose{error\_type}
  \item \tagopen{error\_summary}\tagclose{error\_summary}
  \item \tagopen{hint}\tagclose{hint}
  \item \tagopen{corrected\_solution}\tagclose{corrected\_solution}
\end{itemize}

C) Compare the "Given Solution" (input 3) with the correct logic.
\begin{itemize}
  \item If the Given Solution is actually correct (equivalent to ground truth):
  \begin{itemize}
    \item \tagopen{error\_type}correct\tagclose{error\_type}
    \item \tagopen{error\_summary}\tagclose{error\_summary}
    \item \tagopen{hint}\tagclose{hint}
  \end{itemize}

  \item If the Given Solution is wrong:
  \begin{itemize}
    \item \tagopen{error\_type}...\tagclose{error\_type}
    \item \tagopen{error\_summary}...\tagclose{error\_summary}
    \item \tagopen{hint}...\tagclose{hint}
  \end{itemize}
\end{itemize}

Error reporting requirements (when wrong):
\begin{itemize}
  \item \tagopen{error\_type}: a short label naming the conceptual mistake (e.g., "missing conservation constraint", "wrong objective units", "indexing mismatch", "invalid integrality").
  \item \tagopen{error\_summary}: 1--3 sentences describing the abstract mistake at a conceptual level. Do NOT mention instance-specific numbers.
  \item \tagopen{hint}: a short reusable hint to help future models avoid this mistake. General and abstract.
\end{itemize}

Your output must contain exactly these four tagged blocks in this exact order:
\begin{itemize}
  \item \tagopen{error\_type}...\tagclose{error\_type}
  \item \tagopen{error\_summary}...\tagclose{error\_summary}
  \item \tagopen{hint}...\tagclose{hint}
\end{itemize}

\# Input data:

Question:
\textcolor{blue}{\{question\}}

Ground Truth Solution:
\textcolor{blue}{\{ground\_truth\}}

Given Solution to Analyze:
\textcolor{blue}{\{incorrect\_solution\}}
\end{promptbox}
\end{minipage}
\end{center}

\begin{promptbox}[Prompt for Deduplicating Hints]
    Given a list of hints to solve a specific optimization problem class with Mixed Integer Programming formulation,
create a list that combines duplicated hints into one.

The goal is to reduce the number of hints while preserving their meaning and details from the original hints.

Provide the summarized hints in JSON format as a list of JSON objects. Each object contains:
\begin{itemize}
  \item \texttt{hint}: a single merged hint as a string
  \item \texttt{indices}: the indices from the original list that were combined
\end{itemize}

\textbf{Example input:}
\begin{enumerate}
  \item Hint A.X
  \item Hint B.Y
  \item Hint C.X
  \item Hint A.Y
  \item Hint B.X
\end{enumerate}

\textbf{Example output:}
\begin{lstlisting}
  {"hint": "hint A", "indices": [1, 4]},
  {"hint": "hint B", "indices": [2, 5]},
  {"hint": "hint C", "indices": [3]}
\end{lstlisting}

Here is the list of hints:

\textcolor{blue}{\{hints\}}

Provide the summarized hints in JSON format as a list of JSON objects.

\end{promptbox}

\begin{promptbox}[Prompt for Refining Hints]
I have a mixed-integer linear optimization problem that is incorrectly formulated. I generated a "hint" that says how to fix the formulation for this instance. However, I will apply this "hint" in all my dataset which consists of different instances of variations of the problem. Judge if the hint is *safe* to be applied to the whole dataset, or if it may introduce issues in the formulation of other instances. 

The hint is: \textcolor{blue}{\{hint\}}
\end{promptbox}

\subsection{Open-Source Teacher for Automatic Hint Generation}\label{app:open_teacher_hints}

To test whether automatic hint generation depends on a proprietary teacher, we replaced \textsc{GPT-5} with the open-source \textsc{gpt-oss-120b} teacher, using majority voting over 64 sampled solutions as the reference before running the same hint synthesis, refinement, and filtering pipeline. Table~\ref{tab:open_teacher_hints} shows that \textsc{gpt-oss-120b}-derived hybrid hints consistently outperform the no-hint baseline and are competitive with the \textsc{GPT-5}-derived variant.

\begin{table}[H]
\caption{Performance of \textsc{gpt-oss-20b} using automatic hints distilled from an open-source \textsc{gpt-oss-120b} teacher.}
\label{tab:open_teacher_hints}
\begin{center}
\resizebox{0.95\linewidth}{!}{

\begin{tabular}{l cc cc cc}
\toprule
& \multicolumn{2}{c}{\textbf{IndustryOR}} & \multicolumn{2}{c}{\textbf{Mamo-Complex}} & \multicolumn{2}{c}{\textbf{OptMATH}} \\
\cmidrule(lr){2-3}\cmidrule(lr){4-5}\cmidrule(lr){6-7}
\textbf{Hint strategy} & \textbf{1 turn} & \textbf{5 turns} & \textbf{1 turn} & \textbf{5 turns} & \textbf{1 turn} & \textbf{5 turns} \\
\midrule
No hint & $66.23 \pm 3.77$ & $71.89 \pm 3.05$ & $65.03 \pm 3.91$ & $86.80 \pm 1.33$ & $66.68 \pm 4.70$ & $81.52 \pm 3.60$ \\
Expert & $68.28 \pm 2.61$ & $75.05 \pm 1.84$ & $75.85 \pm 2.74$ & $89.47 \pm 1.27$ & $73.32 \pm 4.05$ & $87.39 \pm 1.91$ \\
Automatic (\textsc{gpt-oss-120b}) & $65.43 \pm 2.61$ & $73.63 \pm 3.44$ & $63.52 \pm 1.76$ & $86.57 \pm 0.89$ & $70.07 \pm 3.77$ & $84.58 \pm 2.57$ \\
Hybrid (expert + \textsc{gpt-oss-120b}) & $\bf{69.29 \pm 2.19}$ & $\bf{75.66 \pm 2.26}$ & $\bf{74.86 \pm 3.31}$ & $\bf{89.90 \pm 1.40}$ & $\bf{74.82 \pm 4.28}$ & $88.15 \pm 1.86$ \\
Automatic (\textsc{GPT-5}) & $61.31 \pm 2.65$ & $73.43 \pm 2.07$ & $68.14 \pm 1.91$ & $85.38 \pm 1.43$ & $73.38 \pm 3.82$ & $\bf{88.83 \pm 2.46}$ \\
Hybrid (expert + \textsc{GPT-5}) & $65.25 \pm 3.96$ & $75.05 \pm 1.65$ & $74.10 \pm 2.71$ & $87.57 \pm 1.15$ & $74.65 \pm 4.57$ & $87.41 \pm 3.14$ \\
\bottomrule
\end{tabular}
}
\end{center}
\end{table}

\subsection{Hybrid Hint Library}
\label{app:hybrid_hint_library}

When both expert-curated and fully automatic hints are available, we merge them into a single pool. The expert-curated hints provide high-precision guidance for common modeling failures observed by human annotators, while the automatic hints provide scalable coverage over larger corpora. During merging, we deduplicate near-identical hints, standardize formatting, and preserve hints that target distinct failure modes. The resulting hybrid library is used in three places:
\begin{enumerate}
    \item during solution regeneration, where teacher models solve training problems with class-specific guidance;
    \item when determining ground-truth objective values for RL and evaluation from the regenerated solutions; and
    \item during inference, where the evaluated model receives class-conditioned hints as a conditional modeling checklist.
\end{enumerate}

The current taxonomy uses the 53 OptMATH seed classes as a practical scaffold. It is not a hard limitation: for a new domain, the same procedure can be applied by collecting representative failures, writing or distilling a small number of domain-specific hints, and adding them to the library.

\clearpage
\newpage
\section{Details of Data Cleaning}\label{app:data_cleaning}

\subsection{Training Data Cleaning}\label{app:training_data_cleaning}

This section provides the details for Section~\ref{sec:training_data_cleaning}. We clean training data along two axes: question repair and solution regeneration.

\subsubsection{Question Repair}
For OR-Instruct, we use OpenAI o4-mini to detect missing or underspecified parameters and to fill them with values that yield feasible optimization instances. This process flags $180/2700$ OR-Instruct items as incomplete. The filled values are verified by optimization experts before being used for training. For OptMATH, we follow the original generation process and regenerate 50 instances for each of the 53 seed classes. We then back-translate the problem descriptions with \textsc{gpt-oss-120b}, remove infeasible instances, and manually inspect three regenerated examples per class, for a total of 159 expert-inspected regenerated instances.

\subsubsection{Solution Regeneration}
After repairing problem statements, we regenerate training solutions using \textsc{gpt-oss-120b} with the class-specific expert-curated hints from Appendix~\ref{app:error_hint_details}. For each instance, we sample $K=64$ candidate solutions and use majority voting over executable objective values to select the final answer. This reduces the chance that a single incorrect teacher trajectory becomes the SFT target or supplies an incorrect ground-truth objective value for RL.

\subsubsection{Class Balancing}
To avoid over-representing large synthetic classes, we sample up to 100 training instances per problem class. If a class contains fewer than 100 usable instances, we keep all available instances. The final cleaned training mixture contains $2.2$k cleaned OR-Instruct items and $2.3$k regenerated OptMATH questions. Among the OR-Instruct items, $602/2700$ final answers differ from the original labels.

\subsubsection{Missing-Data Filling}\label{app:missing-data-prompts}
We identify and fill missing data with the semi-automatic procedure below. Representative before/after examples appear after the prompt.

\paragraph{Automatic missing-data infilling prompt.}\label{sec:missing_data}

We design an automatic procedure to fill in missing data in the public available training datasets OR-Instruct~\cite{tang2024orlm} and OptMATH~\cite{lu2025optmath}. First, we generate a gurobipy code to each incomplete question by prompting OpenAI's reasoning models (o1, o3, o4-mini), which often attempts to fill in synthetic data to complete the question. We then extract the synthesized data from the code which corresponds to the missing values. We then use the following prompt to ask o4-mini to produce a modified question with the missing data filled in, where we provide both the original question and the gurobipy code as inputs. Lastly, we manually inspect a subset of 100 repaired questions from OptMATH to make sure that the missing data are successfully filled in.

\begin{promptbox}[Prompt for Missing Data Infilling]
\ttfamily\footnotesize
You are given a optimization question and its mathematical solution 
as well as python code. The question may have missing data or 
parameters that are filled in by the solution. Your task is to identify 
if the solution introduces new numbers/values that should be added into 
the question.

Question: \textcolor{blue}{\{question\}} \\ 
Solution: \textcolor{blue}{\{completion\}} \\

If the question has missing data and the solution provides specific 
numbers that fills these missing parameters, return the modified 
question with the missing parameters imputed. Do not modify anything 
else except for adding in missing data.
Otherwise, return NO MISSING DATA.
Only return the modified question or NO MISSING DATA, nothing else.

\end{promptbox}

Below is an example question from OptMATH that has missing data infilled by our prompting. The red texts are from the original question with missing data. The \textcolor{green!60!black}{green} texts are the infilled texts.

\begin{promptbox}[Example of Automatic Missing Data Infilling]
\ttfamily\footnotesize
\begin{verbatim}
# Question: Aircraft Assignment for Route Coverage

You are responsible for managing the assignment of aircraft to various 
routes for an airline company. The goal is to minimize the total 
operational costs while ensuring that all route demands are met 
and that the availability of each aircraft type is not exceeded.

#### Aircraft and Routes:
- There are 6 types of aircraft (aircraft_0 to aircraft_5) available 
for assignment.
- There are 7 routes (route_0 to route_6) that need to be serviced.

#### Aircraft Availability:
Each aircraft type has a limited number of units available for 
assignment:
- Aircraft_0: 4 units available
- Aircraft_1: 5 units available
- Aircraft_2: 4 units available
- Aircraft_3: 4 units available
- Aircraft_4: 5 units available
- Aircraft_5: 5 units available

#### Route Demands:
Each route has a specific demand that must be satisfied by the combined 
capabilities of the assigned aircraft. The demands are as follows:
- Route_0: Requires at least 275 units of capacity
- Route_1: Requires at least 213 units of capacity
- Route_2: Requires at least 228 units of capacity
- Route_3: Requires at least 265 units of capacity
- Route_4: Requires at least 226 units of capacity
- Route_5: Requires at least 277 units of capacity
- Route_6: Requires at least 246 units of capacity
\end{verbatim}
\begin{OldParts}
#### Aircraft Capabilities:
Each aircraft type contributes differently to the capacity of each 
route. For example:
- **Aircraft_0** contributes 100 units to Route_0, 101 units to Route_1, 
88 units to Route_2, and so on.
- **Aircraft_1** contributes 80 units to Route_0, 88 units to Route_1, 
111 units to Route_2, and so on.
- Similar contributions are defined for all other aircraft types across 
all routes.
\end{OldParts}
\begin{NewParts}
#### Aircraft Capabilities:
Each aircraft type contributes differently to the capacity of each 
route. The capacity contributions \(p_{ij}\) (aircraft \(i\)
to route \(j\)) are:

- Aircraft_0: [100, 101, 88, 95, 110, 120, 105]
- Aircraft_1: [80, 88, 111, 90, 85, 95, 100]
- Aircraft_2: [120, 110, 105, 115, 125, 130, 110]
- Aircraft_3: [90, 95, 100, 105, 110, 115, 120]
- Aircraft_4: [110, 120, 115, 125, 130, 135, 140]
- Aircraft_5: [95, 100, 105, 110, 115, 120, 125]
\end{NewParts}

\begin{OldParts}
#### Operational Costs:
Assigning an aircraft to a route incurs a specific cost. For example:
- Assigning **Aircraft_0** to **Route_0** costs 3778 units.
- Assigning **Aircraft_0** to **Route_1** costs 3344 units.
- Assigning **Aircraft_1** to **Route_0** costs 3660 units.
- Similar costs are defined for all other aircraft-route combinations.

\end{OldParts}

\begin{NewParts}
#### Operational Costs:
Assigning an aircraft to a route incurs a specific cost \(c_{ij}\) 
(aircraft \(i\) to route \(j\)):

- Aircraft_0: [3778, 3344, 3555, 3666, 3777, 3888, 3999]
- Aircraft_1: [3660, 3444, 3555, 3666, 3777, 3888, 3999]
- Aircraft_2: [4000, 3888, 3777, 3666, 3555, 3444, 3333]
- Aircraft_3: [3555, 3666, 3777, 3888, 3999, 4000, 4111]
- Aircraft_4: [3888, 3777, 3666, 3555, 3444, 3333, 3222]
- Aircraft_5: [3666, 3555, 3444, 3333, 3222, 3111, 3000]
\end{NewParts}

\begin{verbatim}
#### Objective:
Determine the number of each aircraft type to assign to each route 
such that:
1. The total operational cost is minimized.
2. The total capacity provided by the assigned aircraft meets 
or exceeds the demand for each route.
3. The number of aircraft assigned does not exceed the availability of 
any aircraft type.

#### Constraints:
- The total number of aircraft of each type assigned across all routes 
must not exceed its availability.
- The combined capacity of all aircraft assigned to a route must meet 
or exceed the route's demand.
- The number of aircraft assigned to any route must be a non-negative 
integer.
\end{verbatim}
\end{promptbox}

Below we provide an example of cleaning training instance of the Job Shop problem with missing data. We describe the semi-automated procedure for infilling missing data in Appendix~\ref{sec:missing_data}.

\begin{descriptionbox}[Job Shop Problem (Before)]
\ttfamily\scriptsize
\# Question:In a manufacturing facility, six distinct jobs must be processed on a set of machines. Each job consists of two operations, and each operation must be performed on a specific machine. The goal is to schedule these operations to minimize the total completion time, known as the makespan. Each operation has a processing time of 1 unit, and operations within the same job must follow a specific sequence, with at least 1 unit of time between consecutive operations. Each machine can process only one operation at a time, and the order of operations on shared machines is determined by binary decision variables. These binary variables ensure that no two operations overlap on the same machine. The makespan must be at least as large as the completion time of every operation, and all start times must be non-negative. The binary variables take values of 0 or 1, and a large constant (100,000) is used in the machine capacity constraints to enforce the order of operations. The objective is to determine the start times of all operations and the order in which they are processed on each machine, ensuring that all constraints are satisfied and the makespan is minimized.\\

\# Note:\\
- The Code must include:```python

import gurobipy as gp\\
from gurobipy import GRB\\
```\\
- Make sure the model variable is named `model`.\\
- Avoid using "$<$" and "$>$" in Gurobi constraints; instead, use "$<=$" or "$>=$" as appropriate.\\
- Carefully determine whether the variable is an integer or a continuous variable.
\end{descriptionbox}

\begin{descriptionbox}[Job Shop Problem (After)]
\ttfamily\scriptsize
\# Question:In a manufacturing facility, six distinct jobs must be processed on a set of machines. Each job consists of two operations, and each operation must be performed on a specific machine. {\color{ForestGreen} In particular, for Job 0: Operation 0 is assigned to Machine 2, and Operation 1 to Machine 1, for Job 1: Operation 0 is assigned to Machine 0, and Operation 1 to Machine 2, for Job 2: Operation 0 is assigned to Machine 1, and Operation 1 to Machine 0, for Job 3: Operation 0 is assigned to Machine 2, and Operation 1 to Machine 2, for Job 4: Operation 0 is assigned to Machine 1, and Operation 1 to Machine 0, and for Job 5: Operation 0 is assigned to Machine 0, and Operation 1 to Machine 1.} The goal is to schedule these operations to minimize the total completion time, known as the makespan. Each operation has a processing time of 1 unit, and operations within the same job must follow a specific sequence, with at least 1 unit of time between consecutive operations. Each machine can process only one operation at a time, and the order of operations on shared machines is determined by binary decision variables. These binary variables ensure that no two operations overlap on the same machine. The makespan must be at least as large as the completion time of every operation, and all start times must be non-negative. The binary variables take values of 0 or 1, and a large constant (1000) is used in the machine capacity constraints to enforce the order of operations. The objective is to determine the start times of all operations and the order in which they are processed on each machine, ensuring that all constraints are satisfied and the makespan is minimized.\\

\# Note:\\
- The Code must include:```python

import gurobipy as gp\\
from gurobipy import GRB\\
```\\
- Make sure the model variable is named `model`.\\
- Avoid using "$<$" and "$>$" in Gurobi constraints; instead, use "$<=$" or "$>=$" as appropriate.\\
- Carefully determine whether the variable is an integer or a continuous variable.
\end{descriptionbox}

\newpage

\subsection{Benchmark Cleaning}\label{app:benchmark_cleaning}

This section describes representative issues we observed in the benchmark data. \space These fall into five main categories: missing data, ambiguous problem descriptions, integral vs.\ fractional variables, wrong solutions, and infeasible problems. We next provide additional details and examples that highlight these challenges.

All cleaned benchmarks have been manually verified by domain experts. Despite earlier claims of correctness in prior work, we found errors in $50\%$ of the original IndustryOR test instances and $41\%$ of the original OptMATH test instances (Table~\ref{tab:error_breakdown}). We apply a structured and repeatable cleaning process with multiple expert passes to address these issues and document every modification in the cleaning logs.

Table~\ref{tab:bench_examples} shows representative examples of incorrect problems and their fixes in IndustryOR and OptMATH, while Table~\ref{tab:error_breakdown} summarizes the frequency and distribution of major error types across all three benchmarks. A complete, per-instance correction record is available in the HTML summary tables described in our supplementary material.

\begin{table}[t]
\centering
\caption{Representative incorrect problems and fixes in IndustryOR and OptMATH.}
\label{tab:bench_examples}
\resizebox{\linewidth}{!}{
\begin{tabular}{l c l p{0.32\linewidth} p{0.32\linewidth}}
\toprule
\textbf{Benchmark} & \textbf{Problem Index} & \textbf{Optimization Class} &
\textbf{Why the original problem is incorrect} &
\textbf{How we fixed it} \\
\midrule
IndustryOR & 24 & -- &
The problem did not state an objective function. &
We added a well-specified objective (minimize pollution). \\
IndustryOR & 46 & -- &
The original instance was infeasible but had a ground-truth objective value of $472.3$. &
We updated the data to make the instance feasible and adjusted the ground-truth objective accordingly. \\
\midrule
OptMATH & 35 & Facility Location Problem (FLP) &
The problem describes selecting towers to cover regions, but the description is missing data about which towers cover which regions. &
We generated the missing coverage data semi-automatically and then verified it manually. \\
OptMATH & 36 & Job Shop &
The instance omits the number of machines and specifies only a partial assignment of jobs to machines. &
We generated the missing machine and assignment data semi-automatically and then verified it manually. \\
\bottomrule
\end{tabular}}
\end{table}

\begin{table}[t]
\centering
\caption{High-level breakdown of benchmark errors. Some instances belong to multiple error categories. Percentages are relative to the total number of problems in each benchmark.}
\label{tab:error_breakdown}
\resizebox{\linewidth}{!}{
\begin{tabular}{l c c c c c c c c c c}
\toprule
\textbf{Benchmark} &
\textbf{\# (\%) Incorrect} &
\textbf{Missing Data} &
\textbf{Ambiguous} &
\textbf{Infeasible} &
\textbf{Wrong GT} &
\textbf{Underspecified} &
\textbf{Specific Solver} &
\textbf{Non-linear} &
\textbf{Inconsistent} &
\textbf{Integral / Fractional} \\
\midrule
IndustryOR &
50 (50\%) &
0 (0\%) &
11 (11\%) &
2 (2\%) &
25 (25\%) &
5 (5\%) &
5 (5\%) &
1 (1\%) &
2 (2\%) &
4 (4\%) \\
MAMO-Complex &
106 (51.2\%) &
15 (7.1\%) &
3 (1.4\%) &
2 (0.9\%) &
42 (19.9\%) &
0 (0\%) &
0 (0\%) &
0 (0\%) &
4 (1.9\%) &
43 (20.4\%) \\
OptMATH &
68 (41\%) &
48 (29\%) &
11 (7\%) &
1 (0.6\%) &
10 (6\%) &
1 (0.6\%) &
0 (0\%) &
0 (0\%) &
0 (0\%) &
0 (0\%) \\
\bottomrule
\end{tabular}}
\end{table}

\space

Cleaning took a month of review by optimization experts (OR faculty and PhD students), with four expert passes over each benchmark. Every modification is documented per instance in HTML correction tables (original and updated problem/answer, and the reason for each change). We evaluate on the SIRL-, LogiOR-, and Survey-cleaned variants of IndustryOR/Mamo-Complex (Table~\ref{tab:other_benchmarks}). The cleaning logs include \texttt{industryOR\_sirl\_vs\_ours.html}, which documents five remaining issues identified in SIRL's cleaned IndustryOR. This provides a complementary cross-check against parallel cleaning efforts. Appendix~\ref{app:anonymous_repo} describes the cleaning logs and ground-truth objective values recorded after cleaning.

\subsubsection{Missing Data}

We observe that many problems contain missing values for key parameters.
For example, an assignment problem in OptMATH reads {\it ``For example, assigning aircraft\verb|_|0 to route\verb|_|0 costs 2552 units, to route\verb|_|1 costs 4340 units, and so on''} without providing the additional costs.  
Interestingly, when data is missing, the OpenAI o-series models synthesize reasonable values for the missing parameters during the formulation generation process. We therefore manually identify problems with missing data, and leverage the fabricated values in their solutions to fill them back into the question description, followed by manual inspection to validate correctness.

\subsubsection{Ambiguous Problem Descriptions}

We also observe that many problems contain ambiguities or inconsistencies. For example, a facility location problem in IndustryOR had the following objective function \textit{``to achieve the goal of minimizing costs and maximizing coverage area''}, thus containing two conflicting objectives.
    These issues are corrected via manual inspection.

\subsubsection{Integral vs Fractional Variables}

Many problems exhibit ambiguity regarding whether decision variables are integer or continuous. For example, a production optimization problem in  IndustryOR refers to ``units of products'', while the reported ground truth corresponds to the production of fractional products. We resolve this by updating the ground truth value, adding explicit clarification sentences that enforce integrality or continuity.

\begin{descriptionbox}[Example of Integer vs. Fractional]
\ttfamily\scriptsize
A company produces product A and product B. Each unit of product A sold generates a profit of £30, while each unit of product B sold generates a profit of £10. The company can allocate a maximum of 40 hours per week for production. Producing one unit of product A requires 6 hours, while producing one unit of product B requires 3 hours. Market demand requires that the quantity of product B produced must be at least three times the quantity of product A. The storage space occupied by product A is four times that of product B, and a maximum of four units of product A can be stored per week.\\

Formulate a model for this problem.\\

Reported ground truth: 146.667\\
Corrected ground truth: \textcolor{ForestGreen}{160}\\
\end{descriptionbox}

\subsubsection{Wrong Solutions}

The benchmarks also contained problems where the provided ground truth value was incorrect. For instance, some minimum-cost flow problems in the OptMATH dataset reported wrong optimal costs (e.g., one problem reported optimal cost of 6 as the ground truth, while the correct optimal value was 4). These errors were all corrected by optimization experts.

\begin{descriptionbox}[Example of Wrong Solution]
\ttfamily\scriptsize

You are responsible for managing the distribution of emergency medical supplies across eight cities in a region. Each city has a specific supply of medical resources and a demand that must be met to ensure adequate healthcare coverage. The goal is to minimize the total transportation cost while ensuring that all cities receive the necessary supplies and that no distribution routes exceed their capacity.\\

City Supply and Demand:\\
- **City 0** has a net demand of 1 unit of medical supplies.\\
- **City 1** has a balanced supply and demand (net demand of 0 units).\\
- **City 2** has a net supply of 1 unit of medical supplies.\\
- **City 3** has a net demand of 2 units of medical supplies.\\
- **City 4** has a balanced supply and demand (net demand of 0 units).\\
- **City 5** has a balanced supply and demand (net demand of 0 units).\\
- **City 6** has a net supply of 2 units of medical supplies.\\
- **City 7** has a balanced supply and demand (net demand of 0 units).\\

Transportation Costs:
The cost of transporting medical supplies between cities varies depending on the route. Below are the transportation costs per unit of supplies:\\

- **From City 0**: To City 1 costs 3, to City 2 costs 2, to City 3 costs 2, to City 4 costs 2, to City 5 costs 3, to City 6 costs 3, and to City 7 costs 1.
- **From City 1**: To City 0 costs 1, to City 2 costs 2, to City 3 costs 3, to City 4 costs 1, to City 5 costs 2, to City 6 costs 1, and to City 7 costs 2.
- **From City 2**: To City 0 costs 2, to City 1 costs 2, to City 3 costs 3, to City 4 costs 3, to City 5 costs 2, to City 6 costs 1, and to City 7 costs 2.
- **From City 3**: To City 0 costs 1, to City 1 costs 2, to City 2 costs 1, to City 4 costs 3, to City 5 costs 3, to City 6 costs 2, and to City 7 costs 3.
- **From City 4**: To City 0 costs 3, to City 1 costs 2, to City 2 costs 1, to City 3 costs 1, to City 5 costs 3, to City 6 costs 2, and to City 7 costs 2.
- **From City 5**: To City 0 costs 1, to City 1 costs 2, to City 2 costs 1, to City 3 costs 2, to City 4 costs 1, to City 6 costs 2, and to City 7 costs 1.
- **From City 6**: To City 0 costs 2, to City 1 costs 3, to City 2 costs 1, to City 3 costs 1, to City 4 costs 1, to City 5 costs 1, and to City 7 costs 1.
- **From City 7**: To City 0 costs 1, to City 1 costs 1, to City 2 costs 3, to City 3 costs 1, to City 4 costs 2, to City 5 costs 3, and to City 6 costs 2.\\

Route Capacity Constraints:
Each route between cities has a maximum capacity for transporting medical supplies:\\

- **From City 0**: To City 1 (7 units), to City 2 (7 units), to City 3 (7 units), to City 4 (7 units), to City 5 (8 units), to City 6 (8 units), and to City 7 (8 units).
- **From City 1**: To City 0 (8 units), to City 2 (7 units), to City 3 (8 units), to City 4 (8 units), to City 5 (7 units), to City 6 (7 units), and to City 7 (9 units).
- **From City 2**: To City 0 (8 units), to City 1 (7 units), to City 3 (7 units), to City 4 (7 units), to City 5 (7 units), to City 6 (9 units), and to City 7 (7 units).
- **From City 3**: To City 0 (7 units), to City 1 (7 units), to City 2 (9 units), to City 4 (8 units), to City 5 (7 units), to City 6 (7 units), and to City 7 (9 units).
- **From City 4**: To City 0 (9 units), to City 1 (7 units), to City 2 (8 units), to City 3 (9 units), to City 5 (7 units), to City 6 (7 units), and to City 7 (7 units).
- **From City 5**: To City 0 (7 units), to City 1 (8 units), to City 2 (9 units), to City 3 (9 units), to City 4 (8 units), to City 6 (9 units), and to City 7 (8 units).
- **From City 6**: To City 0 (9 units), to City 1 (8 units), to City 2 (7 units), to City 3 (8 units), to City 4 (8 units), to City 5 (7 units), and to City 7 (8 units).
- **From City 7**: To City 0 (9 units), to City 1 (8 units), to City 2 (7 units), to City 3 (9 units), to City 4 (9 units), to City 5 (8 units), and to City 6 (8 units).\\

City Capacity Constraints:
Each city has a maximum capacity for receiving medical supplies:\\

- **City 0**: Can receive up to 19 units.\\
- **City 1**: Can receive up to 15 units.\\
- **City 2**: Can receive up to 15 units.\\
- **City 3**: Can receive up to 14 units.\\
- **City 4**: Can receive up to 15 units.\\
- **City 5**: Can receive up to 15 units.\\
- **City 6**: Can receive up to 14 units.\\
- **City 7**: Can receive up to 16 units.\\

Objective:
Your task is to determine the optimal distribution of medical supplies between the cities to minimize the total transportation cost while ensuring that all cities meet their supply and demand requirements, no route exceeds its capacity, and no city exceeds its receiving capacity.\\

Reported ground truth: 6.\\
Correct ground truth: \textcolor{ForestGreen}{4}. 

\end{descriptionbox}

\subsubsection{Infeasible Problems}

Beyond problems with wrong solutions, we also observe problems that are infeasible. An example is provided below. We fix these problems by appropriately updating the data so that the problem admits a feasible solution.

\begin{descriptionbox}[Example of Infeasible Problem]
\ttfamily\scriptsize

"A university computer lab hires 4 undergraduates (designated 1, 2, 3, and 4) and 2 graduate students (designated 5 and 6) for duty answering questions. The maximum duty hours from Monday to Friday and the hourly wage for each person are shown in Table 5-9.\\

Table 5-9\\
Student ID,Wage (CNY/h),Monday,Tuesday,Wednesday,Thursday,Friday\\
1,10.0,6,0,6,0,7\\
2,10.0,0,6,0,6,0\\
3,9.9,4,8,3,0,5\\
4,9.8,5,5,6,0,4\\
5,10.8,3,0,4,8,0\\
6,11.3,0,6,0,6,3\\

The lab operates from 8:00 AM to 10:00 PM, and there must be one and only one student on duty during open hours. It is also required that each undergraduate must work at least 8 hours per week, and each graduate student must work at least 7 hours per week. Additionally, supplement the following requirements: each student can work no more than 2 shifts per week, and no more than 3 students can be scheduled for duty each day. Based on these conditions, establish a new mathematical model."

\end{descriptionbox}

\begin{descriptionbox}[Example of Fixed Infeasible Problem]
\ttfamily\scriptsize
"A university computer lab hires 4 undergraduates (designated 1, 2, 3, and 4) and 2 graduate students (designated 5 and 6) for duty answering questions. The maximum duty hours from Monday to Friday and the hourly wage for each person are shown in Table 5-9.\\

Table 5-9\\
Student ID | Wage (CNY/h) | Monday | Tuesday | Wednesday | Thursday | Friday\\
1 | 10.0 | 6 | 0 | 6 | 0 | 7\\
2 | 10.0 | 0 | 6 | 0 | 6 | {\textbf{\textcolor{ForestGreen}7}}\\
3 | 9.9 | 4 | 8 | {\textbf{\textcolor{ForestGreen}4}} | 0 | 5\\
4 | 9.8 | 5 | 5 | 6 | 0 | 4\\
5 | 10.8 | {\textbf{\textcolor{ForestGreen}4}} | 0 | 4 | 8 | 0\\
6 | 11.3 | {\textbf{\textcolor{ForestGreen}5}} | 6 | 0 | 6 | 3\\

The lab operates from 8:00 AM to 10:00 PM, and there must be one and only one student on duty during open hours. It is also required that each undergraduate must work at least 8 hours per week, and each graduate student must work at least 7 hours per week. Additionally, each student can work no more than 2 shifts per week, and no more than 3 students can be scheduled for duty each day.\\

Based on these conditions, establish a mathematical model to determine the work schedule that satisfies all requirements."

\end{descriptionbox}

\subsection{Nonlinear Problems in OptMATH}\label{app:nonlinear}

Finally, we study the nonlinear problems in the original OptMATH dataset~\citep{lu2025optmath}. This subset contains 32 problems, including second-order cone programming (SOCP) and quadratically constrained programming (QCP) instances. Their structure differs substantially from the MILP-heavy training distribution targeted by our class-based error analysis and hinting pipeline.

We keep the nonlinear subset separate for two reasons.

First, all second-order test problems in the benchmark are essentially copies of the same synthetic template (see the example below). In addition, these instances suffer from severe missing-data issues: many problem statements mention, for example, the OptMATH problems with index 142, 148, 150 and 154 mentioned ``12 linear constraints'' but only specify parameter values for one or two constraints, so the problems are incomplete and require substantial manual reconstruction by human labelers before they can be meaningfully used. 

Second, when we do run our models on these nonlinear instances, we find that the errors are very concentrated: the solver typically encodes \space $\sum_i y_{k,i}^2 \leq t_k^2$ constraints with $t_k$ unrestricted, making the feasible region a nonconvex union of two cones. This is essentially modeling issue that we believe would be straightforward to fix if similar second-order training examples (with correctly encoded SOCs and linking equations) were available. However, \emph{we do not find any comparable second-order problems of this type in our training data}. This means we cannot derive targeted, class-specific hints from the training corpus to address these errors, in contrast to the MILP classes where our method is designed to operate. 

The closest problem family we see in the available datasets is classical mean--variance portfolio optimization with a quadratic objective and linear constraints, e.g. $$ \begin{aligned} &\min_{w} && \sum_{i=1}^n \sum_{j=1}^n w_i w_j \sigma_{ij} \ &\text{s.t.} && \sum_{i=1}^n w_i r_i \ge R, \ & && \sum_{i=1}^n w_i = 1, \ & && l_i \le w_i \le u_i\ \forall i. \end{aligned} $$ whose description and structure are quite different from the synthetic second-order cone instances in the test set. In practice, hints or error patterns derived from such portfolio problems do not transfer to the benchmark's second-order templates.

\space

\paragraph{Generalization to Nonlinear Formulations.}
As our training corpora contain essentially no SOCP/QCP instances of this type, we seed the two nonlinear hints by sampling three problems from the OptMATH nonlinear test set. We provide the generated hints in the box below. We evaluate the base, SFT, single-turn RL, and multi-turn RL models on 32 held-out nonlinear OptMATH instances, whose SOCP/QCP structures are absent from the training distribution. Table~\ref{tab:optmath_nonlinear_hints} reports results with and without the two nonlinear-specific hints.

\begin{table}[htbp]
\caption{Accuracy (\%) on the held-out nonlinear OptMATH subset. Two nonlinear-specific hints are distilled from three separate seed instances.}
\label{tab:optmath_nonlinear_hints}
\centering
\small
\setlength{\tabcolsep}{5pt}
\begin{tabular}{llcc}
\toprule
\textbf{Model} & \textbf{Hints} & \textbf{1 turn} & \textbf{5 turns} \\
\midrule
GPT-OSS-20B & None & $33.94$ & $61.82 $ \\
GPT-OSS-20B & Nonlinear & $58.79$ & $86.67$ \\
OptiMind-SFT & None & $25.45$ & $49.70$ \\
OptiMind-SFT & Nonlinear & $70.30$ & $89.09$ \\
OptiMind-RL-single & None & 24.24 & 43.18 \\
OptiMind-RL-single & Nonlinear & \textbf{86.74} & 89.39 \\
OptiMind-RL-multi & None & \textbf{39.77} & \textbf{62.50} \\
OptiMind-RL-multi & Nonlinear & 78.03 & \textbf{91.29} \\
\bottomrule
\end{tabular}
\end{table}

The no-hint evaluation measures transfer to an unseen structure, while the seeded-hint evaluation shows that the hint library can be extended to a new family using only a small number of examples. Without hints, multi-turn RL outperforms single-turn RL at both turn budgets. With hints, multi-turn RL initially performs worse, but by five turns surpasses single-turn RL. Multi-turn RL has the highest five-turn mean in both hint settings; single-turn RL is strongest on the first turn with nonlinear hints. The limited template coverage leaves broader nonlinear generalization open.

\begin{descriptionbox}[Automatic Hints Generated from OptMATH Nonlinear Problems]
\ttfamily\scriptsize

  \begin{itemize}
  \item \textbf{Error:} The SOC constraint was encoded as \space $\sum_i y_{k,i}^2 \le t_k^2$
  (equivalently $t_k^2 \ge \sum_i y_{k,i}^2$) while leaving $t_k$ unrestricted in sign,
  which makes the feasible set a nonconvex union of two cones ($\pm t_k$).\\
  \textbf{Hint:} Model the cone in a convex form: use $\lVert y_k\rVert_2 \le t_k$ and enforce $t_k \ge 0$.
  If you keep the squared form, use $\sum_i y_{k,i}^2 \le t_k^2$ \textup{and} explicitly add $t_k \ge 0$;
  do not allow an unrestricted $t_k$ with a squared-radius inequality.

  \item \textbf{Error:} Key mistake: Variables declared \textquotedblleft free/unrestricted\textquotedblright\ in the math description were left at the solver's default $\mathrm{lb}=0$ (nonnegative),
  incorrectly assuming the default is $(-\infty, +\infty)$. This changes the feasible region and can invalidate the SOCP.\\
  \textbf{Hint:} When the formulation says a continuous variable is free, explicitly set $\mathrm{lb}=-\infty$, $\mathrm{ub}=+\infty$
  (or represent $v$ as $v^+ - v^-$ with $v^+, v^- \ge 0$). Also carefully map the coefficient blocks into $Ax=b$
  with the correct RHS per row so the linear equalities match the intended model.
\end{itemize}

\end{descriptionbox}

\begin{descriptionbox}[Example of Non-Linear Problem]
\ttfamily\scriptsize
Minimize the objective function with cost coefficients -0.1919592146476727 for x[0], -1.473647303839492 for x[1], and 2.304735407761341 for x[2]. The problem is subject to three strict equality constraints: (1) -1.635895473616174 x[0] - 0.3973211001807447 x[1] + 0.9471007364101932 x[2] must equal -1.151224267166901, labeled as linear\_eq[0]; (2) -0.5249535603075616 x[0] + 0.3668073807989349 x[1] - 0.7858336216136411 x[2] must equal 0.2111440590441619, labeled as linear\_eq[1]; and (3) -1.91474276776438 x[0] - 0.0172618223950067 x[1] - 0.4534063203075578 x[2] must equal -0.0637133383926026, labeled as linear\_eq[2].\\

Additionally, four auxiliary variables y\_0[0], y\_0[1], y\_0[2], and y\_0[3] are introduced, each subject to linear constraints: (1) -0.1884394252535835 x[0] + 0.0177104249784969 x[1] + 0.0986130842848287 x[2] + y\_0[0] must equal 0.0043955531724021, labeled as R3; (2) -0.0708818331240574 x[0] - 0.0215819660876361 x[1] - 0.0616654799104094 x[2] + y\_0[1] must equal 0.0879019749446116, labeled as R4; (3) -0.0907143884234421 x[0] + 0.0750697315409076 x[1] - 0.1169779589661307 x[2] + y\_0[2] must equal 0.0357726532121181, labeled as R5; and (4) -0.1820309484664378 x[0] + 0.0261698752429371 x[1] + 0.0667294286775777 x[2] + y\_0[3] must equal 0.0118088378651355, labeled as R6.\\

A decision variable t\_0[0] is introduced, subject to the constraint -0.2396501315769039 x[0] + 0.3255209569765619 x[1] + 0.2624492403208943 x[2] + t\_0[0] must equal 1.588464205354944, labeled as R7. A second-order cone constraint labeled as qc0 is imposed, ensuring that the sum of the squares of y\_0[0], y\_0[1], y\_0[2], and y\_0[3] does not exceed the square of t\_0[0], expressed as - t\_0[0]\^{}2 + y\_0[0]\^{}2 + y\_0[1]\^{}2 + y\_0[2]\^{}2 + y\_0[3]\^{}2 <= 0. All decision variables, including x[0], x[1], x[2], y\_0[0], y\_0[1], y\_0[2], y\_0[3], and t\_0[0], are free to take any real value. The goal is to determine the optimal values for these variables to minimize the objective function while satisfying all constraints. This is a Second-Order Cone Programming (SOCP) problem."
\end{descriptionbox}

\clearpage

\section{Experiment Details}
\subsection{Training Details}\label{app:training-details}

We perform supervised fine-tuning (SFT) to adapt the base language model to our formulation workflow, i.e., mapping a natural-language problem description to a complete solution that includes a mathematical model and executable solver code. 
In more formal terms, let the SFT dataset be $\mathcal{D}_{\mathrm{SFT}} = {(x_i,y_i)}_{i=1}^{N}$, where $x_i$ is the problem description and $y_i$ is the completion sequence formed by concatenating the model's thinking tokens, the mathematical formulation, and the solver code. We train with standard sequence-to-sequence loss: $\mathcal{L}_{\mathrm{SFT}}(\theta)= -\mathbb{E}_{(x,y)\sim \mathcal{D}_{\mathrm{SFT}}} \sum_{t=1}^{|y|}\log p_\theta \bigl(y_t \mid y_{<t}, x\bigr)$, where $\theta$ denotes all trainable parameters of the model; $p_\theta(\cdot)$ is the model's output distribution; $y_t$ is the target token at position $t$; and $y_{<t} := (y_1,\ldots,y_{t-1})$ are the prefix of previously generated target tokens.

We performed SFT on a single node with eight NVIDIA HGX B200 GPUs. All solution generation in the data-cleaning pipeline and evaluation tasks were run on 4 compute nodes, each with eight 80GB NVIDIA H100 GPUs.
The data-cleaning and regeneration budget scales with dataset size, decoding rounds, and maximum generation length; in our setting this is roughly $4{,}500$ training instances $\times$ 64 sampled solutions $\times$ 60{,}000 maximum tokens. Test-time inference scales as number of test instances $\times$ turns $\times$ maximum generation length, i.e., about $439 \times T \times 60{,}000$ maximum tokens for $T$ correction turns, with an additional multiplicative factor of $K$ when majority voting with $K$ samples is used.
We use gpt-oss-20B as the base model and adapt the \textsc{Verl} framework for SFT. Concretely, we train from the \texttt{unsloth/openai-gpt-oss-20b-BFloat16} variant to avoid MXFP4 precision pitfalls and to leverage Unsloth's compatibility fixes for gpt-oss, and we use FSDP2 to distribute training across the eight GPUs. We tune the learning rate over $[10^{-6}, 4\times 10^{-5}]$ via a grid search, and select the best configuration on the validation split under the first-turn, no-hint setting. Key SFT hyperparameters and optimizer settings are listed in Table~\ref{tab:sft_params}. 

\begin{table}[ht]
\caption{SFT training parameters for \textsc{gpt-oss-20b}.} \label{tab:sft_params}
\begin{center}
\begin{tabular}{l c}
\textbf{Parameter} & \textbf{Value} \\ 
\hline \\[-0.8em]
Batch size & 16 \\
Learning rate & 2e-5 \\
Max prompt length & 16384 \\
Max response length & 16384 \\
Prompt length truncation & left \\
Max epochs & 3
\end{tabular}
\end{center}
\end{table}

For RL training of \textsc{gpt-oss-20b}, we use GRPO on the cleaned reward environment described in \S\ref{sec:error_analysis}--\S\ref{sec:training_rl}. RL is trained separately from SFT. The multi-turn variant uses the first-turn prompt followed by the self-correction feedback prompt for a total of three turns; after each turn we execute the latest program and feed the resulting stdout/stderr back to the model. The trajectory reward is the average binary printed-answer correctness over the three turns: each candidate receives correctness 1 if its printed objective matches any ground-truth objective value in the cleaned dataset within the evaluation tolerance, and 0 otherwise. The implementation uses \textsc{Slime}; detailed hyperparameters are listed in Table~\ref{tab:rl_params}.

\begin{table}[ht]
\caption{GRPO training parameters for \textsc{gpt-oss-20b}.} \label{tab:rl_params}
\centering
\begin{tabular}{@{}>{\raggedright\arraybackslash}p{0.40\linewidth}>{\raggedright\arraybackslash}p{\dimexpr0.60\linewidth-2\tabcolsep\relax}@{}}
\toprule
\multicolumn{1}{c}{\bf Parameter}  &\multicolumn{1}{c}{\bf Value}
\\ \midrule
Algorithm & GRPO \\
Initialization & \texttt{gpt-oss-20b-bf16} \\
Rollout turns & 3 \\
GRPO group size & 2 \\
Samples per prompt & 8 \\
Prompts per step & 32 \\
Global batch size & 240 \\
KL coefficient & 0.001 \\
Learning rate & $5\times 10^{-7}$ with cosine decay and 1\% warmup \\
Optimizer & Adam ($\beta_1{=}0.9$, $\beta_2{=}0.99$, $\epsilon{=}10^{-8}$) \\
Clip range / gradient clip & 0.2 / 1.0 \\
GRPO iterations per batch & 2 \\
Logged training steps & 4{,}792 \\
Rollout temperature / top-p & 0.9 / 1.0 \\
Rollout max response length (per turn) & 4{,}096 tokens \\
Hardware budget & 2 $\times$ 8 Nvidia B200 \\
\bottomrule
\end{tabular}
\end{table}

\paragraph{Qwen3.5-9B training.}
Qwen3.5-9B SFT uses trajectories from Qwen3.5-122B-A10B filtered to match the ground-truth objective values in the cleaned training corpus, while RL reuses the cleaned instances and ground-truth objective values without new teacher trajectories (Appendix~\ref{app:qwen_transfer}). Table~\ref{tab:qwen_sft_params} gives the SFT hyperparameters. We train the Qwen base model with GRPO in \textsc{Slime}, using three-turn solver-feedback rollouts without early stopping. The base reward is final-turn binary correctness, with DAPO-style soft overlong reward shaping and dynamic sampling of groups with nonzero reward variance. Table~\ref{tab:qwen_rl_params} gives the RL configuration; evaluation settings are reported in Appendix~\ref{app:qwen_transfer}.

\begin{table}[ht]
\caption{SFT training parameters for Qwen3.5-9B.} \label{tab:qwen_sft_params}
\begin{center}
\begin{tabular}{l c}
\textbf{Parameter} & \textbf{Value} \\
\hline \\[-0.8em]
Batch size & 32 \\
Learning rate & 1e-5 \\
Max sequence length & 32768 \\
Sequence length truncation & left \\
Max epochs & 3
\end{tabular}
\end{center}
\end{table}

\begin{table}[ht]
\caption{GRPO training parameters for Qwen3.5-9B.}
\label{tab:qwen_rl_params}
\centering
\small
\begin{tabular}{@{}>{\raggedright\arraybackslash}p{0.40\linewidth}>{\raggedright\arraybackslash}p{\dimexpr0.60\linewidth-2\tabcolsep\relax}@{}}
\toprule
\multicolumn{1}{c}{\bf Parameter} & \multicolumn{1}{c}{\bf Value}
\\ \midrule
Algorithm & GRPO \\
Initialization / precision & Qwen3.5-9B / BF16 \\
Rollout turns & 3 (early stopping disabled) \\
Samples per prompt & 8 \\
Prompts per rollout & 32 \\
Global / micro batch size & 256 / 1 \\
KL coefficient / entropy coefficient & 0.001 / 0 \\
Learning rate & $5\times 10^{-7}$ with cosine decay and 1\% warmup \\
Optimizer & Adam ($\beta_1{=}0.9$, $\beta_2{=}0.99$, $\epsilon{=}10^{-8}$) \\
Weight decay & 0.01 \\
Clip range / gradient clip & 0.2 / 1.0 \\
Optimization steps per rollout & 1 \\
Configured rollout iterations & 1{,}000 \\
Rollout temperature / top-$p$ & 1.0 / 1.0 \\
Rollout max response length & 24{,}576 tokens \\
Actor total token budget & 65{,}536 tokens \\
Tensor / pipeline parallel size & 4 / 1 \\
Hardware budget & 2 $\times$ 8 NVIDIA B200 \\
\bottomrule
\end{tabular}
\end{table}

For solution generation in data cleaning, we host \texttt{openai/gpt-oss-20b} with \textsc{vLLM}. For evaluation of the GPT-OSS models, we serve all fine-tuned checkpoints and the BF16 base (\texttt{unsloth/openai-gpt-oss-20b-BFloat16}) via \textsc{SGLang}, since \textsc{vLLM} did not support our Unsloth-adapted variant when we conducted the evaluation. We use top-p decoding for all generation and inference.
We tune the sampling temperature over $[0.5, 1.0]$ via a grid search, and select the best configuration on the validation split under the first-turn, no-hint setting.
The common sampling settings for our GPT-OSS SFT model are summarized in Table~\ref{tab:gen_params}. 

\begin{table}[t]
\caption{Generation / evaluation sampling parameters for the \textsc{gpt-oss-20b} SFT model.} \label{tab:gen_params}
\begin{center}
\begin{tabular}{ll}
\multicolumn{1}{c}{\bf Parameter}  &\multicolumn{1}{c}{\bf Value}
\\ \hline \\[-0.8em]
Temperature & 0.9 \\
Top p & 0.95 \\
Max tokens & 60000 
\end{tabular}
\end{center}
\end{table}

\subsection{Evaluation Experiment Details}
\label{app:evaluation_experiment_details}

\paragraph{Benchmark selection.}
Our main evaluation uses the cleaned versions of \textbf{IndustryOR}~\citep{tang2024orlm},
\textbf{Mamo-Complex}~\citep{huang2024mamo}, and \textbf{OptMATH}~\citep{lu2025optmath}. Each
contains roughly 100--160 problems, and together they provide 469 carefully validated optimization
instances. Although the cleaned benchmarks are modest in size, we prioritize correctness and
transparent cleaning over raw scale because the task is highly sensitive to missing data, ambiguous
wording, infeasible instances, and incorrect ground-truth objective values. Cleaning details are provided in
Appendix~\ref{app:benchmark_cleaning}.

\space

For completeness, we additionally report results on cleaned versions of IndustryOR released by
SIRL~\citep{chen2025solver}, LogiOR~\citep{logior}, and \citet{survey}, as well as cleaned versions
of Mamo-Complex from SIRL and \citet{survey}. We also evaluate the arXiv dataset to test out-of-taxonomy generalization:
on this dataset, \textsc{OptiMind-SFT}+hints improves first-turn accuracy from $30.00\%$ to $38.75\%$;
see Appendix~\ref{app:ood_generalization}. We additionally evaluate the standalone LogiOR benchmark, whose mathematical formulations allow us to check the ground-truth objective values with a solver (Appendix~\ref{app:standalone_logior}).

\paragraph{Objective extraction and correctness.}
All evaluated models are prompted to generate executable Python code using GurobiPy. To extract the
model's predicted objective value, we insert a print statement with a unique tag and parse the tagged
objective from the execution log, following the protocol of SIRL~\citep{chen2025solver}. A generated
solution is considered correct if the parsed objective value $\hat{z}$ satisfies
\[
    \frac{|\hat{z}-z^\star|}{|z^\star|+1} < 10^{-6},
\]
where $z^\star$ is a ground-truth objective value in the cleaned dataset. If multiple interpretations are
considered valid for a problem, a prediction is correct if it matches any of the corresponding
ground-truth objective values. We use objective-value matching rather than code-level comparison because many
syntactically different formulations can represent the same optimization model.

\paragraph{Multi-turn correction, majority voting, and repeated runs.}
We report accuracy after the first generation turn and after five turns of iterative self-correction.
In each self-correction turn, the previously generated GurobiPy code is executed and the resulting
solver output is returned to the model. The model may then revise its formulation or keep the previous
candidate if it judges the solution to be correct.

For the base and SFT checkpoints, we also evaluate self-consistency by comparing single-sample
decoding ($K=1$) with majority voting over $K=8$ sampled solutions. Majority voting groups
candidate objective values within relative and absolute tolerance $10^{-6}$ to account for
floating-point variation. \space For the main comparisons in Table~\ref{tab:main_results_comprehensive}, non-RL results are averaged over
10 independent evaluations using different random seeds.

Run counts for the additional evaluations are specified with the respective tables in Appendix~\ref{app:additional_eval_results} where available.
The model comparisons use $K=1$; majority voting in benchmark-accuracy evaluation is confined to the test-time-scaling study. The $K=8$ GPT-5 consensus used to generate automatic hints and the $K=64$ solution-generation consensus are separate data-construction procedures.

\paragraph{Baselines.}
We compare against open-source models of similar or larger scale, including \textsc{gpt-oss-20B}
without fine-tuning, \textsc{Qwen3-32B}, and \textsc{SIRL-Gurobi32B}~\citep{chen2025solver}, a
specialized optimization-modeling model that has been considered state-of-the-art at its size. We also
include frontier models \textsc{o4-mini} and \textsc{GPT-5}. We were unable to reliably reproduce the
reported performance of several other open-source optimization-modeling systems, such as OptMATH
and LLMOpt, so we include the strongest small open-source baselines that we could evaluate under
our protocol.

\paragraph{Qwen-based baseline decoding.}
We evaluate Qwen3-32B with temperature $0.6$, top-$p=0.95$, and one sample per turn ($K=1$), for up to five turns without early stopping. For both Qwen3-32B and its derivative SIRL-Gurobi-32B, we use a hard context window of 32{,}768 tokens and a combined prompt-plus-completion limit of 32{,}700 tokens, truncating the earliest conversation history in longer interactions. Given the current prompt length $L_{\mathrm{prompt}}$, we dynamically set
\[
\texttt{max\_new\_tokens}
= \max\!\left(1,\min\!\left(32{,}700-L_{\mathrm{prompt}},\;32{,}768-L_{\mathrm{prompt}}\right)\right).
\]\par
\space

\clearpage
\section{Details of Inference}\label{app:inference_details}

\subsection{Problem Classification Prompt}\label{app:problem_classification}

For problem classification, we use all 53 seed classes from the OptMATH dataset, finding that these categories are diverse and cover a comprehensive proportion of classes in our training set. Furthermore, we obtain corresponding natural language examples for each problem category from the OptMATH repository. \cite{lu2025optmath};
Table \ref{optimization_subclasses} contains the 53 seed class names, and Table \ref{tab:subclass_examples} shows examples of corresponding examples. Some high-level class names appear more than once because they correspond to distinct generator templates or application variants in the original taxonomy. When prompting the LLM to classify the problem, we provide all problem categories and natural language examples to assist the language model. 

\tcbset{
  colback=gray!10,    colframe=gray!80,   boxrule=0.5pt,
  arc=2pt,
  outer arc=2pt,
  left=6pt,
  right=6pt,
  top=6pt,
  bottom=6pt,
}

\begin{promptbox}[Problem Classification Prompt]
\ttfamily\footnotesize
You are a helpful Assistant with expertise in mathematical modeling 
and the Gurobi solver. You will be given a natural language question. 
Please classify the question into one or more of the following categories:
\\
\textcolor{blue}{\{list of categories\}} \\ \\
If the question does not fit any of the above categories, please provide 
a label for the Category that best describes the problem type.
\\ \\
Category examples: \\
\textcolor{blue}{\{natual language example of each category\}}
\\ \\
Please classify the following question into one or more categories: \\
\textcolor{blue}{\{question\} }
\\
Return only the categories in a python list, without any additional text.
\end{promptbox}

\begin{table}[ht]
\centering
\caption{Problem categories obtained from OptMATH}
\label{optimization_subclasses}
\renewcommand{\arraystretch}{1.2} \begin{tabular}{@{}*{3}{m{\dimexpr(\linewidth-4\tabcolsep)/3\relax}}@{}}
\toprule
\centering Aircraft Assignment & \centering Aircraft Landing & \centering Bin Packing \tabularnewline
\centering Blending Problem & \centering Capacitated Facility Location Problem & \centering Capacitated Lot-sizing Problem (CLSP) \tabularnewline
\centering Car Selection Assignment & \centering Contract Allocation & \centering Diet Problem \tabularnewline
\centering Factory Planning Problem & \centering Flow Shop Scheduling & \centering Job Shop \tabularnewline
\centering Knapsack & \centering Multicommodity Capacitated Network Design & \centering MarketShare \tabularnewline
\centering Set Multi-Cover & \centering PortfolioOptimization & \centering Revenue Management Problem \tabularnewline
\centering Assignment Problem & \centering Set Cover & \centering Discrete Lot-sizing and Scheduling Problem \tabularnewline
\centering Static Line Planning & \centering Structure Based Assignment Problem & \centering SupplyChain \tabularnewline
\centering TravelingSalesman & \centering Facility Dispersion Problem & \centering Military Personnel Deployment Problem \tabularnewline
\centering Production Planning Problem & \centering Facility Dispersion Problem & \centering Network Optimization \tabularnewline
\centering Lot-Sizing Problem & \centering Operations Optimization & \centering Capacitated Vehicle Routing Problem with Time Windows (CVRPTW) \tabularnewline
\centering Facility Location Problem & \centering Cutting Stock Problem & \centering Unit Commitment Problem \tabularnewline
\centering Farm Planning & \centering Transportation, Airline Industry, Resource Allocation & \centering Multi-Commodity Transportation Problem \tabularnewline
\centering Minimum Cost Flow Problem & \centering Assignment Problem & \centering Multi-Commodity Network Flow Problem \tabularnewline
\centering Transportation Problem & \centering Profit Maximization Problem & \centering Revenue Maximization Problem \tabularnewline
\centering Facility Location Problem & \centering Production Planning Problem & \centering Team Formulation Problem \tabularnewline
\centering Transportation Problem & & \tabularnewline
\bottomrule
\end{tabular}
\end{table}

\begin{table}[ht]
\centering
\caption{Examples of OptMATH classes and their corresponding natural language examples. Original generators: \url{https://github.com/optsuite/OptMATH/tree/main/generators}.}
\label{tab:subclass_examples}
\renewcommand{\arraystretch}{1.2}
\begin{tabular}{@{}>{\raggedright\arraybackslash}p{0.25\linewidth}>{\raggedright\arraybackslash}p{\dimexpr0.75\linewidth-2\tabcolsep\relax}@{}}
\toprule
\textbf{Subclass} & \textbf{Natural Language Example} \\
\midrule
Knapsack & A hiker is preparing for a 3-day outdoor hiking trip. They need to select the most valuable combination of equipment and supplies from many available options within the limited backpack capacity. The items include:\newline* Tent: weight 5kg, value 90 points\newline* Sleeping bag: weight 3kg, value 80 points* Portable stove: weight 1kg, value 60 points\newline* Drinking water: weight 2kg, value 70 points\newline* Dry food: weight 1.5kg, value 65 points\newline* First aid kit: weight 0.5kg, value 50 points\newline Assuming the backpack has a maximum weight capacity of 10kg, the hiker's goal is to select the combination of items with the highest total value while not exceeding the weight limit. Each item must be either taken in its entirety or left behind; partial items cannot be taken.
 \\
\midrule
Team Formulation Problem & As a team leader, you have several projects requiring different expertise levels in different areas. You want to assign a set of people to these projects. Each person has a level of expertise in each of the areas. How to assign people to projects to minimize the maximum skill shortage from the project expertise requirement? \\
\midrule
Transportation Problem & A company has three warehouses (origins) with supplies of 100, 150, and 200 units, respectively. There are four retail stores (destinations) with demands of 80, 120, 90, and 160 units, respectively. The cost of transporting one unit from each warehouse to each retail store is given. The goal is to determine the optimal transportation plan that minimizes the total transportation cost while meeting all supply and demand constraints. \\
\bottomrule
\end{tabular}
\end{table}

\newpage
\subsection{Multi-Turn Inference Prompts}\label{app:inference-prompts}

This appendix contains the inference details summarized in Section~\ref{sec:inference}. At the first turn, the model receives the original problem statement and, optionally, class-conditioned error-analysis hints. The hints are presented as a conditional checklist: the model is instructed to inspect the problem statement, determine which hints are applicable, and apply only those hints. This is important because a class-level hint may apply only to a subset of instances. For example, a backlog-balance hint should be used only when backorders are explicitly allowed; otherwise, adding backlog variables changes the intended optimization problem.

After the model produces a GurobiPy program, we execute the latest code block and return the solver output to the model during self-correction turns. The feedback includes stdout, stderr, solver status, logs, and the printed objective value when available. If the previous program is correct, the model is not required to produce a new code block; if it detects an error, it can revise the formulation. Majority voting can be applied at the first turn or after each correction turn by grouping candidate objective values within the evaluation tolerance.

\begin{algorithm}[tb]
\caption{\textsc{MultiTurnInferencewithMajorityVoting}}
  \label{alg:multi-turn-inference}
  \begin{algorithmic}
\STATE \textbf{Require:} problem instance $Q$, error-analysis library $H$ (maps problem type $\to$ list of (error, hint) pairs), LLM Generator $G$, majority voting number $K$, number of correction rounds $M$.
\STATE $t \gets \textsc{ClassifyType}_G(Q)$ \COMMENT{predict problem type}
\STATE $\texttt{hints} \gets H[t]$ \COMMENT{retrieve hints of corresponding type}
\STATE $\big(s_1^{(0)}, \cdots s_K^{(0)} \big) \gets \textsc{FirstTurnGenerate}_G(Q, \texttt{hints}, K)$ \COMMENT{generate $K$ solutions}
\STATE $a^{(0)}, \texttt{stdout}^{(0)}, \texttt{stderr}^{(0)} \gets \textsc{GetMajorityResults}\big(s_1^{(0)}, \cdots s_K^{(0)} \big)$
\STATE
\FOR{$m=1$ to $M-1$}
\STATE $\big(s_1^{(m)}, \cdots s_K^{(m)}  \big) \gets \textsc{SelfCorrectionGenerate}_G(Q, \texttt{stdout}^{(m-1)}, \texttt{stderr}^{(m-1)}, K)$
\COMMENT{get $K$ self-correction responses}
\STATE $a^{(m)}, \texttt{stdout}^{(m)}, \texttt{stderr}^{(m)} \gets \textsc{GetMajorityResults}\big(s_1^{(m)}, \cdots s_K^{(m)}  \big)$
\ENDFOR
\STATE \textbf{Return} $\hat{a}^{(M-1)}$  \COMMENT{Return the final result}
  \end{algorithmic}
\end{algorithm}

\begin{algorithm}[tb]
\caption{\textsc{GetMajorityResults}($s_1, \cdots s_K  $)}
\label{alg:majority}
\begin{algorithmic}
\STATE \textbf{Require:} Solution trajectories $s_1, \cdots s_K $
\FOR{$k = 1$ to $K$}
    \STATE $\big( a_k, \texttt{stdout}_k, \texttt{stderr}_k \big) \gets \textsc{ExtractAnswer}(s_k)$ \COMMENT{Extract code and get system output for each solution}
\ENDFOR
\STATE $\hat{k} \gets \textsc{GetMajorityVoteIndex}\big(a_1, \cdots, a_K\big)$ 
\STATE \textbf{Return} $\big(a_{\hat{k}},  \texttt{stdout}_{\hat{k}}, \texttt{stderr}_{\hat{k}}  \big)$
\end{algorithmic}
\end{algorithm}

We use the following prompts during our multi-turn inference. At the first turn, we prompt the LLM with the question and asks for the corresponding gurobipy code. Optionally, we may include an error-analysis hint for the problem's category, which we find generally improve the performance. Algorithm \ref{alg:multi-turn-inference} shows a pseudocode of our inference strategy.

\begin{promptbox}[First Turn User Prompt (Not Using Error Analysis Hint)]
\ttfamily\footnotesize
You are an expert in optimization and mixed integer programming. You are given an optimization problem and you need to solve it using gurobipy.\\

\textcolor{blue}{\{question\}}\\

Reason step by step before generating the gurobipy code.

When you respond, first think carefully.

After thinking, output the math modeling of the problem.

Finally output a ```python ...``` code block that solves the problem.\\

The code must include:

import gurobipy as gp \\
from gurobipy import GRB

\end{promptbox}

\begin{promptbox}[First Turn User Prompt (Using Error Analysis Hint)]
\ttfamily\footnotesize
You are an expert in optimization and mixed integer programming. You are given an optimization problem and you need to solve it using gurobipy.\\

\textcolor{blue}{\{question\}}\\

Below are hints for avoiding common mistakes often seen for this problem type. Avoid them if applicable. \\

Error analysis \#1: \textcolor{blue}{\{Error Summary 1\}, \{Hint 1\}}

...

Error analysis \#k: \textcolor{blue}{\{Error Summary k\}, \{Hint k\}} \\

Instructions for applying error-analysis hints: \\
- Review the provided hints and identify which ones are applicable to this problem. \\
- Please apply ALL applicable hints. \\
- Before applying any hint or writing constraints, check the sign and direction of every variable and coefficient for consistency (e.g., profit = revenue - cost; capacities as <= flow conservation as =). \\

General modeling checklist (follow rigorously): \\
- Units: Use correct units everywhere and ensure the objective's units match the goal (e.g., dollars for cost/profit, distance for TSP, time for scheduling). Do not mix units (e.g., minutes with hours, dollars with 1000 dollars) without converting. \\

Reason step by step before generating the gurobipy code.

When you respond, first think carefully.

After thinking, output the math modeling of the problem.

Finally output a ```python ...``` code block that solves the problem.\\

The code must include:

import gurobipy as gp \\
from gurobipy import GRB

\end{promptbox}

From the LLM's response, we extract and execute the gurobipy code to obtain standard output (STDOUT) and standard error (STDERR). Our multi-turn pipeline then feeds this execution feedback back to the LLM, prompting it to self-correct its previous solution, as shown below.

\begin{promptbox}[Self-Correction Feedback Prompt]
\ttfamily\footnotesize
Running the given gurobipy code block result in the following standard output and error.  \\
Please analyze the standard output and error and provide your thoughts on the correctness of the code and the output, and if it is not correct, provide a corrected version of the code. \\

The standard output and error message (if any) for your generated code are as follows: \\

[STDOUT] \textcolor{blue}{\{Standard Output of the gurobipy program (gurobipy log)\}} \\

[STDERR] \textcolor{blue}{\{Standard Error of the gurobipy program (if exists)\}} \\

Based on the output and error messa   ge (if any), please think step by step in the <think> ... </think> block to determine if the code is correct.  \\

Based on your thoughts, please provide an (improved) gurobipy code in the ```python\verb|\n|``` code block if you have identified mistakes in the previous code. \\

If the previous code is correct, you do not need to provide a new gurobi code block again. Your performance will be evaluated based on the correctness of the final ```python\verb|\n|``` code block. \\

\verb|#| Note: \\
- The Code must include:```python
import gurobipy as gp
from gurobipy import GRB
```\\
- Make sure the model variable is named `model`.\\
- Avoid using "<" and ">" in Gurobi constraints; instead, use "<=" or ">=".\\
- Carefully determine whether the variable is an integer or continuous. \\
\end{promptbox}

\subsection{Example of the Effect of Multi-Turn Inference}\label{app:multiTurnEx}

In this production optimization instance from \textbf{IndustryOR}, a factory has 15,000 cubic meters of internal warehouse capacity and can rent external storage at an additional cost. The first-turn program returns an optimal solution, but its printed decisions reveal that external storage does not cover inventory exceeding internal capacity.

The first turn of the inference produced a model with the \textbf{incorrect} constraint {\color{red} ``\verb|e[t] <= vol - wh_vol_cap|''}, where \verb|e[t]| is a decision variable choosing how many cubic meters of external warehouse to use in period $t$, \verb|vol| denotes the total volume of (internal + external) warehouse used, and \verb|wh_vol_cap| $= 15,000$ is the internal warehouse capacity. Notice that this constraint is incorrect because it allows the volume used \verb|vol| to be bigger than the capacity \verb|wh_vol_cap| while having external warehouse usage \verb|e[t]| $=0$. The correct inequality is in the opposite direction, namely ``{\color{blue} \verb|e[t] >= vol - wh_vol_cap|}'', which forces that the external warehouse usage to cover for any excess over the internal warehouse capacity. 

After inspecting the first-turn formulation and the solver's decisions, the model identifies and reverses the incorrect inequality in its second response. The relevant reasoning is shown below.

\begin{descriptionbox}[Thought of the second turn of inference (snippet)]
\ttfamily \small
Volume per month: compute volume each month. July: 0.2*45k + 0.4*15k = 9000 + 6000 = 15000 cubic meters. That equals capacity. So no external. August: 0.2*75k + 0.4*0 = 15000. OK. September: 0.2*127.5k + 0.4*0 = 25500. \textbf{Exceeds 15000 by 10500}. But model forced \verb|e_t <= vol - wh_vol_cap|. So \verb|e_t| would be 10500. But output shows \verb|e_t| 0. So maybe \verb|e_t| not used? Let's check: The constraint {\color{red} \verb|e[t] <= vol - wh_vol_cap|}. For September \verb|vol|=25500, \verb|wh_vol_cap|=15000 => RHS=10500. So \verb|e_t| <= 10500. But objective includes 0.5*\verb|e_t|. To minimize cost, \verb|e_t| should be 0. But we didn't enforce \verb|e_t >= vol - wh_vol_cap|? No, we only enforced \verb|e_t <= vol - wh_vol_cap|. That allows \verb|e_t| to be 0 even if \verb|vol > wh_vol_cap|. But we need to enforce that external volume equals excess volume. \textbf{We used inequality incorrectly}. We should enforce {\color{blue} \verb|e_t >= vol - wh_vol_cap|}, not <=. 
\end{descriptionbox}

A solution can be optimal for the generated model while violating the problem statement. Here, the printed decisions reveal such a violation and help the model identify the incorrect constraint. Including this feedback in RL rollouts trains the model to use solver outputs when revising its formulation. Appendix~\ref{app:rl-trajectories} illustrates RL trajectories for semantic repair, infeasibility diagnosis, and preservation of correct formulations.

\newpage
\section{ Code, Data, and Model Release}
\label{app:anonymous_repo}

We release the full hint library, cleaned test data, cleaning logs, and evaluation code at \url{https://github.com/microsoft/OptiGuide/tree/main/optimind}. The cleaning logs include benchmark annotations and IndustryOR/OptMATH correction tables; the evaluation code includes inference scripts. We release our trained OptiMind-RL model at \url{https://huggingface.co/microsoft/optimind-rl}.

Below, we describe the cleaned test data and cleaning logs included in the release.

\subsection{Cleaned Test Data}

The expert--cleaned test files are available in the \href{https://github.com/microsoft/OptiGuide/tree/main/optimind/data}{repository data directory}:
\begin{itemize}
    \item \href{https://github.com/microsoft/OptiGuide/blob/main/optimind/data/optimind_cleaned_classified_industryor.csv}{\texttt{optimind\_cleaned\_classified\_industryor.csv}} (99 items), 
    \item \href{https://github.com/microsoft/OptiGuide/blob/main/optimind/data/optimind_cleaned_classified_mamo_complex.csv}{\texttt{optimind\_cleaned\_classified\_mamo\_complex.csv}} (210 items), and
    \item \href{https://github.com/microsoft/OptiGuide/blob/main/optimind/data/optimind_cleaned_classified_optmath.csv}{\texttt{optimind\_cleaned\_classified\_optmath.csv}} (130 items),
\end{itemize}
Each CSV is UTF--8 encoded
and contains three columns: \texttt{problem\_class} (the assigned problem category),
\texttt{question} (the complete natural--language problem statement), and
\texttt{answer} (the ground--truth optimal objective value after cleaning).

The \texttt{answer} field is either (i) a single numeric value, or (ii) a JSON array of numeric values
when multiple interpretations are reasonable (e.g., integer vs.\ fractional formulations) and thus
multiple ground-truth objective values are recorded. During evaluation, a prediction is considered correct if it matches
\emph{any} provided value within absolute and relative tolerances of $10^{-6}$.

These files reflect our corrections for missing parameters, ambiguity, infeasibility, wrong reference
answers, and scope mismatches identified during manual review. The sets are intended strictly for
\emph{evaluation} (not training) to ensure comparability across studies. We will update the archive if
community feedback identifies additional issues; any changes will be documented with version notes.

\subsection{Cleaning Logs}\label{app:supp-data-correction}

The cleaning logs are available in the \href{https://github.com/microsoft/OptiGuide/tree/main/optimind/data/comparisons}{repository correction-table directory}.

\begin{figure}[ht]
\begin{center}
\includegraphics[width=0.9\linewidth]{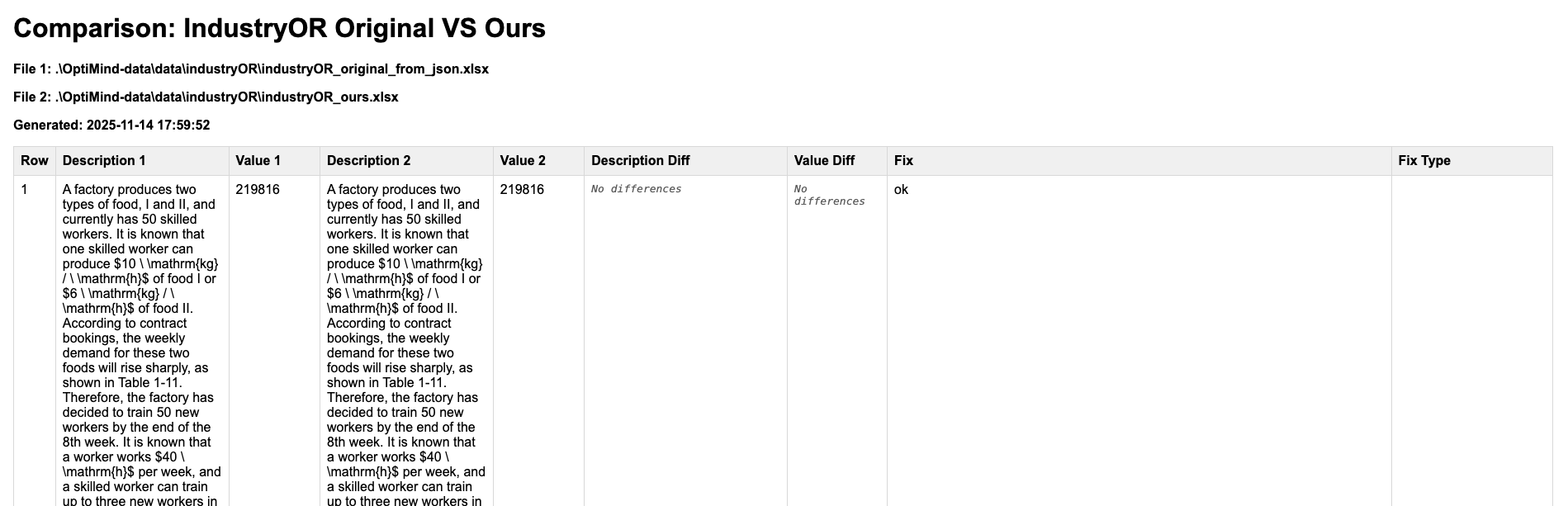}  
\includegraphics[width=0.9\linewidth]{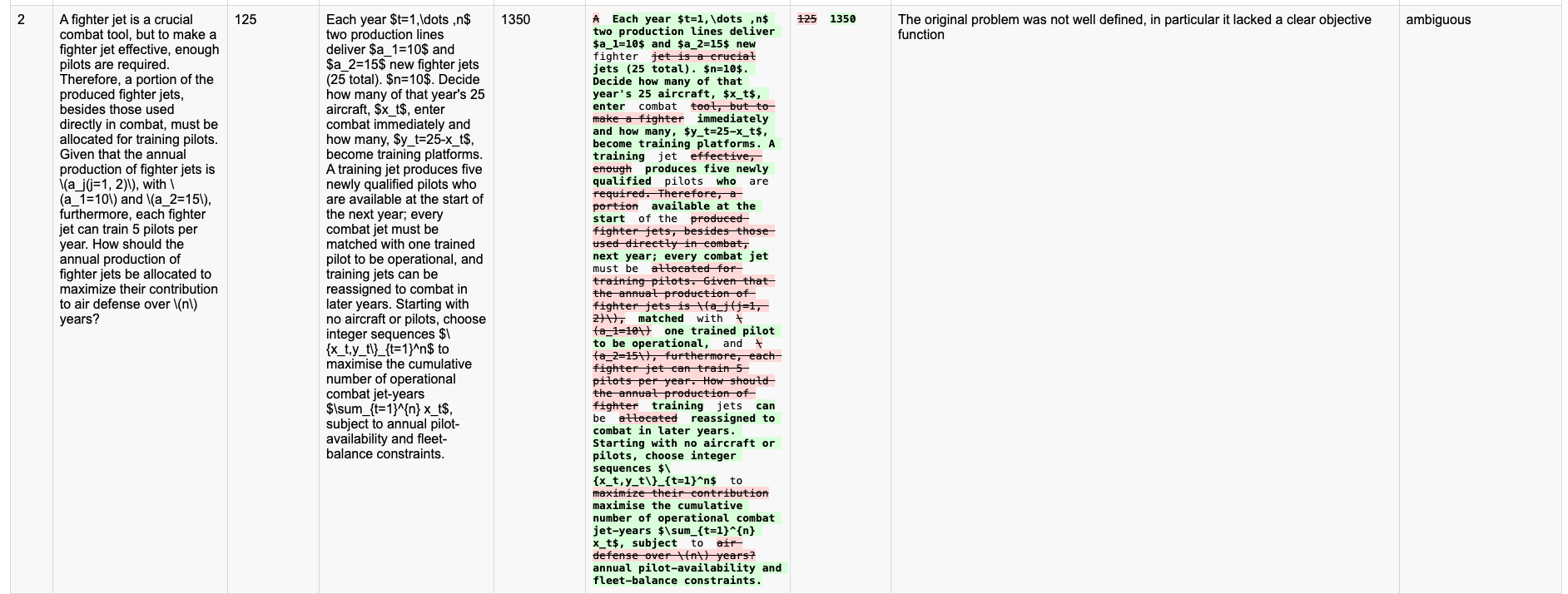}  
\end{center}
\caption{A screenshot of the HTML correction table in our supplementary material showing our edits for the IndustryOR benchmark.}
\label{fig:screenshot}
\end{figure}

In addition to the cleaned CSV test sets, for the IndustryOR and OptMATH benchmarks the cleaning logs include HTML summary tables that list each correction and the reason for the change. Each HTML file contains a table with columns [Problem Index, Original Problem, Original Answer, Updated Problem, Updated Answer, How did we fix it]. Figure~\ref{fig:screenshot} shows a screenshot of one such table, illustrating the layout and content of the HTML files. The cleaning logs also include the comparison files \href{https://github.com/microsoft/OptiGuide/blob/main/optimind/data/comparisons/industryOR_original_vs_ours.html}{\texttt{industryOR\_original\_vs\_ours.html}} and \href{https://github.com/microsoft/OptiGuide/blob/main/optimind/data/comparisons/optmath_original_vs_ours.html}{\texttt{optmath\_original\_vs\_ours.html}}, which align the original instances with our cleaned versions. 

In addition, \href{https://github.com/microsoft/OptiGuide/blob/main/optimind/data/comparisons/industryOR_sirl_vs_ours.html}{\texttt{industryOR\_sirl\_vs\_ours.html}} compares our cleaned IndustryOR set against the SIRL-cleaned version: although the latter was reported as corrected, we identify five instances that still exhibit issues and document them in this file. Together, these tables offer a clear, human-readable view of the structure and details of our dataset corrections.

\newpage
\section{Limitations and Broader Impacts}\label{app:limitations-impact}
Our evaluations prioritize carefully reviewed optimization benchmarks, but their finite size, limited problem-family coverage, and remaining interpretation ambiguity restrict generalization claims. The nonlinear hints are derived from separate seed instances and evaluated on held-out problems. Objective agreement is a practical evaluation proxy, not a formal certificate that every constraint is correct. Generated formulations can support access to optimization expertise, but incorrect assumptions or constraints can lead to costly or unfair decisions. Deployment therefore requires validation of the intended variables, objectives, constraints, and resulting decisions, beyond checking solver status or objective values alone. Our implementation also depends on Gurobi for solver execution and on the model-generation and training resources described in Appendix~\ref{app:training-details}. The released full hint library, cleaned test data, cleaning logs, evaluation code, and trained OptiMind-RL model support inspection and reproducible evaluation (Appendix~\ref{app:anonymous_repo}).

\end{document}